\pdfoutput=1
\documentclass[11pt]{article}
\usepackage[final]{acl}
\usepackage{times} 
\usepackage{latexsym}
\usepackage[T1]{fontenc}
\usepackage[utf8]{inputenc}
\usepackage{microtype}
\usepackage{inconsolata}
\usepackage{graphicx}
\usepackage{enumitem}
\usepackage{soul}
\usepackage[colorinlistoftodos,textsize=scriptsize]{todonotes}
\usepackage{tabularx}
\usepackage{hyperref}
\usepackage{booktabs}
\usepackage{footmisc}
\usepackage{multirow}
\usepackage[most,skins,theorems]{tcolorbox}
\usepackage{caption, subcaption}
\usepackage[capitalise,nameinlink,noabbrev]{cleveref}
\usepackage{amssymb}
\usepackage[utf8]{inputenc}  
\usepackage[T1]{fontenc}     
\usepackage{fontawesome5}    
\usepackage{pifont}    
\usepackage{makecell}

\usepackage{soul}

\crefname{figure}{Figure}{Figures}
\definecolor{lightblue}{rgb}{0.22,0.45,0.70}%
\definecolor{gblue}{HTML}{4f79a7}
\definecolor{ggreen}{HTML}{77b7b2}
\definecolor{gred}{HTML}{e1575a}
\definecolor{gorange}{HTML}{f28e2a}
\definecolor{strongorange}{RGB}{255,87,34}
\definecolor{ppurple}{HTML}{603a70}
\definecolor{purple}{rgb}{0.5,0,0.5}
\definecolor{mydarkblue}{rgb}{0,0.08,0.45}

\newif\ifanirudh
\newif\ifyscomment
\newif\ifAAaddressed
\newif\ifmk
\newif\ifanmol

\anirudhfalse
\yscommentfalse
\AAaddressedfalse
\mkfalse
\anmolfalse
\newcommand{\Dataset}{\raisebox{0pt}[0.8ex][0pt]{\faDatabase}\quad\textbf{Dataset:}\,}
\newcommand{\Task}   {\faClipboardList\quad\textbf{Task:}\,}
\newcommand{\Metric} {\faChartLine\quad\textbf{Metric:}\,}
\newcommand{\deltacc}{$\Delta$-accuracy}
\newcommand{\qwen}{Qwen-2.5}
\newcommand{\llama}{Llama-3.1}
\newcommand{\oneruler}{\textsc{OneRuler}}
\newcommand{\ruler}{\textsc{Ruler}}
\newcommand{\nocha}{\textsc{NoCha}}
\newcommand{\veriscore}{\textsc{VeriScore}}
\newcommand{\factscore}{\textsc{FActScore}}
\newcommand{\csfour}{\textsc{CS4}}
\newcommand{\bookscore}{BooookScore}
\newcommand{\eqcont}{\textsuperscript{\scalebox{1.0}{$\bigstar$}}}

\usepackage{booktabs}
\usepackage{graphicx}
\usepackage{amsmath}
\usepackage{caption}
\usepackage{siunitx}

\usepackage{cleveref}
\usepackage{newfloat}     % allows figure* without caption if desired
\usepackage{graphicx}     % for \framebox
\usepackage{varwidth}     % flexible box width
\usepackage{xcolor}       % light‑gray background
\usepackage{etoolbox}  
\title{Does quantization affect models' performance on long-context tasks?}

\author{%
  Anmol Mekala\textsuperscript{$\bigstar$\faChessKnight} \quad
  Anirudh Atmakuru\textsuperscript{$\bigstar$\faChessKnight} \\  
  \texttt{\{amekala, aatmakuru\}@umass.edu} \\[1ex]
  \textbf{Yixiao Song}\textsuperscript{\faChessKnight} \quad
  \textbf{Marzena Karpinska}\textsuperscript{\faChessQueen} \quad
  \textbf{Mohit Iyyer}\textsuperscript{\faChessKnight}\textsuperscript{\faChessRook} \\  
  \texttt{yixiaosong@umass.edu} \quad
  \texttt{mkarpinska@microsoft.com} \quad
  \texttt{miyyer@umd.edu} \\[1ex] 
  \textsuperscript{\faChessKnight}UMass~Amherst \quad
  \textsuperscript{\faChessQueen}Microsoft \quad
  \textsuperscript{\faChessRook}University~of~Maryland,~College~Park
}

\begin{document}
\maketitle

\def\thefootnote{\eqcont}\footnotetext{These authors contributed equally to this work.}\def\thefootnote{\arabic{footnote}}

\begin{abstract}

Large language models (LLMs) now support context windows exceeding 128K tokens, but this comes with significant memory requirements and high inference latency. Quantization can mitigate these costs, but may degrade performance. In this work, we present the first systematic evaluation of quantized LLMs on tasks with long inputs ($\ge$64K tokens) and long-form outputs. Our evaluation spans 9.7K test examples, five quantization methods (FP8, GPTQ-int8, AWQ-int4, GPTQ-int4, BNB-nf4), and five models (\llama{} 8B and 70B; \qwen{} 7B, 32B, and 72B). 
We find that, on average, 8-bit quantization preserves accuracy (\textasciitilde0.8\% drop), whereas 4-bit methods lead to substantial losses, especially for tasks involving long-context inputs (drops of up to 59\%).
This degradation tends to worsen when the input is in a language other than English. 
Crucially,  the effects of quantization depend heavily on the quantization method, model, and task. For instance, while \qwen{} 72B remains robust under BNB-nf4, \llama{} 70B experiences a 32\% performance drop on the same task. These findings highlight the importance of a careful, task-specific evaluation before deploying quantized LLMs, particularly in long-context scenarios and for languages other than English. 
% Code and results at \href{https://github.com/molereddy/long-context-quantization}{\scriptsize \texttt{github.com/molereddy/long-context-quantization}}.

\end{abstract}
% \makeatletter
% \ifacl@linenumbers
%   % Do nothing in review mode
% \else
%   \blfootnote{*Equal contribution.}
% \fi
% \makeatother

\begin{center}
  \texttt{\scriptsize \faGithub \quad \href{https://github.com/molereddy/long-context-quantization}{github.com/molereddy/long-context-quantization}}
\end{center}
\section{Introduction}

Large language models (LLMs) serve a vast community of users through APIs and web services. However, their inference demands substantial resources, severely constraining scalability and increasing energy consumption \citep{samsi2023words, luccioni2024power,fernandez2025energy}. Post-training quantization aims to solve this problem by approximating models at lower precision \citep{zhou2024survey,zhu2024survey}. 8-bit quantization \citep{micikevicius2022fp8,frantar2023gptq} provides accuracy at half the cost, and more aggressive 4-bit approaches \citep{lin2024awq,dettmers_qlora_bnbint4,kim2023memory} offer even greater cost-effectiveness. Model providers widely offer quantized models \citep{meta2024llamaquant, google2024jetstream, databricks2023llminference}, but the impact on performance has not been explored exhaustively.

% % left bottom right top
\begin{figure}[t] 
    \centering    
    \includegraphics[width=\linewidth, clip, trim=0pt 0pt 50pt 0pt]{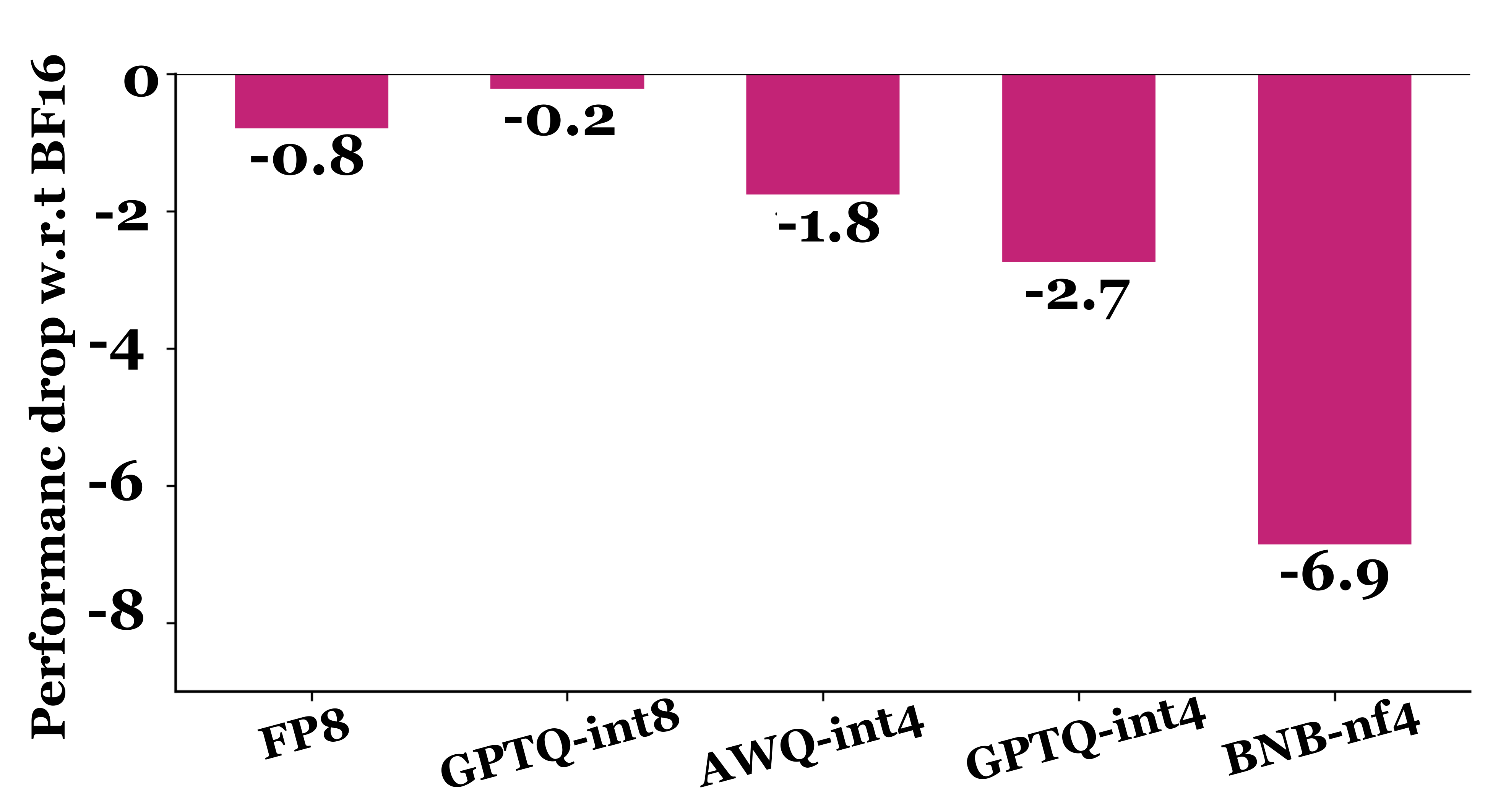}
    \caption{Average performance drop in percentage points caused by quantization on \textbf{long-context tasks}. The drop is computed by subtracting the performance of each quantization method (averaged over tasks and models) from the corresponding BF16 baseline.}
    \label{fig:overall-quant}
\end{figure}

\begin{figure*}[t]  
    \centering
    \resizebox{0.99\textwidth}{!}{
        \includegraphics{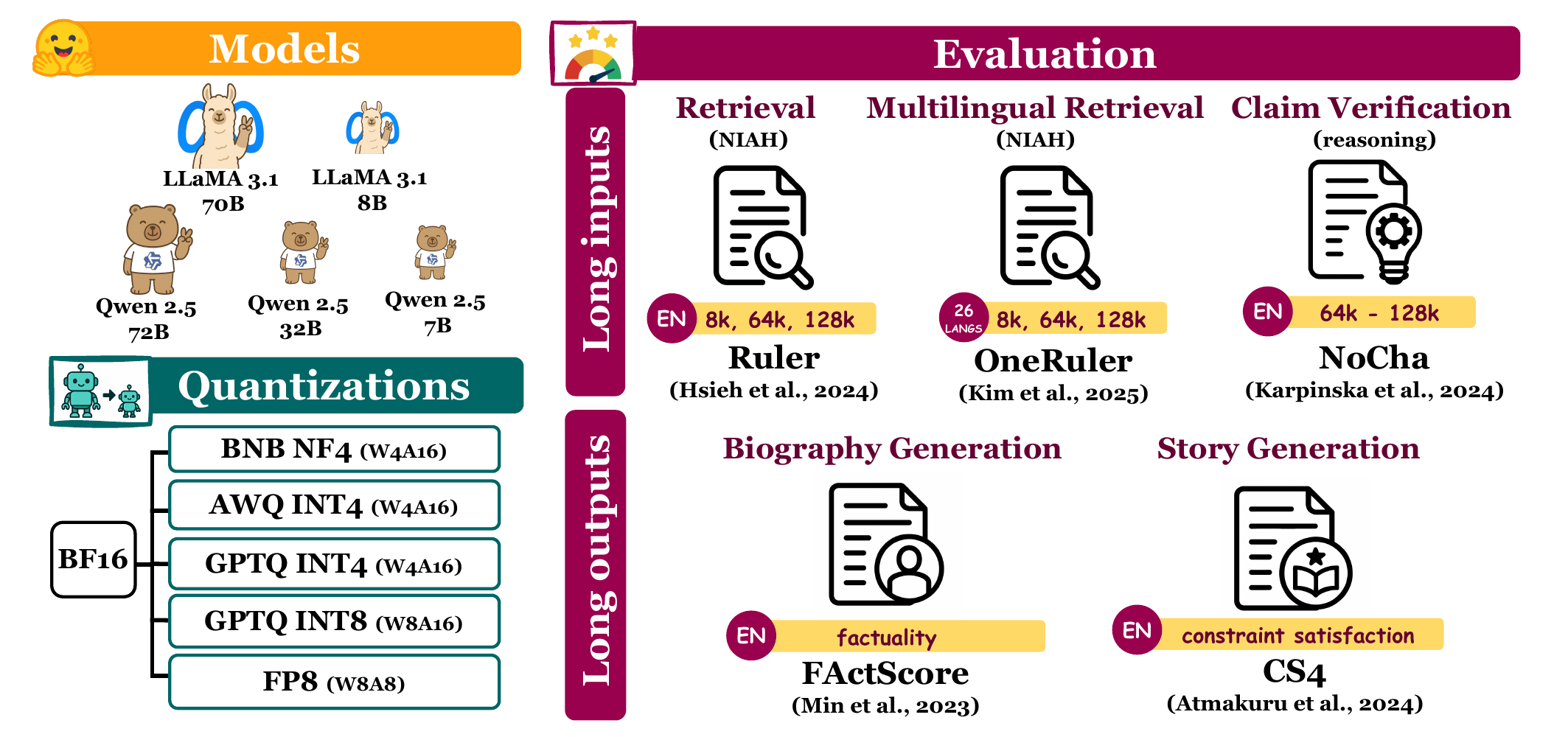}
    }
    \caption{Overview of the evaluation pipeline. We assess five models from two model families (right), comparing their performance in BF16 precision against their quantized variants across five quantization methods. The models are evaluated on three long-input tasks (top left) and two long-form output tasks (bottom left).}
    \label{fig:description}
\end{figure*}

While previous work has shown that quantized LLMs largely maintain performance on standard benchmarks \citep[inter alia]{badshah2024quantifying, liu2024quantizegeneralization, jin2024comprehensive, huang2024empirical}, these studies have mainly focused on tasks with short-form inputs and outputs \citep{jin2024comprehensive, lee2024comprehensive}, rarely exceeding 2K input and 100 output tokens. This limited scope leaves open a critical gap: the impact of quantization on tasks requiring long-context inputs or long-form outputs.\footnote{We define long-form inputs as over 64K tokens, and long-form outputs as typically 250-650 tokens long.}
This is particularly salient because, as context length increases, the associated memory and compute requirements grow quadratically, making low-precision methods particularly attractive for efficiency. 
Moreover, many modern LLMs now support context windows of 128K tokens or more, allowing them to process entire codebases, legal documents, or full novels. Such models, when aggressively quantized (e.g., to 4 bits), could theoretically handle significantly more context under the same memory budget.

However, to our knowledge, only \citet{kurtic2024give} have benchmarked long-context performance, and their study was confined to the \ruler{} \citep{hsiehruler} dataset. Critically, they only reported a single, aggregated score and did not provide a breakdown of performance at different context lengths, which can mask significant performance drops on very long inputs.

% However, to the best of our knowledge, only \citet{kurtic2024give} include the \ruler{} \citep{hsiehruler} long-context task, but their coarse result aggregation across tasks and context lengths may obscure the effects of individual factors. Thus, it remains unclear whether quantized LLMs can maintain accuracy on very long inputs or generate high-quality long-form outputs.

To address this gap, we conduct a systematic analysis of quantization performance across multiple models, quantization methods, and benchmark tasks. This includes five datasets across 9.7K test examples, with key factors varying such as context length, language, and output constraints.
We evaluate five quantization methods (FP8, GPTQ-int8, AWQ-int4, GPTQ-int4, and BNB-NF4) on five models (\llama{} 8B and 70B; \qwen{} 7B, 32B, and 72B), comparing against their BF16 baselines. For long-context input tasks, we use \ruler{} \citep{hsiehruler} and \oneruler{} \citep{kim2025one} to assess retrieval, and \nocha{} \citep{karpinska2024one} to test reasoning over book-length texts. For long-form output tasks, we employ \factscore{} \citep{min2023factscore} to evaluate factuality in biography generation and \csfour{} \citep{atmakuru2024cs4} to assess constrained story generation (see \autoref{fig:description}). The results reveal the following:

\vspace{3pt}

\setlist{nolistsep}
\begin{enumerate}[leftmargin=12pt, nolistsep]

    \item \textbf{The 8-bit quantized models remain mostly robust}, with FP8 and GPTQ-int8 showing average drops of 0.2\% and 0.8\%, respectively.\footnote{This result is in line with prior studies \citep{lee2024comprehensive,kurtic2024give}.} \textbf{4-bit methods incur larger losses}, with AWQ showing 1.8\%, GPTQ-int4 2.7\%, and BNB-nf4 6.9\% average drops. BNB-nf4 can be particularly inaccurate on long contexts, with up to 59\% loss on \oneruler{} for \llama{} 70B. 
    
    \item \textbf{Quantization degrades long-input retrieval performance}, with drops worsening as input length increases. 4-bit quantization shows the most significant drop, with an average decrease of up to 23\% across models at 128K tokens.

    \item \textbf{The adverse effects of quantization on performance are more severe in languages other than English.} Our long-context retrieval experiments reveal that, under quantization, accuracy in these languages can drop up to five times more than in English.

    \item \textbf{The impact of quantization is not uniform between models.} While \qwen{} 72B shows minimal loss under BNB-nf4 across tasks, similarly sized \llama{} 70B suffers a 32\% drop.
   
\end{enumerate}

\section{Experimental setup}
\label{sec:exp_setup}

This section outlines our experimental setup. We evaluate five models from two model families (Llama and Qwen) across six precision settings on five benchmarks. Additional details are in \S\ref{app:sec-exp_setup}.

\paragraph{Models:} We evaluate five open-weight models from two families: \textbf{Llama 3.1} 8B and 70B \citep{grattafiori2024llama} and \textbf{Qwen 2.5} 7B, 32B, and 72B \citep{qwen2.5,yang2024qwen2}. All models support up to 128K input tokens.\footnote{\qwen{} models originally support 32k tokens but can be extended to 128K using \textsc{YaRN} \citep{peng2024yarn}. See \S\ref{app:qwen_long_ctx} for details.}

\paragraph{Quantization methods:} We use BF16 as the full-precision baseline and compare four quantization methods at 4-bit and 8-bit precisions against it: 

\begin{enumerate}[leftmargin=12pt,itemsep=0pt]
    \item \textbf{AWQ}: 4-bit integer ({\small W4A16})\footnote{\label{wxay}W$x$A$y$ denotes $x$-bit weights and $y$-bit activations.} quantization \citep{lin2024awq}, referred to as AWQ-int4.
    
    \item \textbf{BNB}: 4-bit floating-point quantization using bitsandbytes ({\small W4A16}) \citep{dettmers_qlora_bnbint4}, referred to as BNB-nf4.
    
    \item \textbf{GPTQ}: (a) 4-bit integer ({\small W4A16}), referred to as GPTQ-int4, and (b) 8-bit integer ({\small W8A16}), referred to as GPTQ-int8 \citep{frantar2023gptq}.
    
    \item \textbf{FP8}: Dynamic 8-bit floating-point quantization ({\small W8A8}) \citep{micikevicius2022fp8}, referred to as FP8.
\end{enumerate}

\noindent In total, we benchmark five models across five quantization setups and the BF16 baselines, resulting in 30 configurations.

\paragraph{Inference setup:} All models are run using {\small \textsc{vLLM}} \citep{kwon2023efficientvllm} on one to four A100-80G GPUs, except for the FP8 models, which require H100 GPUs due to the lack of FP8 support on A100 \citep{vllmdocs}. All experiments use greedy decoding with a temperature of $0.0$ to ensure deterministic outputs.

\paragraph{Benchmark tasks:}
\label{para:benchmark_tasks} The BF16 models and their quantized variants are evaluated on five benchmarks: (1) \textbf{\ruler{}} \citep{hsiehruler} for English long‑context retrieval, (2) \textbf{\oneruler{}} \citep{kim2025one} for long‑context retrieval across languages, (3) \textbf{\nocha{}} \citep{karpinska2024one} for long‑context reasoning, (4) \textbf{\factscore{}} \citep{min2023factscore} for factual biography generation, and (5) \textbf{\csfour{}} \citep{atmakuru2024cs4} for constrained creative story generation. The first three tasks evaluate model performance on long inputs, while the last two focus on long‑form outputs. See \S\ref{app:sec-benchmarks} for more details.

\paragraph{\Dataset} \ruler\ \cite{hsiehruler}
\vspace{-0.4em}
\paragraph{\Task} \ruler\ evaluates model's \emph{effective context window} with needle‑in‑a‑haystack ({\small NIAH}) retrieval tasks in English. The model must locate one or more target key-value pairs (the ``needles'') hidden in passages of varying lengths (the ``haystacks''). Following \citet{yen2024helmet}, the task includes three variants: (1) \textit{multi‑key needle}, (2) \textit{multi‑key UUID}, and (3) \textit{multi‑value}. Each variant is tested at three context lengths (8K, 64K, and 128K),\footref{rulercontext} with emphasis on longer contexts (64K and 128K). Each {\{\texttt{task, context-length}\}} combination includes 100 examples, yielding 900 test examples in total.
\vspace{-0.4em}
\paragraph{\Metric} Following \cite{hsiehruler}, performance is measured using exact match accuracy against the gold answer.
\vspace{0.2em}
\paragraph{\Dataset} \oneruler\ \cite{kim2025one}
\vspace{-0.4em}
\paragraph{\Task} \oneruler{} extends \ruler{} by evaluating {\small NIAH} retrieval tasks not only in English but also across multiple other languages. It also introduces no-needle scenarios, where models must explicitly predict the absence of a needle. \oneruler{} serves as an important complement to \ruler{} in light of the performance degradation of quantized models observed by \citet{marchisio2024does} on non-Latin scripts. Each model is evaluated on two tasks (NIAH-single and NIAH-none; see \S\ref{app:sec-benchmarks}), three context lengths (8K, 64K, and 128K),\footnote{\label{rulercontext}Unless explicitly stated, we report average performance across 64K and 128K context lengths, as our focus is on long-context scenarios. Results at 8K are included only in a targeted ablation to highlight the impact of extended context.} and 26 languages. Each \texttt{\{task, context-length, language\}} combination consists of 50 examples, totaling 7,800 test examples.
\vspace{-0.3em}
\paragraph{\Metric} As with \ruler{}, exact match accuracy is used as the metric.
\vspace{0.2em}
\begin{figure*}[ht]
    \centering
    \resizebox{0.90
    \linewidth}{!}{
        \includegraphics[clip, trim=0pt 0pt 0pt 0pt]{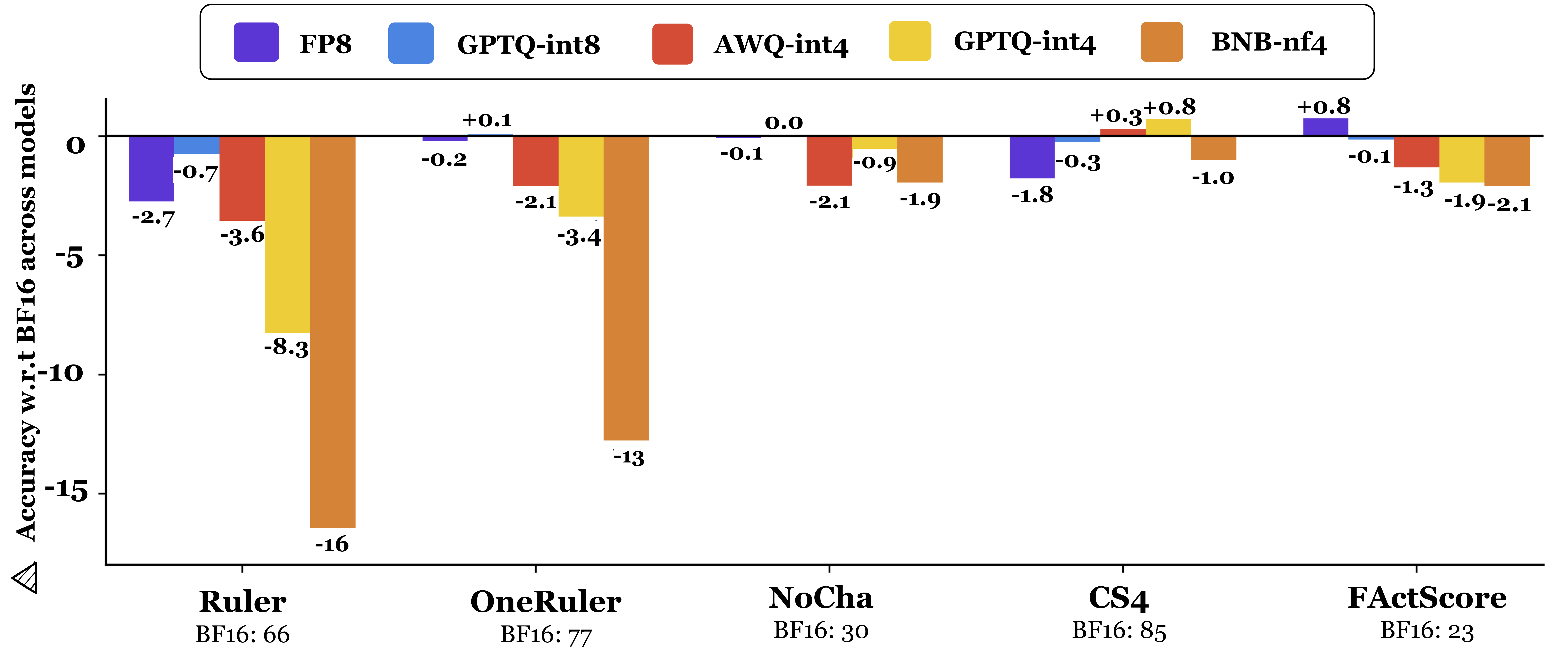}
    }
  \caption{Average performance of each quantization method by task, computed by averaging across all models. Performance is reported using \deltacc{}, representing the difference relative to the BF16 baseline. The BF16 baseline values (averaged across models for each task) are shown below the task names. For \oneruler{} and \ruler{}, only long-context results from the 64K and 128K context lengths are included in the aggregation.}
  \label{fig:all_taskwise}
\end{figure*}

\paragraph{\Dataset} \nocha\ \cite{karpinska2024one}
\vspace{-0.4em} 
\paragraph{\Task} \nocha\ evaluates long-context reasoning by asking models to verify claims against book-length contexts. Each example includes a minimal pair of true and false claims with subtle but meaningful differences. Models are prompted to explain their reasoning before making a decision. The evaluation includes data points with contexts up to 128K tokens. In total, the dataset contains 723 claim pairs drawn from 46 novels and three short story collections.
\vspace{-0.4em}
\paragraph{\Metric} Following \citet{karpinska2024one}, we report accuracy at the claim-pair level: an answer is considered correct only if the model correctly labels both claims in the pair (i.e., identifies which is true and which is false).
\vspace{0.2em}
\paragraph{\Dataset} \factscore\ \cite{min2023factscore}
\vspace{-0.4em}
\paragraph{\Task} \factscore\ evaluates a model's ability to generate factual long-form biographies for named entities. The evaluation covers 100 entities, evenly distributed across three frequency categories: \textit{very rare}, \textit{medium}, and \textit{very frequent}.\footnote{We adopt the definition from \factscore{} \citep{min2023factscore}, which uses Wikipedia occurrences as frequency.} 
\vspace{-0.4em}
\paragraph{\Metric} We report \veriscore\ \citep{song2024veriscore} as the factuality metric. Responses without verifiable content receive a score of zero.

\vspace{0.2em}

\paragraph{\Dataset} \csfour\ \cite{atmakuru2024cs4}
\vspace{-0.4em}
\paragraph{\Task} \csfour\ evaluates a model's instruction-following and story-writing abilities by measuring how well it generates coherent short stories under increasing levels of constraint (9, 23, and 39 constraints). Each model generates 150 stories in total, with 50 per constraint level.
\vspace{-0.4em}
\paragraph{\Metric} Following \citet{atmakuru2024cs4}, we use GPT-4o to evaluate constraint satisfaction \citep{liu2023g}, computed per story as 
$100 \times \frac{\text{\# of satisfied constraints}}{\text{\# of total constraints}}$. We use \bookscore\ \citep{chang2024booookscore} to evaluate coherence and internal consistency of the generated stories.\footnote{We use \texttt{gpt-4o-2024-08-06} as the judge LLM for constraint satisfaction and BooookScore.} %We also report Self-BLEU \citep{zhu2018texygen} and 5-gram repetition rates to assess lexical diversity. \mk{we don't use selfbleu anyway, and 5-gram repetition fig 29 is unclear to me, so may drop, i guess looking if there are loopy repetitions would make some more sense} 

\section{Results and discussion}
\label{sec:results}

\begin{figure}[hbt!]
  \centering

  \begin{subfigure}[b]{\columnwidth}
    \centering
    \includegraphics[
      clip,
      trim=0pt 0pt 0pt 0pt,
      width=\linewidth
    ]{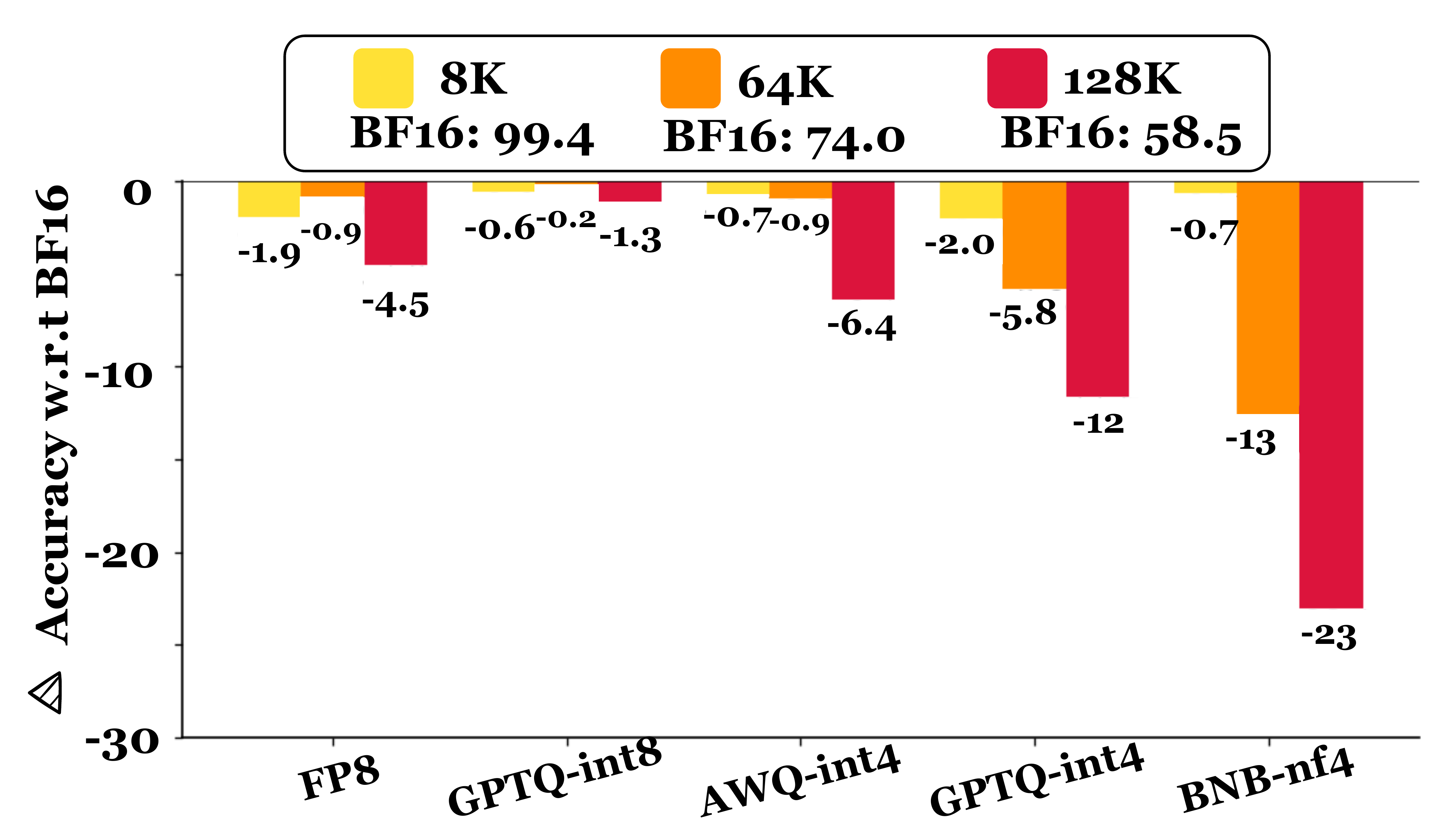}
    %\vspace{-2pt}
    \caption{\textbf{\ruler{}:} drop in performance at three context lengths.}
    \label{fig:ruler-contextwise}
  \end{subfigure}

  \vspace{1em}

  \begin{subfigure}[b]{\columnwidth}
    \centering
    \includegraphics[
      clip,
      trim=0pt 5pt 0pt 5pt,
      width=\columnwidth
    ]{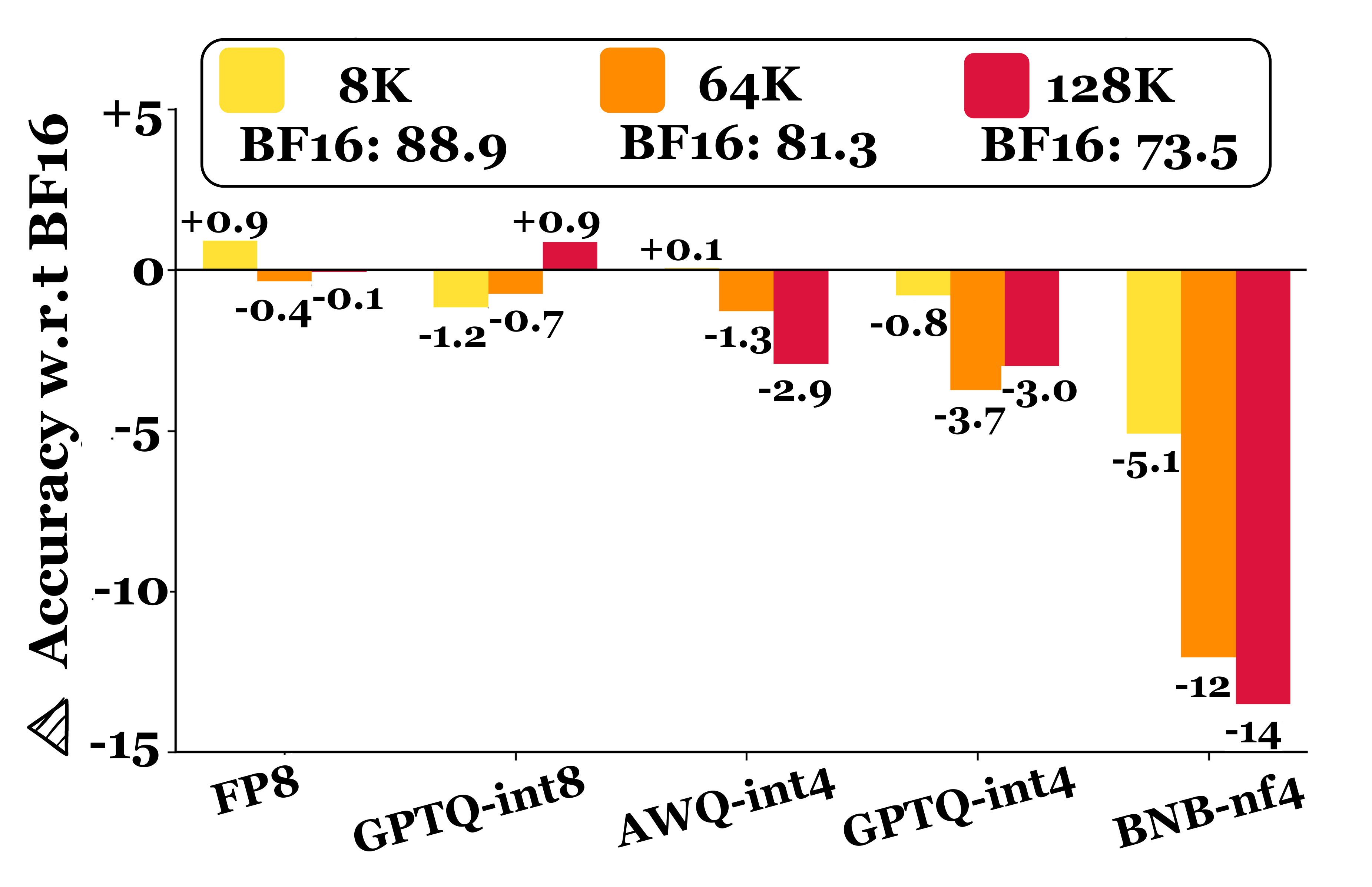}
    %\vspace{-5pt}
    \caption{\textbf{\oneruler:} drop in performance at three context lengths.}
    \label{fig:oneruler-contextwise}
  \end{subfigure}

  \vspace{1em}

  \begin{subfigure}[b]{\columnwidth}
    \centering
    \includegraphics[
      clip,
      trim=0pt 5pt 0pt 5pt,
      width=\linewidth
    ]{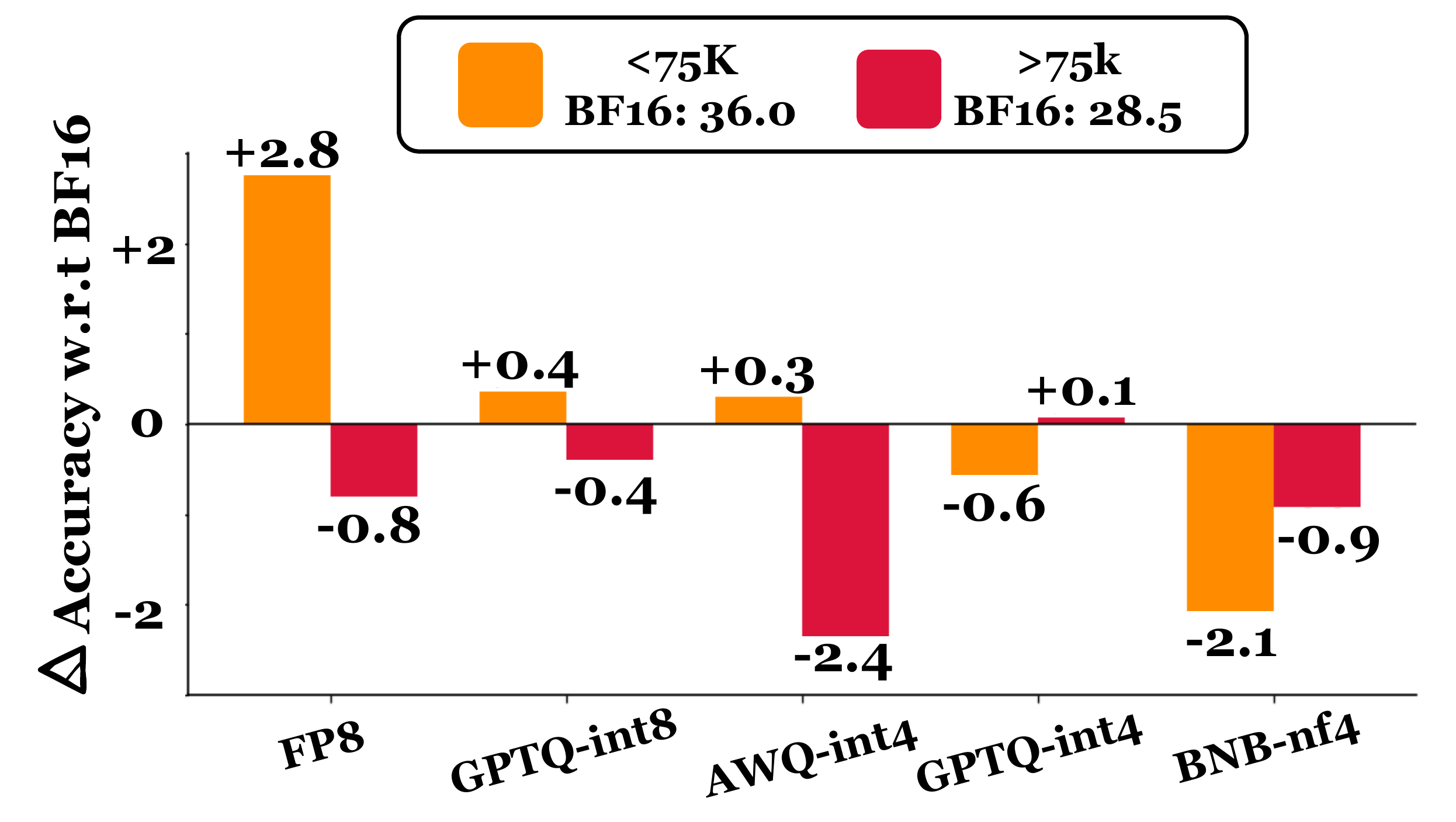}
    %\vspace{-2pt}
    \caption{\textbf{\nocha{}:} drop in performance at varying context lengths.}
    \label{fig:nocha-context-length}
  \end{subfigure}

  \vspace{-5pt}
  \caption{Quantization performance by context length for three long-context benchmarks (\ruler{}, \oneruler{}, and \nocha{}), shown as \deltacc{} values averaged across models for each quantization method. The corresponding BF16 baseline scores (model-averaged) are shown in the legend.}
  \label{fig:quant-context-length-combined}
\end{figure}

This section presents the results of how quantization affects performance on the benchmarks in \S\ref{sec:exp_setup}. We examine results at the following levels of detail: model family, task category, and sub-task settings such as context length and language. For each setting, we report \deltacc{}, defined as the change in performance between the BF16 model and its quantized version:
\vspace{-2pt}
\[
  \Delta\text{-accuracy} = 
    \operatorname{acc}_{\text{quantized}} - \operatorname{acc}_{\text{BF16}}
\]
% \mk{for now i put it as clear equation could merge withing text if short on space but this is important to make really clear!}
\vspace{-8pt}
% We also repo

\noindent{We summarize key findings below.}

\begin{figure*}[hbt!]
 \centering

 \vspace{1em}

 \begin{subfigure}[b]{0.8\textwidth}
   \centering
    \includegraphics[width=\textwidth]{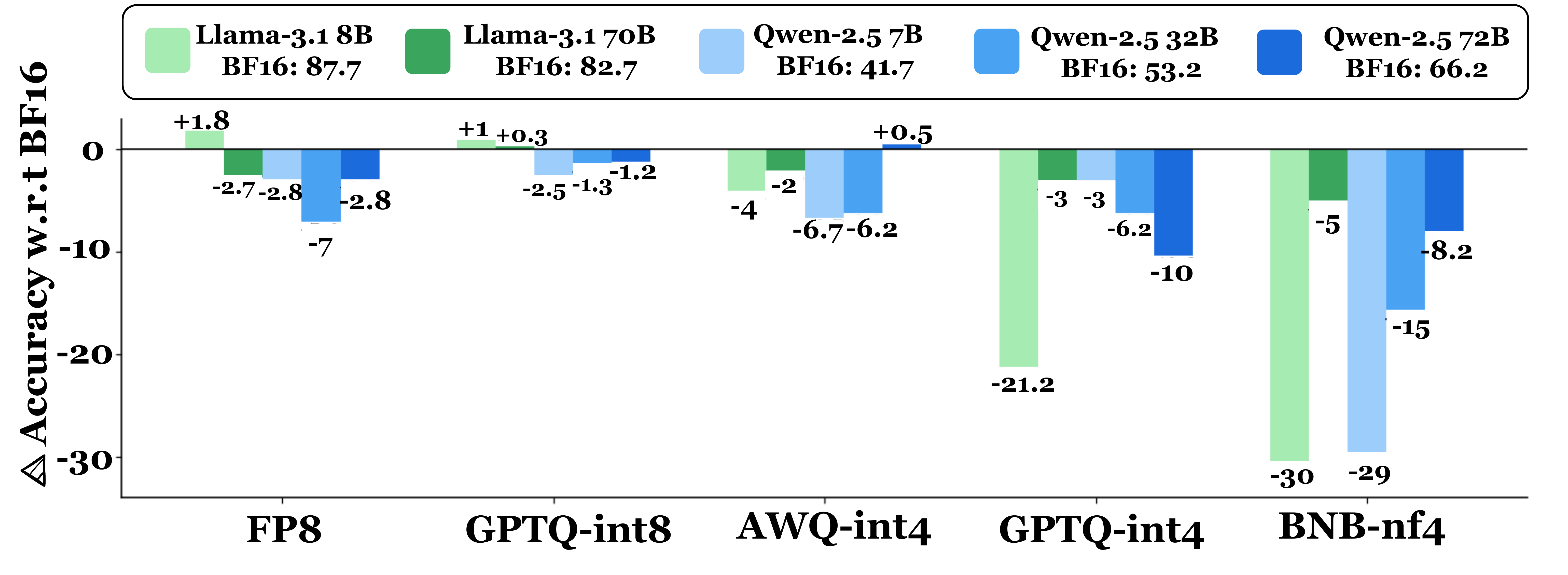}
    \caption{\textbf{\ruler:} Average $\Delta$-accuracy relative to BF16 across models.}
    \label{fig:ruler-model-wise}
  \end{subfigure}

  \vspace{1em}

  \begin{subfigure}[b]{0.8\textwidth}
    \centering
    \resizebox{\textwidth}{!}{\includegraphics[clip, trim=5pt 5pt 10pt 5pt]{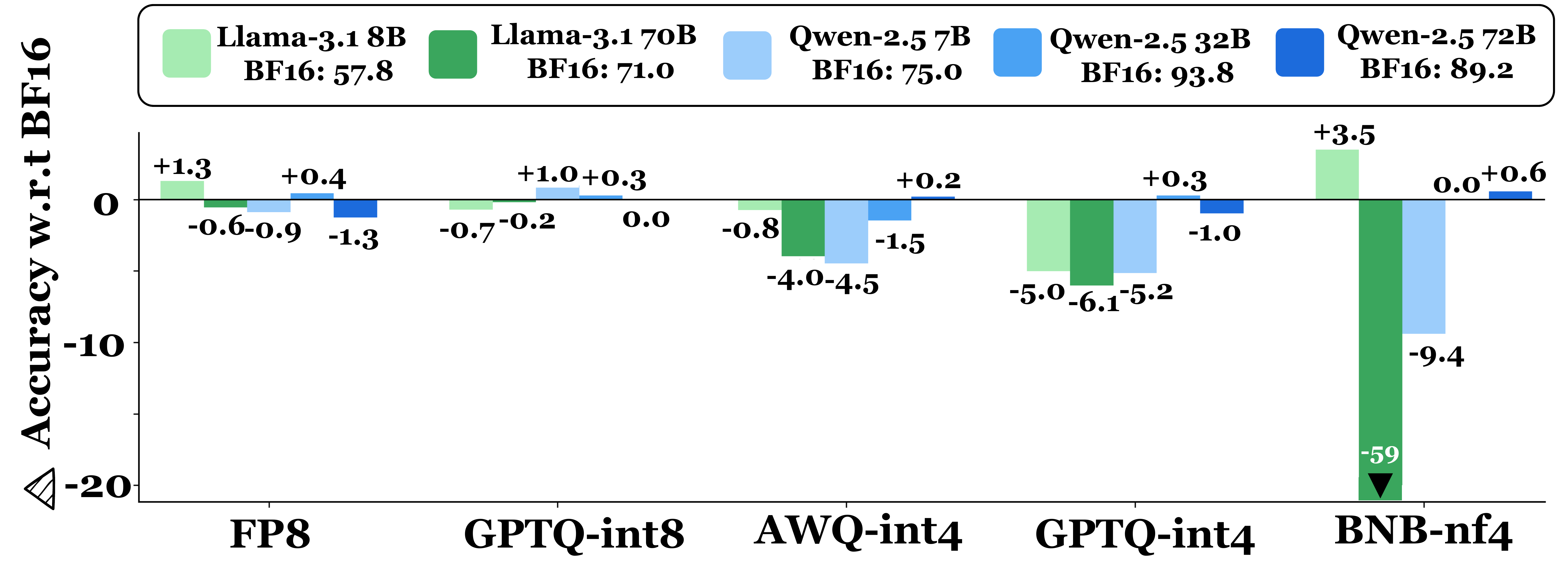}}
    \caption{\textbf{\oneruler:} Average $\Delta$-accuracy relative to BF16 across models.}
    \label{fig:oneruler_modelwise}
  \end{subfigure}

  \vspace{1em}

  \begin{subfigure}[b]{0.8\textwidth}
    \centering
    \includegraphics[width=\textwidth]{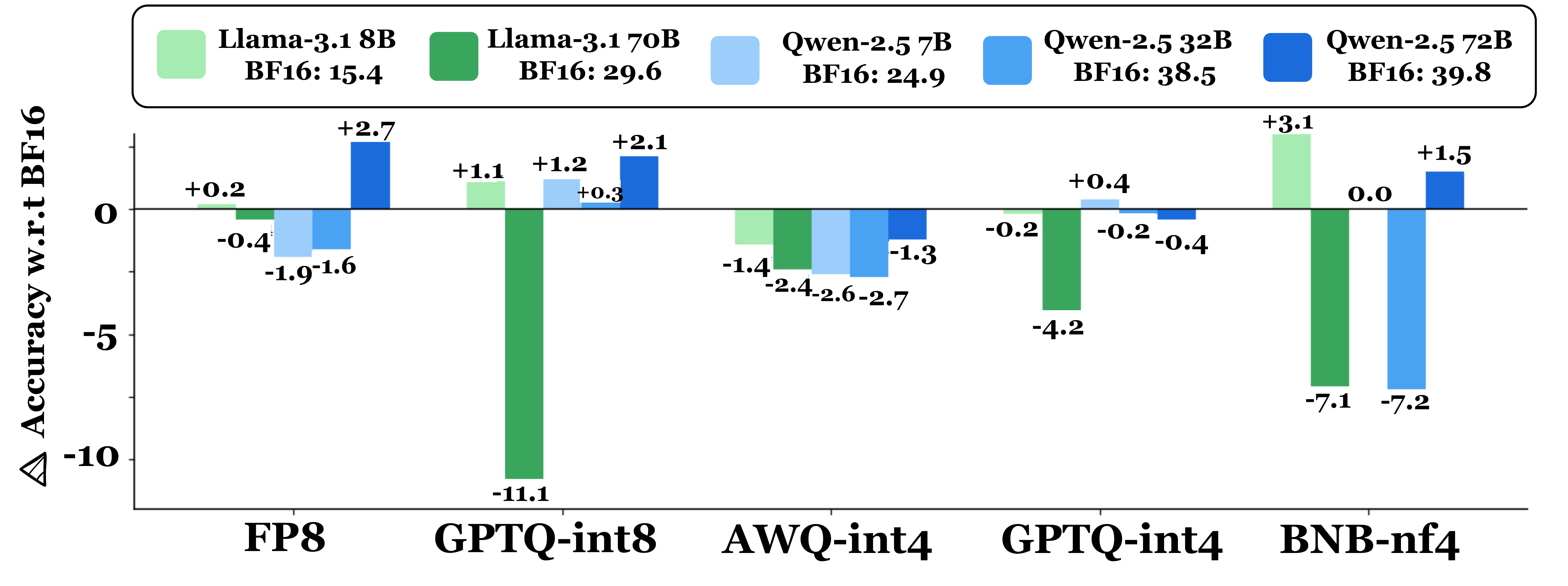}
    \caption{\textbf{\nocha:} Average $\Delta$-accuracy relative to BF16 across models.}
    \label{fig:nocha_modelwise}
  \end{subfigure}

  \vspace{-5pt}
  \caption{Average performance of quantization methods on long-context tasks, by model. The BF16 baseline results over which deltas are computed are given in the legend. \ruler{} and \oneruler{} results are computed over long (64K, 128K) context lengths only.}
  \label{fig:longctx-modelwise}
\end{figure*}

\paragraph{Long-context retrieval performance degrades with 4-bit quantization.}
Long-context retrieval tasks are particularly sensitive to aggressive compression (\autoref{fig:all_taskwise}, \textit{p}<0.05).\footnote{We report the details of statistical analysis reported throughout this section in \S\ref{app:stats-test}.} Specifically, accuracy on \ruler{} and \oneruler{} drops by up to 16\% when model precision is reduced to 4-bits, regardless of the quantization method. For the \nocha{} long-context reasoning benchmark, the overall accuracy decrease is smaller. This is likely because the accuracy of the baseline BF16 is already near chance level around 30\%, leaving little room for further degradation.\footnote{As the model has to evaluate correctly two claims in the pair, the random baseline is at 25\%.} Nevertheless, model performance still tends to decline slightly for most quantization methods as input length increases from $<$75K to $>$75K tokens (\autoref{fig:nocha-context-length}). While the extent of this drop varies, it can be observed across most models for all three tasks (\autoref{fig:longctx-modelwise}).
This sensitivity of quantization performance at long contexts is possibly due to the rounding errors in RoPE embeddings accumulated over long context  \citep{wang2025when}.

\begin{figure}[t]
  \centering
  \begin{subfigure}[t]{\linewidth}
    \includegraphics[clip,trim=0pt 5pt 0pt 5pt,width=\linewidth]{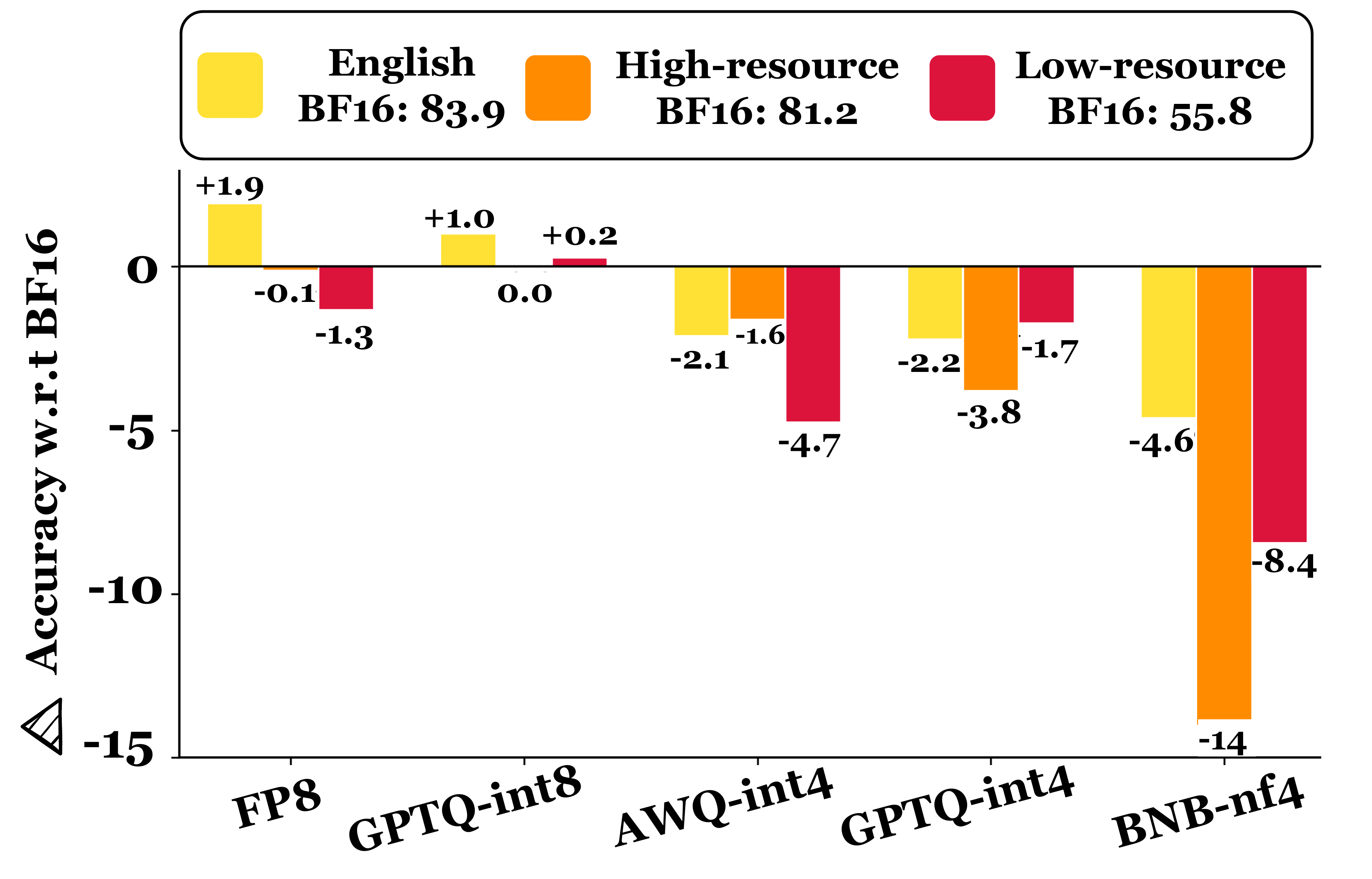}
    \caption{\textbf{\oneruler:} drop in performance by language resource level}
    \label{fig:oneruler-barplot-resource}
  \end{subfigure}
  \vspace{4pt}
  \begin{subfigure}[t]{\linewidth}
    \includegraphics[clip,trim=0pt 5pt 0pt 5pt,width=\linewidth]{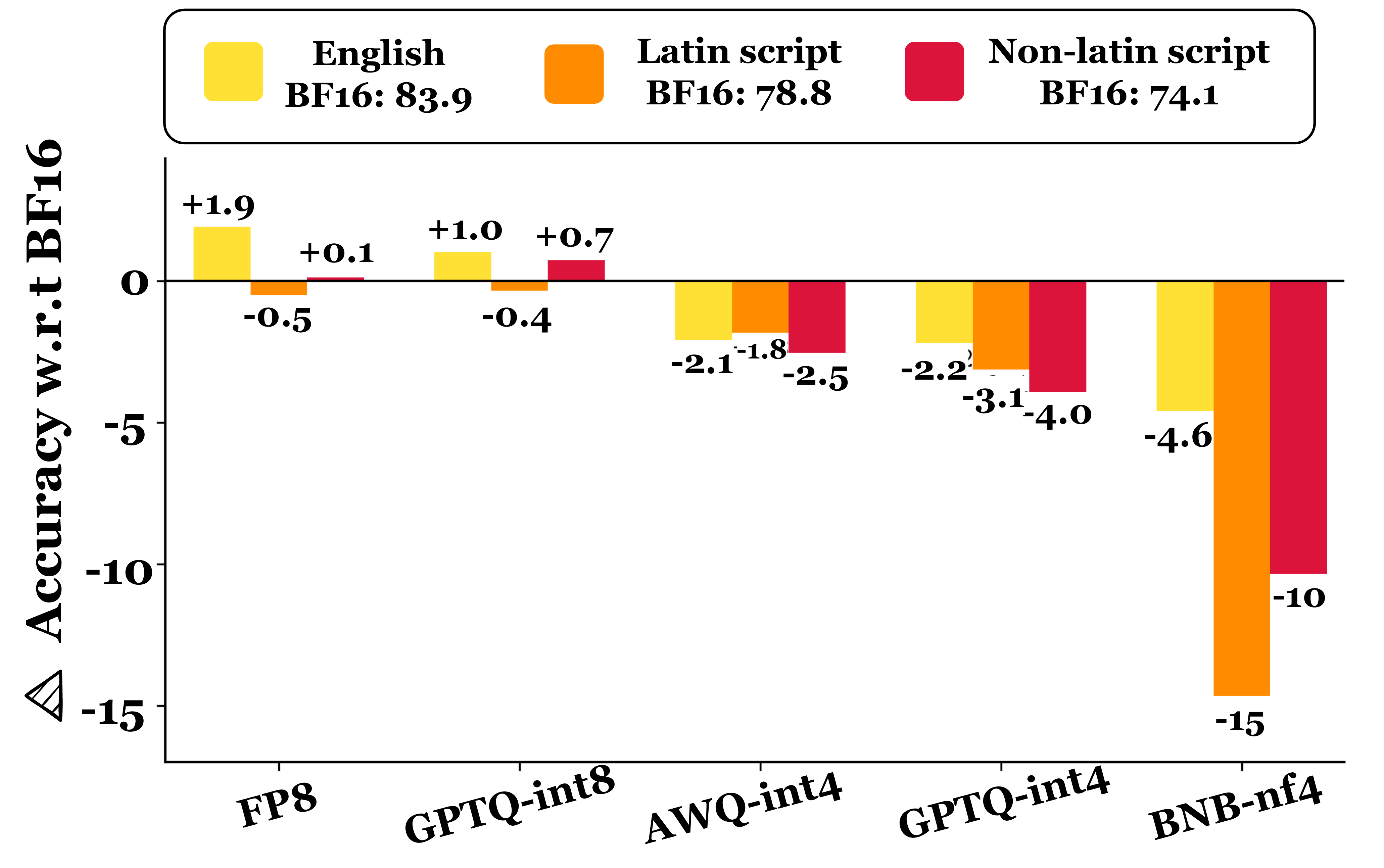}
    \caption{\textbf{\oneruler:} drop in performance by script family}
    \label{fig:oneruler-barplot-script}
  \end{subfigure}
  \caption{\textbf{\oneruler:} Average performance of each quantization method on each model, aggregated over long-context results (64K and 128K lengths). Performance is shown as \deltacc{}, the difference relative to the BF16 baseline.}
  \label{fig:oneruler-barplots}
\end{figure}

\paragraph{Quantization Degrades Accuracy More Severely for High- and Low-Resource Languages.}
Our analysis of \oneruler{} performance groups languages into three categories: English, other high-resource languages, and low-resource languages.\footnote{We adopt the definition from \oneruler{} \citep{kim2025one}, which classifies languages with under 250k Wikipedia articles (Hindi, Sesotho, Swahili, and Tamil) as low-resource.} While 8-bit quantization methods like FP8 and GPTQ-int8 show minimal impact across all groups (ranging from +1.9\% to –1.3\% vs. the BF16 baseline), 4-bit methods uniformly degrade performance (see \autoref{fig:oneruler-barplot-resource}).

Specifically, larger drops occur for \textbf{low-resource languages} when using AWQ-int4 (–4.7\%) and BNB-nf4 (–8.4\%). Similarly, \textbf{high-resource languages} face significant performance decreases with GPTQ-int4 (–3.8\%) and particularly with BNB-nf4 (–14\%). While performance on English also drops, the degradation range is narrower (–2.1\% to –4.6\%). Post‑hoc tests confirm that drops for non‑English languages are statistically significant ($p$<0.05, \autoref{tab:oneruler-posthoc-frequency} in \S\ref{app:stats-test}).\footnote{We also report performance by language script following \citet{marchisio2024does}, in (\autoref{fig:oneruler-barplot-script}). However, we do not observe a consistent pattern where quantization disproportionately affects languages with non-Latin scripts. Instead, the performance drops vary depending on the quantization method: for example, GPTQ-int4 results in slightly larger drops for non-Latin script languages, while BNB-nf4 leads to greater degradation for languages using the Latin script.}

% the 5x1 version
% Variant 2: 5×1 vertical layout in figure*

\begin{figure*}[hbt!]
 \centering

 \vspace{1em}
  \begin{subfigure}[b]{\textwidth}
    \centering
    \includegraphics[width=0.8\textwidth]{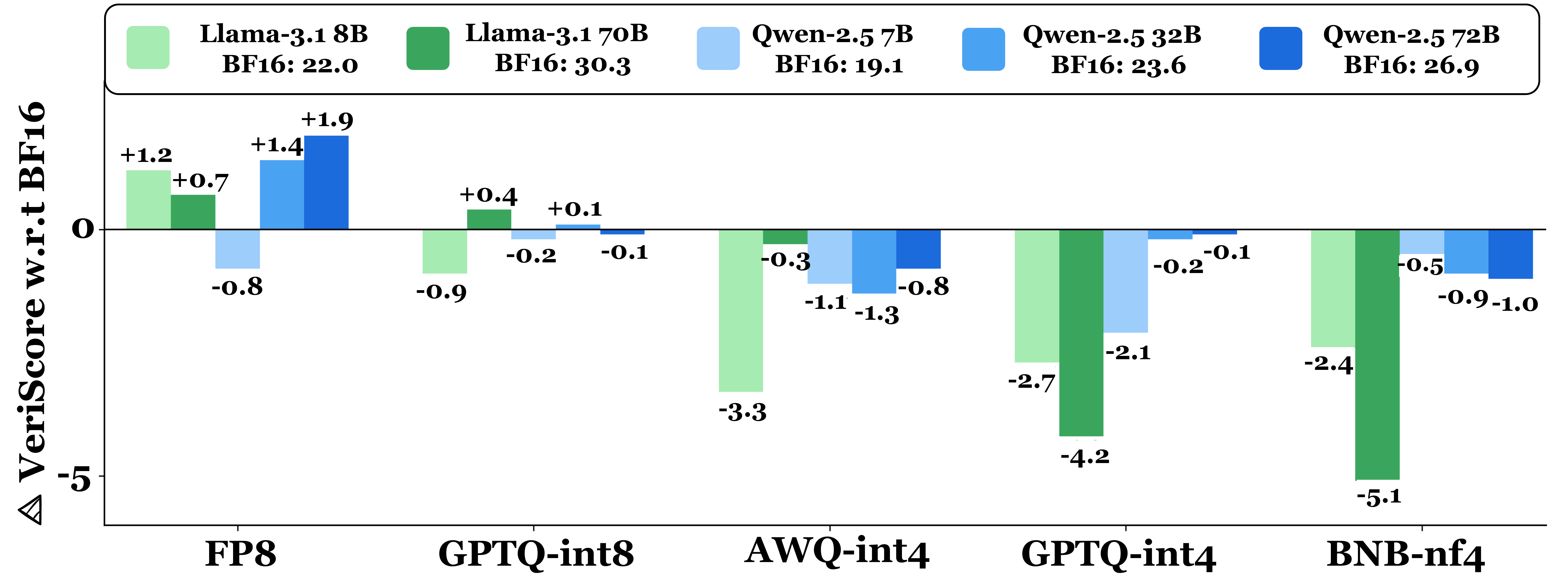}
    \caption{\textbf{\factscore:} Average $\Delta$-\veriscore{} across models.}
    \label{fig:veriscore_modelwise}
  \end{subfigure}

  \vspace{1em}

  \begin{subfigure}[b]{0.8\textwidth}
    \centering
    \includegraphics[width=\textwidth]{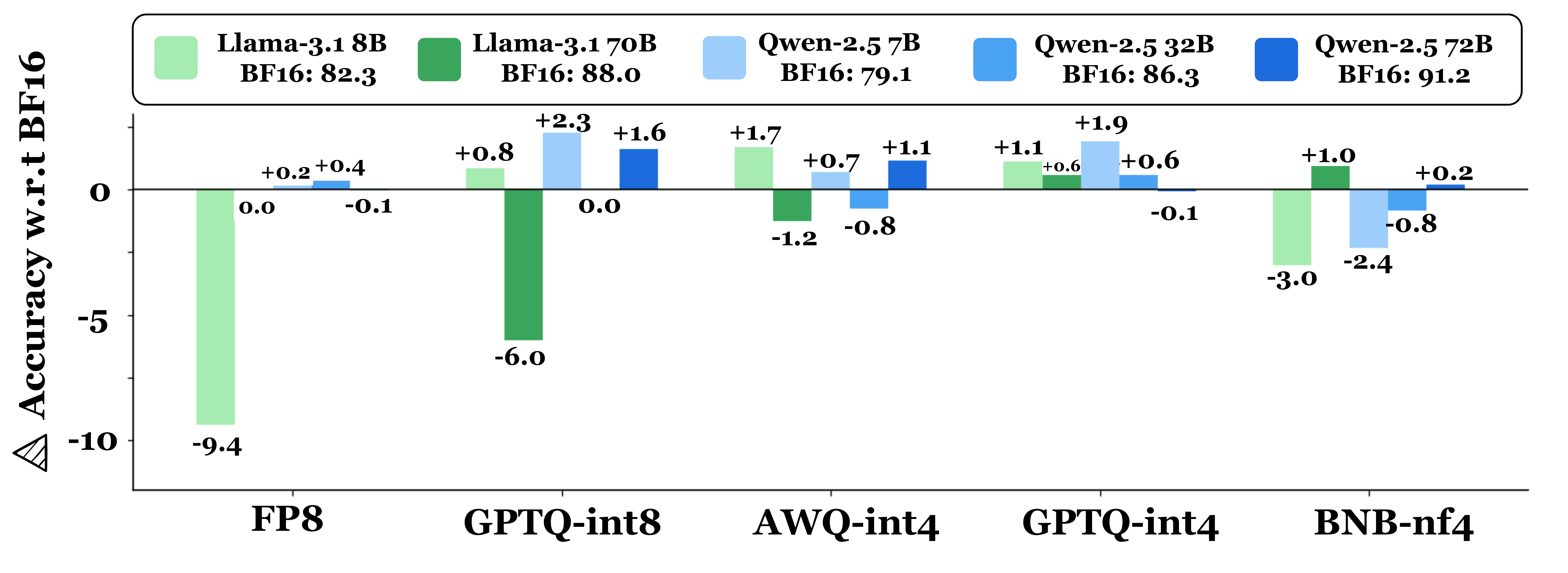}
    \caption{\textbf{\csfour:} Average $\Delta$-constraint-satisfaction across models.}
    \label{fig:cs4_modelwise}
  \end{subfigure}

  \vspace{1em}
  \begin{subfigure}[b]{0.8\textwidth}
    \centering
    \includegraphics[width=\textwidth]{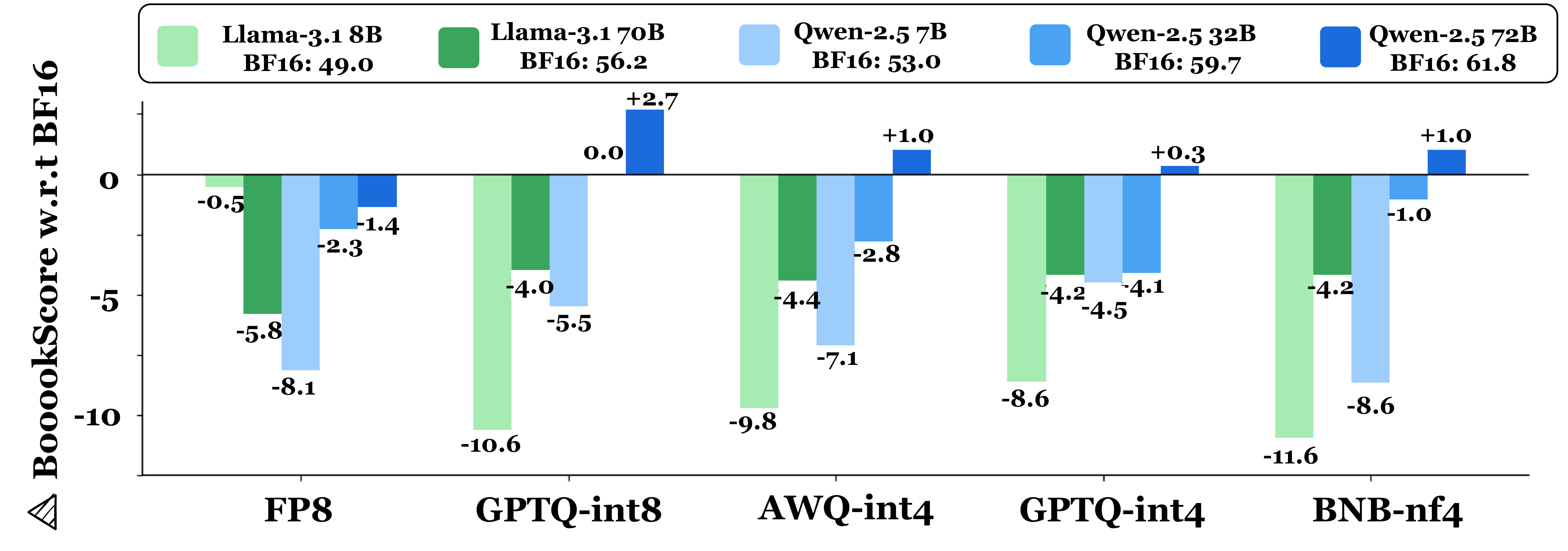}
    \caption{\textbf{\csfour:} Average $\Delta$-BooookScore across models.}
    \label{fig:BooookScore_modelwise}
  \end{subfigure}
  \vspace{-5pt}
  \caption{Average performance of quantization methods on long-form generation tasks, by model. The BF16 baseline results over which deltas are computed are given in the legend.}
  \label{fig:longgen-modelwise}
\end{figure*}

% left bottom right top
\begin{figure}[t] 
    \centering    \includegraphics[width=0.9\linewidth]{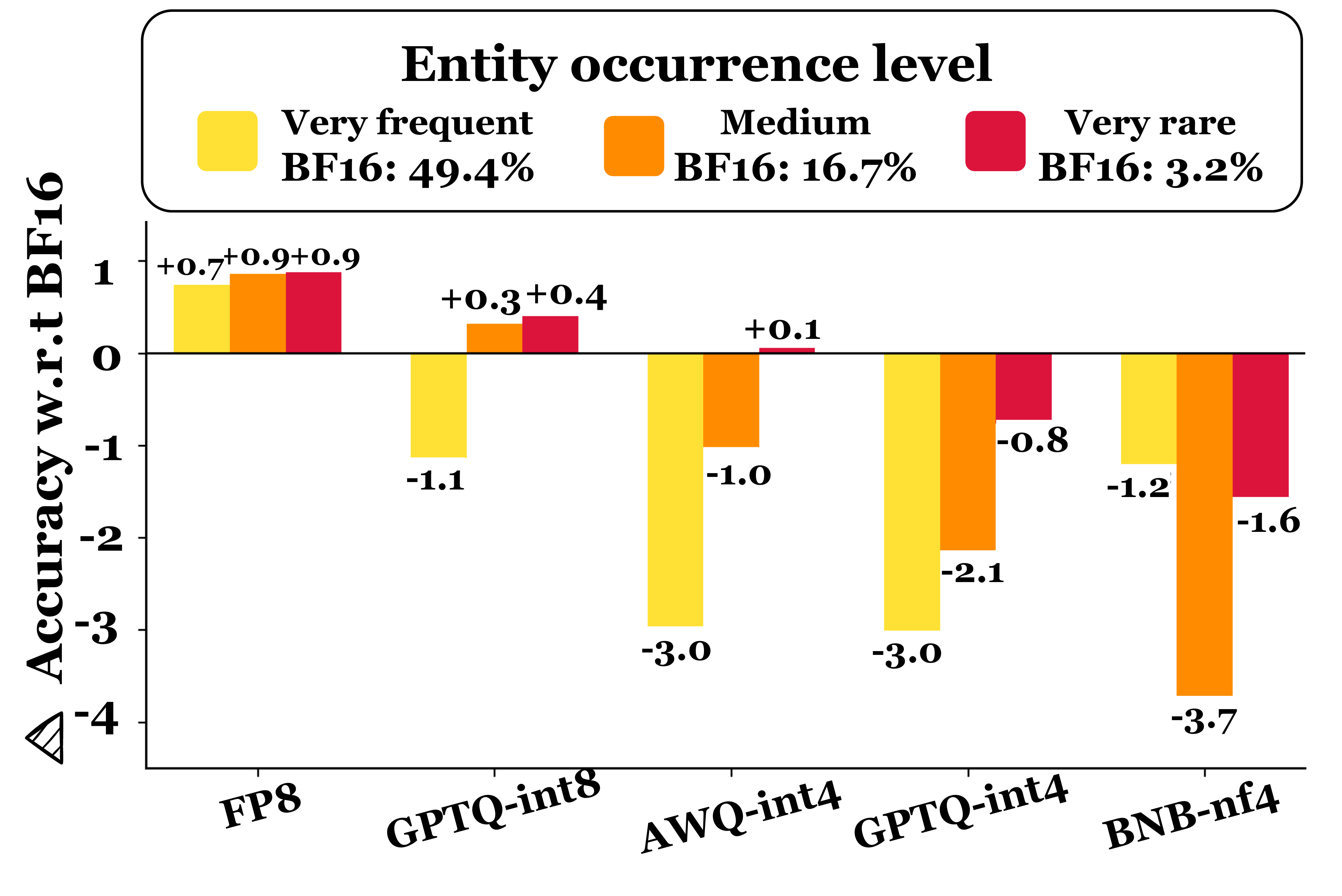}
    \caption{\textbf{\factscore:} Average $\Delta$-\veriscore{} (relative to BF16), varying by level of occurrence of the input entity}
    \label{fig:veriscore_raritywise}
\end{figure}

\paragraph{Factuality is preserved at 8-bit but drops at 4-bit precision.}
On the \factscore{} biography generation task, 8-bit quantizations remain essentially on par with the BF16 baseline (FP8 +0.8\%, GPTQ-int8 –0.1\%; see \autoref{fig:all_taskwise}).  
In contrast, 4-bit precision reduces scores: AWQ-int4 (–1.3\%), GPTQ-int4 (–1.9\%), BNB-nf4 (–2.1\%), with model-specific variation (\Cref{fig:veriscore_modelwise}).\footnote{Most of this decline stems from higher abstention rates: responses with no verifiable claim are counted as errors. For example, \llama{}-70B abstains in 29\% of BF16 outputs, 30\% of FP8, and 52\% of BNB-nf4. Full statistics appear in \autoref{fig:veriscore-noclaims-heatmap} (\S\ref{appsec:factscore}).}
\autoref{fig:veriscore_raritywise} shows that the largest drops (up to –3.7\%) occur for very frequent entities, while medium- and low-frequency entities are less affected. These smaller declines likely reflect their already low baseline accuracy, which leaves limited room for further degradation. All figures show that drops are small, with only BNB-nf4 showing statistically significant  effects on \veriscore{}, no claim rates and effect of entity frequency ($p$<0.05, Tables \ref{tab:veriscore-quant-posthoc}, \ref{tab:veriscore-empty_claims-posthoc} \& \ref{tab:veriscore-category-posthoc} in \S\ref{app:stats-test}).

\begin{figure}[t] 
    \centering
    \includegraphics[width=\linewidth]{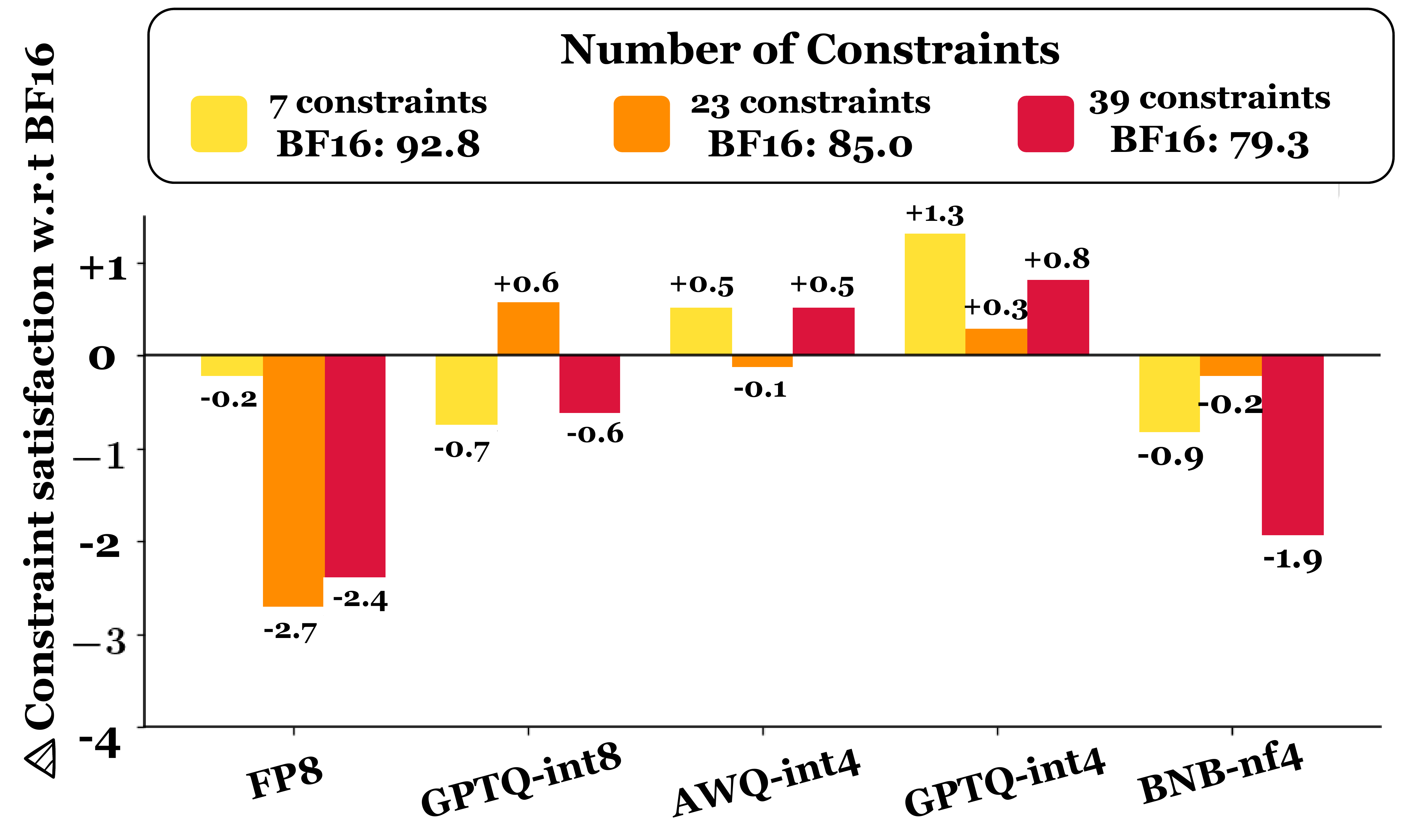}
    %\vspace{-2pt}
    \caption{\textbf{\csfour:} Average $\Delta$-constraint-satisfaction of different quantization methods relative to BF16 on the \csfour~benchmark across varying input constraints.}
    \label{fig:cs-barplot}
\end{figure}

%\begin{figure}[t] 
%    \centering
%    \includegraphics[width=\linewidth]{images/BooookScore_Model.pdf}
    %\vspace{-2pt}
%    \caption{Average BooookScore win rates of different quantization methods against BF16 on \csfour.}
%    \label{fig:BooookScore-heatmap}
%\end{figure}

\paragraph{GPTQ-int8, GPTQ-int4, and AWQ-int4 preserve constraint satisfaction on \csfour{}, while FP8 and BNB-nf4 incur losses.}
% \autoref{fig:all_taskwise} plots the $\Delta$-constraint-satisfaction for each quantization on \csfour{}. AWQ-int4 and GPTQ-int4 slightly outperform BF16 (+0.3\% and +0.8\%, respectively), and GPTQ-int8 drops only –0.3\%. In contrast, FP8 and BNB-nf4 suffer significant drops of –1.8\% and –1.0\% (\textit{p}<0.05; \Cref{tab:csfour-stats-model-quant,tab:cs4-posthoc-model} in \S\ref{app:stats-test})). Breaking down by constraint count (\autoref{fig:cs-barplot}), FP8 and BNB-nf4 show larger declines at higher constraint levels (\textit{p}<0.05; \Cref{tab:cs4-interact-posthoc} in \S\ref{app:stats-test}). The pronounced FP8 drop on instruction following is unexpected given its robustness on other tasks.
On the \csfour{} task, most quantization methods preserve constraint satisfaction. GPTQ and AWQ methods perform on par with the BF16 baseline, with scores ranging from +0.8\% to -0.3\% (\autoref{fig:all_taskwise}). In contrast, FP8 and BNB-nf4 incur statistically significant drops of -1.8\% and -1.0\%, respectively (\textit{p}<0.05; \Cref{tab:csfour-stats-model-quant,tab:cs4-posthoc-model} in \S\ref{app:stats-test}). For these two methods, the degradation worsens as the number of constraints increases (\autoref{fig:cs-barplot}, \textit{p}<0.05; \Cref{tab:cs4-interact-posthoc} in \S\ref{app:stats-test}). The pronounced drop from FP8 is particularly noteworthy, as it contrasts with its strong performance on other tasks.

 % Splitting by the number of constraints (\autoref{fig:cs-barplot}) reveals that at seven constraints, AWQ-int4 and GPTQ-int4 improve by 0.5\% and 1.1\%, whereas FP8, GPTQ-int8, and BNB-nf4 drop by 0.2\%, 0.7\%, and 0.9\%, respectively. At twenty-three constraints, GPTQ-int8 and GPTQ-int4 achieve improvements of 0.6\% and 0.3\%, respectively; BNB-nf4 and AWQ-int4 remain within 0.2\% of the baseline, while FP8 declines by 2.7\%. Under thirty-nine constraints, GPTQ-int4 and AWQ-int4 deliver improvements of 0.8\% and 0.5\%, respectively, while FP8 drops by 2.4\%, GPTQ-int8 by 0.6\%, and BNB-nf4 by 1.9\%. Statistically\AAaddressed{add table}, the performance drops of FP8 and BNB-nf4 are significant, with \textit{p}-values less than 0.01. While the instruction-following results of FP8 are surprising, BNB-nf4’s performance aligns with its degradation trends observed in other tasks. 

\paragraph{Quantization degrades story coherence, especially in smaller models.} 
Story coherence, evaluated by BooookScore, declines with quantization for all models except \qwen{} 72B (\autoref{fig:BooookScore_modelwise}). Within each model family, the drop is generally more pronounced for smaller models. For instance, in the \qwen{} family, the 7B model loses 6.8\% on average, while the 32B model loses only 2.0\% and the 72B model gains 0.7\%. This pattern is also visible in the \llama{} family: the 70B model sees a modest 4.5\% average drop, while the 8B model shows a much larger 10\% loss. The notable exception is FP8 quantization, which preserves coherence exceptionally well for \llama{} 8B, resulting in a mere 0.5\% decline.

\paragraph{FP8 and GPTQ-int8 preserve overall accuracy most effectively, while 4-bit quantization causes noticeable degradation.}
FP8 and GPTQ-int8 have the smallest average accuracy drops ($\Delta$-accuracy 0.8\% and 0.2\%, respectively) relative to BF16 across tasks (\autoref{fig:overall-quant}). Post‑hoc tests for \ruler{} (\Cref{tab:ruler-posthoc-model}), \oneruler{} (\autoref{tab:oneruler-posthoc-model}), \nocha{} (\Cref{tab:nocha-stats}), and \factscore{} (\autoref{tab:oneruler-posthoc-model}) confirm that neither differs significantly from BF16 ($p$>0.05).
% When broken down by model (Figures \ref{fig:longctx-modelwise}, \ref{fig:longgen-modelwise}), only 4 of 50 8‑bit model–quantization combinations show large drops: GPTQ‑int8 on \llama{} 70B for \nocha{} and \csfour{}, and FP8 on \qwen{} 32B for \ruler{} and \llama{} 8B for \csfour{}. Averaged across models, even the worst case 8-bit performance is just -2.7\% for FP8 on \ruler{} (\autoref{fig:all_taskwise}). This suggests that 8-bit quantization methods can be deployed with minimal impact.
In contrast to 8-bit quantization methods, 4-bit methods exhibit larger losses ($p$<0.05 in \Cref{tab:ruler-posthoc-model,tab:oneruler-posthoc-model,tab:nocha-posthoc-model,tab:cs4-posthoc-model,tab:veriscore-quant-posthoc} in \S\ref{app:stats-test}). AWQ-int4 and GPTQ-int4 suffer average drops of 1-3\%, while BNB-nf4 loses an average of 6.9\% (\autoref{fig:overall-quant}).  Specific cases show very large drops: on \oneruler{}, \llama{} 70B under BNB-nf4 drops by 59\% (\autoref{fig:oneruler_modelwise}); on \ruler{}, \qwen{} 8B, \qwen{} 32B, and \llama{} 8B each drop by over 15\% (\autoref{fig:ruler-model-wise}), etc.
% ; and on \nocha{}, \qwen{} 32B and \llama{} 70B each lose over 7\% (\autoref{fig:nocha_modelwise}). 
Notably, BNB-nf4 is the default 4-bit quantization in libraries like HuggingFace and vLLM, so its pronounced performance drops have significant practical implications. The results highlight that 4-bit quantization, especially BNB-bf4, should be used with caution. Overall, as seen in \autoref{fig:overall-quant}, the average trend across quantization methods indicates that FP8 and GPTQ-int8 are comparable with BF16, with the 4-bit quantization methods ranked as AWQ-int4 > GPTQ-int4 > BNB-nf4. 
% This aligns with observations by \citet{kurtic2024give,lee2024comprehensive}.

\paragraph{The effect of quantization varies with model and task.}
Our results demonstrate that quantization effects are highly context-dependent, varying significantly with both model family and the evaluation task. This model-specific behavior is evident even between models of comparable size. On the \oneruler{} benchmark, for example, BNB-nf4 quantization leads to a 59\% accuracy drop for \llama{} 70B, whereas the similarly-sized \qwen{} 72B gains 0.6\% with the same method (\autoref{fig:oneruler_modelwise}). 
% Similarly, models of comparable size but different families respond differently. \autoref{fig:oneruler_modelwise} shows that on \oneruler{}, BNB-nf4 leads to a 59\% accuracy drop for \llama{} 70B, while \qwen{} 72B sees a 0.6\% gain with the same method.
% Similarly, the impact of quantization changes with the task. For instance, \llama{} 70B with GPTQ-int8 drops 11.1\% on \nocha{} (\autoref{fig:nocha_modelwise}), but its performance on \ruler{} (\autoref{fig:ruler-model-wise}) and \oneruler{} (\autoref{fig:oneruler_modelwise}) does not decline similarly. The impact of quantization is not uniform across model families as well. In  \autoref{fig:all_modelwise}, \llama{} models are more affected by quantization than \qwen{}. While \qwen{} 7B also sees a notable performance decrease, its 32B and 72B versions are largely resilient, suggesting larger \qwen{} models handle quantization better. Despite similar model sizes, \autoref{fig:oneruler_modelwise} shows that on \oneruler{}, BNB-nf4 leads to a 59\% accuracy drop for \llama{} 70B, while \qwen{} 72B sees a 0.6\% gain with the same method.
% This pair of models exhibits starkly different responses to quantization, on \oneruler{} across multiple context lengths (\autoref{fig:oneruler-barplots-llama-qwen}). % need to decide on inclusion
% These differences contrast with previous studies focusing on a single model family that reported minimal performance degradation \citep{kurtic2024give,huang2024empirical,kumar2025scaling,jin2024comprehensive}, highlighting the need for evaluation across a diverse set of models. 
Sensitivity also varies within a model family; the \llama{} models are generally more affected by quantization than the \qwen{} models, whose larger 32B and 72B variants show higher resilience.\footnote{For AWQ and GPTQ, the \llama{} and \qwen{} models originate from different sources and may have been quantized with different calibration sets. The same overall patterns, however, also appear in BNB-fn4 and FP8 quantizations.} This effect can even be unpredictable, as seen when BNB-nf4 causes a 32\% performance drop on \llama{} 70B but a 0.6\% gain on \llama{} 8B (\autoref{fig:all_modelwise}).

Furthermore, a model's robustness to quantization is also task-dependent. For instance, that same \llama{} 70B model, when quantized with GPTQ-int8, drops 11.1\% on \nocha{} (\autoref{fig:nocha_modelwise}) but shows no comparable decline on \ruler{} or \oneruler{} (\autoref{fig:ruler-model-wise}, \autoref{fig:oneruler_modelwise}). 
These findings differ from prior work that examined a single model family and observed only minor performance drops \citep{kurtic2024give,huang2024empirical,kumar2025scaling,jin2024comprehensive}, highlighting the importance of evaluating a diverse set of models. 

\paragraph{Why does quantization affect long-context tasks?}
While studies on short context tasks show minimal degradation, we observe larger drops, particularly for tasks involving long-context inputs. A possible explanation for this sink could be the accumulation of errors in RoPE embeddings \citep{wang2025when}. This issue could be potentially addressed by using long-context samples during the calibration stage of quantization \citep{williams2023impact}, including multilingual calibration data, and applying attention-calibration techniques like found-in-the-middle to reduce attention errors \citep{hsieh2024found}. We leave the evaluation of these strategies for future work.

% \AAaddressed{
% \paragraph{Underlying causes and potential solutions to tackle performance drops caused by quantization.} As noted by \citet{wang2025when}, a possible factor contributing to the degraded performance of quantized models in long-context tasks is the accumulation of errors in RoPE embeddings. We hypothesize that such issues could be mitigated by incorporating long-context in-distribution samples during the calibration stage of post-training quantization \citep{williams2023impact}, using multilingual calibration data, and applying attention-calibration techniques such as found-in-the-middle \citep{hsieh2024found} to reduce attention-sink errors and improve long-context performance. However, evaluating these strategies would require substantial model-level modifications and re-evaluation across benchmarks, which we leave for future work.
% }

\section{Related works}

\noindent\textbf{Prior studies on quantization effects:} Recent work shows that 8‑bit quantization preserves performance, while 4‑bit methods incur only marginal accuracy loss \citep{kurtic2024give,badshah2024quantifying,liu2024quantizegeneralization,jin2024comprehensive,huang2024empirical,xu2024beyond,li2024evaluating,del2025aggregating,zheng2025empirical}. \citet{lee2024comprehensive} and \citet{kurtic2024give} investigate \llama{} and find that FP8 is nearly lossless, while AWQ-int4 and GPTQ-int4 exhibit small degradations, particularly in reasoning, coding tasks and on Arena-Hard. Results on Mistral \citep{badshah2024quantifying}, Baichuan \citep{liu2024quantizegeneralization}, and Qwen-Chat \citep{jin2024comprehensive} are also similar, with losses rarely exceeding 3\% and diminishing as model size grows. Results on safety, alignment, and interpretability suggest that 4-bit quantization may \citep{egashira2024exploiting} or may not \citep{xu2024beyond} lead to drops in safety, protect against data-poisoning \citep{hussain2025capturing}, reverse unlearning \citep{zhang2025catastrophic}, and have mixed results on interpretability \citep{wang2025through}. Some studies have also shown that models are robust to quantization even at 3-bit precision or lower \citep{chee2023quip,ma2024era,egiazarian2024extreme,kumar2025scaling}. However, these evaluations are generally limited in terms of the input context or the output token ranges, leaving longer-range effects unknown. A significant body of work has studied accuracy trade-offs using different quantization methods on academic benchmarks \citep{yao2023zeroquant,liu2024emergent,gong2024llmc,gong2024makes}. However, these benchmarks may not fully reflect performance in realistic scenarios, such as those involving long-form reasoning or multilingual settings. For instance, \citet{li2025quantization} and \citet{liu2025quantization} report significant drops for GPTQ-int4 and AWQ-int4 on such tasks. For multilingual LLMs, \citet{marchisio2024does} observe increasing drops on reasoning for non-Latin script languages upon quantization. We develop this picture further by evaluating long-form generation (\csfour{}, \factscore{}) and across languages in \oneruler{}.

\noindent\textbf{Long‑context focus:} To our knowledge, only \citet{kurtic2024give} evaluate quantization on long‑context tasks, and they report coarse averages over \ruler{} (which includes context lengths as low as 8K) to conclude negligible performance loss. Our analysis differs by focusing on the effects of parameters like context length, task-type, especially at context lengths >64K. While they evaluate only the \llama{} family and use a single 4-bit quantization (GPTQ-int4), we extend the scale of long context analysis by including \oneruler{} and \nocha{}, the \qwen{} model family, and the AWQ-int4 and BNB-nf4 quantizations.  We find that \qwen{} and \llama{} respond to quantization very differently, with \llama{} showing larger drops upon quantization (\autoref{fig:ruler-model-wise}). In addition, the AWQ-int4 quantization performs better than GPTQ-int4, while the popular BNB-nf4 quantization performs worse (\autoref{fig:all_taskwise}).

\section{Conclusion}

In this paper, we presented a systematic evaluation of the performance of five open-weight models (7B to 72B parameters) across five quantization methods (4-bit and 8-bit). We conducted experiments on five tasks requiring the models to either process long-context inputs (in English or other languages) or return long-form outputs in English. While prior work often suggests 8-bit quantization is nearly lossless and 4-bit methods are \textit{good enough}, our findings demonstrate that the impact of quantization becomes more pronounced in long-context scenarios. In particular, we observed that model's performance under 4-bit quantization tends to degrade progressively as input context length increases. 4-bit quantization has a significant impact when the input is in a language other than English. In contrast, 8-bit precision methods largely preserve accuracy across context lengths. Furthermore, we observed considerable heterogeneity in quantization effects---the performance impact of reduced precision varies significantly across different model architectures and task types. Taken together, these results indicate that quantization should be applied with careful consideration in long-context applications and in multilingual setups.

% In summary, we conducted a comprehensive evaluation of quantized LLMs across five open-source models (7B to 72B parameters) using five quantization methods (with 8- and 4-bit precisions). The five tasks evaluated span retrieval, multilingual ability, and reasoning, requiring long contexts and long-form outputs. While prior works generally claim that 8-bit is cost-free and 4-bit \textit{good enough}, our results reveal that quantization’s impact is far more pronounced in long-context scenarios. In particular, 4-bit quantization incurs substantial performance degradation in high-context tasks, with the popular BNB-nf4 scheme showing especially large accuracy drops. 8-bit-precision schemes largely preserve accuracy. We also observed that the effect of quantization is highly heterogeneous across different models and task types, indicating that each model–task combination responds differently to reduced precision. 

% These findings indicate that quantization should be applied with care in long-context applications. Conclusions drawn from short-context evaluations do not fully hold when sequence lengths grow. Our study provides insights to guide deployment of LLMs under cost constraints, by revealing areas where cost-performance tradeoffs are most significant. 
% --- Additional Essential Sections ---

% Place your limitations here. This section is required for ACL Rolling Review.

\section*{Limitations}

While we conduct a large-scale evaluation of quantized LLMs on long-context and long-form tasks, we identify the following limitations and hope future work will improve on these aspects to enable more comprehensive evaluations of quantization methods.
\paragraph{Scope of experiments:} Our evaluations cover five quantization methods and five LLMs, with the largest model having 72 billion parameters. We largely followed publicly available, off-the-shelf quantized models and recipes (e.g., from the official Qwen repository, HuggingQuants, and NeuralMagic) and did not explore tuning each quantization method further. As a result, our findings might not fully generalize to other models or to quantization techniques beyond those tested.

% \newpage
\paragraph{Evaluation methodology and limited output lengths:} Although we tested input contexts as long as 128K tokens, we restricted the generated outputs in our experiments to relatively short passages (< 650 tokens). We made this choice because our evaluation relies solely on automated metrics and LLM-based judge models, which are not reliable for evaluating very long text. Consequently, without human annotations, our assessment may overlook subtle errors in logical flow or nuanced instruction-following behavior over long-form outputs.

\section*{Ethical Considerations}
We rely solely on publicly available datasets that have been vetted and validated in prior peer-reviewed studies. We do not introduce any new data collection or human subject research procedures. Our paper poses no ethical concerns. 
\section*{Acknowledgments}
This project was partially supported by awards IIS-2046248, IIS-2312949, and IIS-2202506 from the National Science Foundation (NSF). We also thank the UMass NLP community for their insightful comments.

AI assistants like ChatGPT were used to polish writing and debug code for experiments. They were not used to generate paper content or result tables and plots. 

% --- End Additional Sections ---

\bibliography{custom}

\begin{thebibliography}{65}
\providecommand{\natexlab}[1]{#1}

\bibitem[{Atmakuru et~al.(2024)Atmakuru, Nainani, Bheemreddy, Lakkaraju, Yao, Zamani, and Chang}]{atmakuru2024cs4}
Anirudh Atmakuru, Jatin Nainani, Rohith Siddhartha~Reddy Bheemreddy, Anirudh Lakkaraju, Zonghai Yao, Hamed Zamani, and Haw-Shiuan Chang. 2024.
\newblock {CS}4: Measuring the creativity of large language models automatically by controlling the number of story-writing constraints.
\newblock In \emph{6th Workshop on Narrative Understanding (WNU)}.

\bibitem[{Badshah and Sajjad(2024)}]{badshah2024quantifying}
Sher Badshah and Hassan Sajjad. 2024.
\newblock \href {https://arxiv.org/abs/2405.03146} {Quantifying the capabilities of llms across scale and precision}.
\newblock \emph{arXiv preprint arXiv:2405.03146}.

\bibitem[{Bates et~al.(2015)Bates, M{\"a}chler, Bolker, and Walker}]{bates2015fitting}
Douglas Bates, Martin M{\"a}chler, Ben Bolker, and Steve Walker. 2015.
\newblock Fitting linear mixed-effects models using lme4.
\newblock \emph{Journal of statistical software}, 67:1--48.

\bibitem[{Chang et~al.(2024)Chang, Lo, Goyal, and Iyyer}]{chang2024booookscore}
Yapei Chang, Kyle Lo, Tanya Goyal, and Mohit Iyyer. 2024.
\newblock \href {https://openreview.net/forum?id=7Ttk3RzDeu} {{BooookScore}: A systematic exploration of book-length summarization in the era of {LLM}s}.
\newblock In \emph{The Twelfth International Conference on Learning Representations}.

\bibitem[{Chee et~al.(2023)Chee, Cai, Kuleshov, and De~Sa}]{chee2023quip}
Jerry Chee, Yaohui Cai, Volodymyr Kuleshov, and Christopher~M De~Sa. 2023.
\newblock {QuIP}: 2-bit quantization of large language models with guarantees.
\newblock \emph{Advances in Neural Information Processing Systems}, 36:4396--4429.

\bibitem[{{Databricks}(2023)}]{databricks2023llminference}
{Databricks}. 2023.
\newblock \href {https://www.databricks.com/blog/llm-inference-performance-engineering-best-practices} {{LLM} inference performance engineering: Best practices}.
\newblock Accessed: May 18, 2025.

\bibitem[{del Rey et~al.(2025)del Rey, Santos, Travassos, Franch, and Mart{\'\i}nez-Fern{\'a}ndez}]{del2025aggregating}
Santiago del Rey, Paulo S{\'e}rgio Medeiros~dos Santos, Guilherme~Horta Travassos, Xavier Franch, and Silverio Mart{\'\i}nez-Fern{\'a}ndez. 2025.
\newblock Aggregating empirical evidence from data strategy studies: a case on model quantization.
\newblock \emph{arXiv preprint arXiv:2505.00816}.

\bibitem[{Dettmers et~al.(2022)Dettmers, Lewis, Belkada, and Zettlemoyer}]{dettmers2022gpt3}
Tim Dettmers, Mike Lewis, Younes Belkada, and Luke Zettlemoyer. 2022.
\newblock \href {https://proceedings.neurips.cc/paper_files/paper/2022/file/c3ba4962c05c49636d4c6206a97e9c8a-Paper-Conference.pdf} {{GPT3.int8()}: 8-bit matrix multiplication for transformers at scale}.
\newblock \emph{Advances in Neural Information Processing Systems}, 35:30318--30332.

\bibitem[{Dettmers et~al.(2023)Dettmers, Pagnoni, Holtzman, and Zettlemoyer}]{dettmers_qlora_bnbint4}
Tim Dettmers, Artidoro Pagnoni, Ari Holtzman, and Luke Zettlemoyer. 2023.
\newblock \href {https://proceedings.neurips.cc/paper_files/paper/2023/file/1feb87871436031bdc0f2beaa62a049b-Paper-Conference.pdf} {{QLoRA}: Efficient finetuning of quantized {LLMs}}.
\newblock In \emph{Advances in Neural Information Processing Systems}, volume~36, pages 10088--10115. Curran Associates, Inc.

\bibitem[{Egashira et~al.(2024)Egashira, Vero, Staab, He, and Vechev}]{egashira2024exploiting}
Kazuki Egashira, Mark Vero, Robin Staab, Jingxuan He, and Martin Vechev. 2024.
\newblock \href {https://openreview.net/pdf?id=d8CAXiITZ3} {Exploiting {LLM} quantization}.
\newblock In \emph{ICML 2024 Next Generation of AI Safety Workshop}.

\bibitem[{Egiazarian et~al.(2024)Egiazarian, Panferov, Kuznedelev, Frantar, Babenko, and Alistarh}]{egiazarian2024extreme}
Vage Egiazarian, Andrei Panferov, Denis Kuznedelev, Elias Frantar, Artem Babenko, and Dan Alistarh. 2024.
\newblock Extreme compression of large language models via additive quantization.
\newblock In \emph{Proceedings of the 41st International Conference on Machine Learning}, ICML'24. JMLR.org.

\bibitem[{Fernandez et~al.(2025)Fernandez, Na, Tiwari, Bisk, Luccioni, and Strubell}]{fernandez2025energy}
Jared Fernandez, Clara Na, Vashisth Tiwari, Yonatan Bisk, Sasha Luccioni, and Emma Strubell. 2025.
\newblock \href {https://openreview.net/forum?id=aXNty1YGe0} {Energy considerations of large language model inference and efficiency optimizations}.
\newblock In \emph{LTI Student Research Symposium 2025}.

\bibitem[{Frantar et~al.(2023)Frantar, Ashkboos, Hoefler, and Alistarh}]{frantar2023gptq}
Elias Frantar, Saleh Ashkboos, Torsten Hoefler, and Dan Alistarh. 2023.
\newblock \href {https://openreview.net/pdf?id=tcbBPnfwxS} {{GPTQ}: Accurate post-training quantization for generative pre-trained transformers}.
\newblock In \emph{The Eleventh International Conference on Learning Representations}.

\bibitem[{Gong et~al.(2024{\natexlab{a}})Gong, Yong, Gu, Huang, Lv, Zhang, Tao, and Liu}]{gong2024llmc}
Ruihao Gong, Yang Yong, Shiqiao Gu, Yushi Huang, Chengtao Lv, Yunchen Zhang, Dacheng Tao, and Xianglong Liu. 2024{\natexlab{a}}.
\newblock \href {https://doi.org/10.18653/v1/2024.emnlp-industry.12} {{LLMC}: Benchmarking large language model quantization with a versatile compression toolkit}.
\newblock In \emph{Proceedings of the 2024 Conference on Empirical Methods in Natural Language Processing: Industry Track}, pages 132--152, Miami, Florida, US. Association for Computational Linguistics.

\bibitem[{Gong et~al.(2024{\natexlab{b}})Gong, Liu, Wang, Cai, Zhao, and Yan}]{gong2024makes}
Zhuocheng Gong, Jiahao Liu, Jingang Wang, Xunliang Cai, Dongyan Zhao, and Rui Yan. 2024{\natexlab{b}}.
\newblock What makes quantization for large language model hard? {An} empirical study from the lens of perturbation.
\newblock In \emph{Proceedings of the AAAI Conference on Artificial Intelligence}, volume~38, pages 18082--18089.

\bibitem[{{Google Cloud}(2024)}]{google2024jetstream}
{Google Cloud}. 2024.
\newblock \href {https://cloud.google.com/blog/products/compute/accelerating-ai-inference-with-google-cloud-tpus-and-gpus} {{Accelerating AI inference with Google Cloud TPUs and GPUs}}.
\newblock Accessed: May 18, 2025.

\bibitem[{Grattafiori et~al.(2024)Grattafiori, Dubey, Jauhri, Pandey, Kadian, Al-Dahle, Letman, Mathur, Schelten, Vaughan et~al.}]{grattafiori2024llama}
Aaron Grattafiori, Abhimanyu Dubey, Abhinav Jauhri, Abhinav Pandey, Abhishek Kadian, Ahmad Al-Dahle, Aiesha Letman, Akhil Mathur, Alan Schelten, Alex Vaughan, et~al. 2024.
\newblock \href {https://arxiv.org/abs/2407.21783} {The {Llama 3} herd of models}.
\newblock \emph{arXiv preprint arXiv:2407.21783}.

\bibitem[{Hsieh et~al.(2024{\natexlab{a}})Hsieh, Sun, Kriman, Acharya, Rekesh, Jia, and Ginsburg}]{hsiehruler}
Cheng-Ping Hsieh, Simeng Sun, Samuel Kriman, Shantanu Acharya, Dima Rekesh, Fei Jia, and Boris Ginsburg. 2024{\natexlab{a}}.
\newblock \href {https://openreview.net/pdf?id=kIoBbc76Sy} {{RULER}: What's the real context size of your long-context language models?}
\newblock In \emph{First Conference on Language Modeling}.

\bibitem[{Hsieh et~al.(2024{\natexlab{b}})Hsieh, Chuang, Li, Wang, Le, Kumar, Glass, Ratner, Lee, Krishna et~al.}]{hsieh2024found}
Cheng-Yu Hsieh, Yung-Sung Chuang, Chun-Liang Li, Zifeng Wang, Long~T Le, Abhishek Kumar, James Glass, Alexander Ratner, Chen-Yu Lee, Ranjay Krishna, et~al. 2024{\natexlab{b}}.
\newblock Found in the middle: Calibrating positional attention bias improves long context utilization.
\newblock \emph{arXiv preprint arXiv:2406.16008}.

\bibitem[{Huang et~al.(2024)Huang, Zheng, Ma, Qin, Lv, Chen, Luo, Qi, Liu, and Magno}]{huang2024empirical}
Wei Huang, Xingyu Zheng, Xudong Ma, Haotong Qin, Chengtao Lv, Hong Chen, Jie Luo, Xiaojuan Qi, Xianglong Liu, and Michele Magno. 2024.
\newblock \href {https://arxiv.org/abs/2404.14047} {An empirical study of {LLaMA3} quantization: From {LLMs} to {MLLMs}}.
\newblock \emph{Visual Intelligence}, 2(1):36.

\bibitem[{Hussain et~al.(2025)Hussain, Zarkouei, Rabin, Alipour, Lin, and Xu}]{hussain2025capturing}
Aftab Hussain, Sadegh AlMahdi~Kazemi Zarkouei, Md~Rafiqul~Islam Rabin, Mohammad~Amin Alipour, Sen Lin, and Bowen Xu. 2025.
\newblock Capturing the effects of quantization on trojans in code {LLMs}.
\newblock \emph{arXiv preprint arXiv:2505.14200}.

\bibitem[{Jin et~al.(2024)Jin, Du, Huang, Liu, Luan, Wang, and Xiong}]{jin2024comprehensive}
Renren Jin, Jiangcun Du, Wuwei Huang, Wei Liu, Jian Luan, Bin Wang, and Deyi Xiong. 2024.
\newblock \href {https://aclanthology.org/2024.findings-acl.726/} {A comprehensive evaluation of quantization strategies for large language models}.
\newblock In \emph{Findings of the Association for Computational Linguistics: ACL 2024}, pages 12186--12215.

\bibitem[{Karpinska et~al.(2024)Karpinska, Thai, Lo, Goyal, and Iyyer}]{karpinska2024one}
Marzena Karpinska, Katherine Thai, Kyle Lo, Tanya Goyal, and Mohit Iyyer. 2024.
\newblock \href {https://arxiv.org/abs/2406.16264} {One thousand and one pairs: A “novel” challenge for long-context language models}.
\newblock In \emph{Proceedings of the 2024 Conference on Empirical Methods in Natural Language Processing}, pages 17048--17085.

\bibitem[{Kim et~al.(2023)Kim, Lee, Kim, Park, Yoo, Kwon, and Lee}]{kim2023memory}
Jeonghoon Kim, Jung~Hyun Lee, Sungdong Kim, Joonsuk Park, Kang~Min Yoo, Se~Jung Kwon, and Dongsoo Lee. 2023.
\newblock Memory-efficient fine-tuning of compressed large language models via sub-4-bit integer quantization.
\newblock \emph{Advances in Neural Information Processing Systems}, 36:36187--36207.

\bibitem[{Kim et~al.(2025)Kim, Russell, Karpinska, and Iyyer}]{kim2025one}
Yekyung Kim, Jenna Russell, Marzena Karpinska, and Mohit Iyyer. 2025.
\newblock \href {https://arxiv.org/abs/2503.01996} {One ruler to measure them all: Benchmarking multilingual long-context language models}.
\newblock \emph{arXiv preprint arXiv:2503.01996}.

\bibitem[{Kumar et~al.(2025)Kumar, Ankner, Spector, Bordelon, Muennighoff, Paul, Pehlevan, R{\'e}, and Raghunathan}]{kumar2025scaling}
Tanishq Kumar, Zachary Ankner, Benjamin~Frederick Spector, Blake Bordelon, Niklas Muennighoff, Mansheej Paul, Cengiz Pehlevan, Christopher R{\'e}, and Aditi Raghunathan. 2025.
\newblock \href {https://openreview.net/pdf?id=wg1PCg3CUP} {Scaling laws for precision}.
\newblock In \emph{The Thirteenth International Conference on Learning Representations}.

\bibitem[{Kurtic et~al.(2024)Kurtic, Marques, Pandit, Kurtz, and Alistarh}]{kurtic2024give}
Eldar Kurtic, Alexandre Marques, Shubhra Pandit, Mark Kurtz, and Dan Alistarh. 2024.
\newblock \href {https://arxiv.org/abs/2411.02355} {``{G}ive me {BF16} or give me death”? {A}ccuracy-performance trade-offs in llm quantization}.
\newblock \emph{arXiv preprint arXiv:2411.02355}.

\bibitem[{Kwon et~al.(2023)Kwon, Li, Zhuang, Sheng, Zheng, Yu, Gonzalez, Zhang, and Stoica}]{kwon2023efficientvllm}
Woosuk Kwon, Zhuohan Li, Siyuan Zhuang, Ying Sheng, Lianmin Zheng, Cody~Hao Yu, Joseph~E. Gonzalez, Hao Zhang, and Ion Stoica. 2023.
\newblock \href {https://dl.acm.org/doi/10.1145/3600006.3613165} {Efficient memory management for large language model serving with {PagedAttention}}.
\newblock In \emph{Proceedings of the ACM SIGOPS 29th Symposium on Operating Systems Principles}.

\bibitem[{Lee et~al.(2024)Lee, Park, Kwon, Oh, and Kwon}]{lee2024comprehensive}
Jemin Lee, Sihyeong Park, Jinse Kwon, Jihun Oh, and Yongin Kwon. 2024.
\newblock \href {https://arxiv.org/abs/2409.11055} {A comprehensive evaluation of quantized instruction-tuned large language models: An experimental analysis up to 405b}.
\newblock \emph{arXiv preprint arXiv:2409.11055}.

\bibitem[{Li et~al.(2024)Li, Ning, Wang, Liu, Shi, Yan, Dai, Yang, and Wang}]{li2024evaluating}
Shiyao Li, Xuefei Ning, Luning Wang, Tengxuan Liu, Xiangsheng Shi, Shengen Yan, Guohao Dai, Huazhong Yang, and Yu~Wang. 2024.
\newblock \href {https://arxiv.org/abs/2402.18158} {Evaluating quantized large language models}.
\newblock In \emph{Proceedings of the 41st International Conference on Machine Learning}, pages 28480--28524.

\bibitem[{Li et~al.(2025)Li, Su, Yang, Xie, Wong, and Yang}]{li2025quantization}
Zhen Li, Yupeng Su, Runming Yang, Zhongwei Xie, Ngai Wong, and Hongxia Yang. 2025.
\newblock \href {https://arxiv.org/abs/2501.03035} {Quantization meets reasoning: Exploring {LLM} low-bit quantization degradation for mathematical reasoning}.
\newblock \emph{arXiv preprint arXiv:2501.03035}.

\bibitem[{Lin et~al.(2024)Lin, Tang, Tang, Yang, Chen, Wang, Xiao, Dang, Gan, and Han}]{lin2024awq}
Ji~Lin, Jiaming Tang, Haotian Tang, Shang Yang, Wei-Ming Chen, Wei-Chen Wang, Guangxuan Xiao, Xingyu Dang, Chuang Gan, and Song Han. 2024.
\newblock \href {https://proceedings.mlsys.org/paper_files/paper/2024/file/42a452cbafa9dd64e9ba4aa95cc1ef21-Paper-Conference.pdf} {{AWQ}: Activation-aware weight quantization for on-device llm compression and acceleration}.
\newblock \emph{Proceedings of Machine Learning and Systems}, 6:87--100.

\bibitem[{Liu et~al.(2024{\natexlab{a}})Liu, Liu, Gao, Gao, Zhao, Li, Ding, and Wen}]{liu2024emergent}
Peiyu Liu, Zikang Liu, Ze-Feng Gao, Dawei Gao, Wayne~Xin Zhao, Yaliang Li, Bolin Ding, and Ji-Rong Wen. 2024{\natexlab{a}}.
\newblock \href {https://aclanthology.org/2024.lrec-main.461/} {Do emergent abilities exist in quantized large language models: An empirical study}.
\newblock In \emph{Proceedings of the 2024 Joint International Conference on Computational Linguistics, Language Resources and Evaluation (LREC-COLING 2024)}, pages 5174--5190, Torino, Italia. ELRA and ICCL.

\bibitem[{Liu et~al.(2025)Liu, Sun, Zhang, Bai, Yu, Yu, Yuan, and Hou}]{liu2025quantization}
Ruikang Liu, Yuxuan Sun, Manyi Zhang, Haoli Bai, Xianzhi Yu, Tiezheng Yu, Chun Yuan, and Lu~Hou. 2025.
\newblock Quantization hurts reasoning? an empirical study on quantized reasoning models.
\newblock \emph{arXiv preprint arXiv:2504.04823}.

\bibitem[{Liu et~al.(2023)Liu, Iter, Xu, Wang, Xu, and Zhu}]{liu2023g}
Yang Liu, Dan Iter, Yichong Xu, Shuohang Wang, Ruochen Xu, and Chenguang Zhu. 2023.
\newblock \href {https://doi.org/10.18653/v1/2023.emnlp-main.153} {{G-Eval}: {NLG} evaluation using {GPT}-4 with better human alignment}.
\newblock In \emph{Proceedings of the 2023 Conference on Empirical Methods in Natural Language Processing}, pages 2511--2522, Singapore. Association for Computational Linguistics.

\bibitem[{Liu et~al.(2024{\natexlab{b}})Liu, Meng, Wu, Peng, Yao, Guan, Tang, Ma, Wang, and Zhu}]{liu2024quantizegeneralization}
Yijun Liu, Yuan Meng, Fang Wu, Shenhao Peng, Hang Yao, Chaoyu Guan, Chen Tang, Xinzhu Ma, Zhi Wang, and Wenwu Zhu. 2024{\natexlab{b}}.
\newblock \href {https://arxiv.org/abs/2406.12928} {Evaluating the generalization ability of quantized {LLMs}: Benchmark, analysis, and toolbox}.
\newblock \emph{arXiv preprint arXiv:2406.12928}.

\bibitem[{Luccioni et~al.(2024)Luccioni, Jernite, and Strubell}]{luccioni2024power}
Sasha Luccioni, Yacine Jernite, and Emma Strubell. 2024.
\newblock \href {https://doi.org/10.1145/3630106.3658542} {Power hungry processing: Watts driving the cost of {AI} deployment?}
\newblock In \emph{Proceedings of the 2024 ACM Conference on Fairness, Accountability, and Transparency}, FAccT '24, page 85–99, New York, NY, USA. Association for Computing Machinery.

\bibitem[{Ma et~al.(2024)Ma, Wang, Ma, Wang, Wang, Huang, Dong, Wang, Xue, and Wei}]{ma2024era}
Shuming Ma, Hongyu Wang, Lingxiao Ma, Lei Wang, Wenhui Wang, Shaohan Huang, Lifeng Dong, Ruiping Wang, Jilong Xue, and Furu Wei. 2024.
\newblock The era of 1-bit {LLM}s: All large language models are in 1.58 bits.
\newblock \emph{arXiv preprint arXiv:2402.17764}, 1.

\bibitem[{Marchisio et~al.(2024)Marchisio, Dash, Chen, Aumiller, {\"U}st{\"u}n, Hooker, and Ruder}]{marchisio2024does}
Kelly Marchisio, Saurabh Dash, Hongyu Chen, Dennis Aumiller, Ahmet {\"U}st{\"u}n, Sara Hooker, and Sebastian Ruder. 2024.
\newblock \href {https://aclanthology.org/2024.findings-emnlp.935.pdf} {How does quantization affect multilingual {LLMs}?}
\newblock In \emph{Findings of the Association for Computational Linguistics: EMNLP 2024}, pages 15928--15947.

\bibitem[{{Meta AI}(2024)}]{meta2024llamaquant}
{Meta AI}. 2024.
\newblock \href {https://ai.meta.com/blog/meta-llama-quantized-lightweight-models/?utm_source=chatgpt.com} {{Meta Llama}: Introducing quantized lightweight {Llama} 3 models}.
\newblock Accessed: 2025-05-18.

\bibitem[{Micikevicius et~al.(2022)Micikevicius, Stosic, Burgess, Cornea, Dubey, Grisenthwaite, Ha, Heinecke, Judd, Kamalu et~al.}]{micikevicius2022fp8}
Paulius Micikevicius, Dusan Stosic, Neil Burgess, Marius Cornea, Pradeep Dubey, Richard Grisenthwaite, Sangwon Ha, Alexander Heinecke, Patrick Judd, John Kamalu, et~al. 2022.
\newblock \href {https://arxiv.org/abs/2209.05433} {{FP8} formats for deep learning}.
\newblock \emph{arXiv preprint arXiv:2209.05433}.

\bibitem[{Min et~al.(2023)Min, Krishna, Lyu, Lewis, Yih, Koh, Iyyer, Zettlemoyer, and Hajishirzi}]{min2023factscore}
Sewon Min, Kalpesh Krishna, Xinxi Lyu, Mike Lewis, Wen-tau Yih, Pang Koh, Mohit Iyyer, Luke Zettlemoyer, and Hannaneh Hajishirzi. 2023.
\newblock \href {https://aclanthology.org/2023.emnlp-main.741/} {{FActScore}: Fine-grained atomic evaluation of factual precision in long form text generation}.
\newblock In \emph{Proceedings of the 2023 Conference on Empirical Methods in Natural Language Processing}, pages 12076--12100.

\bibitem[{Nakagawa et~al.(2017)Nakagawa, Johnson, and Schielzeth}]{nakagawa2017coefficient}
Shinichi Nakagawa, Paul~CD Johnson, and Holger Schielzeth. 2017.
\newblock The coefficient of determination r2 and intra-class correlation coefficient from generalized linear mixed-effects models revisited and expanded.
\newblock \emph{Journal of the Royal Society Interface}, 14(134):20170213.

\bibitem[{NeuralMagic(2024)}]{llmcompressorcode}
NeuralMagic. 2024.
\newblock \href {https://github.com/vllm-project/llm-compressor} {{LLMCompressor} {GitHub} repository}.
\newblock Accessed: April 7, 2025.

\bibitem[{Peng et~al.(2024)Peng, Quesnelle, Fan, and Shippole}]{peng2024yarn}
Bowen Peng, Jeffrey Quesnelle, Honglu Fan, and Enrico Shippole. 2024.
\newblock \href {https://openreview.net/forum?id=wHBfxhZu1u} {{YaRN}: Efficient context window extension of large language models}.
\newblock In \emph{The Twelfth International Conference on Learning Representations}.

\bibitem[{Qwen(2024{\natexlab{a}})}]{quantcollectionqwen25}
Qwen. 2024{\natexlab{a}}.
\newblock \href {https://huggingface.co/collections/Qwen/qwen25-66e81a666513e518adb90d9e} {Qwen-2.5 quantization collection on {HuggingFace}}.
\newblock Accessed: April 7, 2025.

\bibitem[{Qwen(2024{\natexlab{b}})}]{qwenvllmdocs}
Qwen. 2024{\natexlab{b}}.
\newblock \href {https://qwen.readthedocs.io/} {Qwen documentation}.
\newblock \url{https://qwen.readthedocs.io/en/latest/deployment/vllm.html#extended-context-support} (Deployment with vLLM: Extended Context Support).
\newblock Accessed: March 4, 2025.

\bibitem[{Qwen(2024{\natexlab{c}})}]{qwen2.5}
Qwen. 2024{\natexlab{c}}.
\newblock \href {https://qwenlm.github.io/blog/qwen2.5/} {Qwen2.5: A party of foundation models}.

\bibitem[{Samsi et~al.(2023)Samsi, Zhao, McDonald, Li, Michaleas, Jones, Bergeron, Kepner, Tiwari, and Gadepally}]{samsi2023words}
Siddharth Samsi, Dan Zhao, Joseph McDonald, Baolin Li, Adam Michaleas, Michael Jones, William Bergeron, Jeremy Kepner, Devesh Tiwari, and Vijay Gadepally. 2023.
\newblock From words to watts: Benchmarking the energy costs of large language model inference.
\newblock In \emph{2023 IEEE High Performance Extreme Computing Conference (HPEC)}, pages 1--9. IEEE.

\bibitem[{Shaw et~al.(2024)Shaw, Kurtz, Adkins, and Fineran}]{llmcompressorblog}
Robert Shaw, Mark Kurtz, Sara Adkins, and Benjamin Fineran. 2024.
\newblock \href {https://developers.redhat.com/articles/2024/08/14/llm-compressor-here-faster-inference-vllm} {{LLM Compressor}: Faster inference with {vLLM}}.
\newblock Published by Red Hat; authors affiliated with NeuralMagic.
\newblock Accessed: April 7, 2025.

\bibitem[{Song et~al.(2024)Song, Kim, and Iyyer}]{song2024veriscore}
Yixiao Song, Yekyung Kim, and Mohit Iyyer. 2024.
\newblock \href {https://arxiv.org/abs/2406.19276} {{VeriScore}: Evaluating the factuality of verifiable claims in long-form text generation}.
\newblock In \emph{Findings of the Association for Computational Linguistics: EMNLP 2024}, pages 9447--9474.

\bibitem[{vLLM(2024)}]{vllmdocs}
vLLM. 2024.
\newblock \href {https://docs.vllm.ai/} {{vLLM} documentation}.
\newblock \url{https://docs.vllm.ai/}, including subsections: \url{https://docs.vllm.ai/en/latest/features/quantization/fp8.html} and \url{https://docs.vllm.ai/en/latest/features/quantization/bnb.html}.
\newblock Accessed: April 7, 2025.

\bibitem[{Wang et~al.(2025{\natexlab{a}})Wang, Liu, Du, Zhu, Du, Kawaguchi, and Pang}]{wang2025when}
Haonan Wang, Qian Liu, Chao Du, Tongyao Zhu, Cunxiao Du, Kenji Kawaguchi, and Tianyu Pang. 2025{\natexlab{a}}.
\newblock \href {https://openreview.net/forum?id=gwXfZ3xkUq} {When precision meets position: {BF}loat16 breaks down {R}o{PE} in long-context training}.
\newblock \emph{Transactions on Machine Learning Research}.

\bibitem[{Wang et~al.(2025{\natexlab{b}})Wang, Wang, Feldhus, Ostermann, Cao, Sch{\"u}tze, M{\"o}ller, and Schmitt}]{wang2025through}
Qianli Wang, Mingyang Wang, Nils Feldhus, Simon Ostermann, Yuan Cao, Hinrich Sch{\"u}tze, Sebastian M{\"o}ller, and Vera Schmitt. 2025{\natexlab{b}}.
\newblock Through a compressed lens: Investigating the impact of quantization on {LLM} explainability and interpretability.
\newblock \emph{arXiv preprint arXiv:2505.13963}.

\bibitem[{Westfall et~al.(1997)Westfall, Johnson, and Utts}]{westfall1997bayesian}
Peter~H Westfall, Wesley~O Johnson, and Jessica~M Utts. 1997.
\newblock A {B}ayesian perspective on the {B}onferroni adjustment.
\newblock \emph{Biometrika}, 84(2):419--427.

\bibitem[{Williams and Aletras(2023)}]{williams2023impact}
Miles Williams and Nikolaos Aletras. 2023.
\newblock On the impact of calibration data in post-training quantization and pruning.
\newblock \emph{arXiv preprint arXiv:2311.09755}.

\bibitem[{Xiao et~al.(2023)Xiao, Lin, Seznec, Wu, Demouth, and Han}]{xiao2023smoothquant}
Guangxuan Xiao, Ji~Lin, Mickael Seznec, Hao Wu, Julien Demouth, and Song Han. 2023.
\newblock \href {https://proceedings.mlr.press/v202/xiao23c/xiao23c.pdf} {{SmoothQuant}: Accurate and efficient post-training quantization for large language models}.
\newblock In \emph{Proceedings of the 40th International Conference on Machine Learning}.

\bibitem[{Xu et~al.(2024)Xu, Gupta, Li, Bentham, and Srikumar}]{xu2024beyond}
Zhichao Xu, Ashim Gupta, Tao Li, Oliver Bentham, and Vivek Srikumar. 2024.
\newblock \href {https://aclanthology.org/2024.findings-emnlp.901.pdf} {Beyond perplexity: Multi-dimensional safety evaluation of llm compression}.
\newblock In \emph{Findings of the Association for Computational Linguistics: EMNLP 2024}, pages 15359--15396.

\bibitem[{Yang et~al.(2024)Yang, Yang, Zhang, Hui, Zheng, Yu, Li, Liu, Huang, Wei et~al.}]{yang2024qwen2}
An~Yang, Baosong Yang, Beichen Zhang, Binyuan Hui, Bo~Zheng, Bowen Yu, Chengyuan Li, Dayiheng Liu, Fei Huang, Haoran Wei, et~al. 2024.
\newblock \href {https://arxiv.org/abs/2412.15115} {Qwen2.5 technical report}.
\newblock \emph{arXiv preprint arXiv:2412.15115}.

\bibitem[{Yao et~al.(2024)Yao, Wu, Li, Youn, and He}]{yao2023zeroquant}
Zhewei Yao, Xiaoxia Wu, Cheng Li, Stephen Youn, and Yuxiong He. 2024.
\newblock \href {https://doi.org/10.1609/aaai.v38i17.29908} {Exploring post-training quantization in llms from comprehensive study to low rank compensation}.
\newblock In \emph{Proceedings of the Thirty-Eighth AAAI Conference on Artificial Intelligence and Thirty-Sixth Conference on Innovative Applications of Artificial Intelligence and Fourteenth Symposium on Educational Advances in Artificial Intelligence}, AAAI'24/IAAI'24/EAAI'24. AAAI Press.

\bibitem[{Yen et~al.(2025)Yen, Gao, Hou, Ding, Fleischer, Izsak, Wasserblat, and Chen}]{yen2024helmet}
Howard Yen, Tianyu Gao, Minmin Hou, Ke~Ding, Daniel Fleischer, Peter Izsak, Moshe Wasserblat, and Danqi Chen. 2025.
\newblock \href {https://openreview.net/forum?id=293V3bJbmE} {{HELMET}: How to evaluate long-context models effectively and thoroughly}.
\newblock In \emph{The Thirteenth International Conference on Learning Representations}.

\bibitem[{Zhang et~al.(2025)Zhang, Wang, Li, Wu, Tang, Liu, He, Yin, and Wang}]{zhang2025catastrophic}
Zhiwei Zhang, Fali Wang, Xiaomin Li, Zongyu Wu, Xianfeng Tang, Hui Liu, Qi~He, Wenpeng Yin, and Suhang Wang. 2025.
\newblock \href {https://openreview.net/forum?id=lHSeDYamnz} {Catastrophic failure of {LLM} unlearning via quantization}.
\newblock In \emph{The Thirteenth International Conference on Learning Representations}.

\bibitem[{Zheng et~al.(2025)Zheng, Li, Chu, Feng, Ma, Luo, Guo, Qin, Magno, and Liu}]{zheng2025empirical}
Xingyu Zheng, Yuye Li, Haoran Chu, Yue Feng, Xudong Ma, Jie Luo, Jinyang Guo, Haotong Qin, Michele Magno, and Xianglong Liu. 2025.
\newblock An empirical study of {Qwen3} quantization.
\newblock \emph{arXiv preprint arXiv:2505.02214}.

\bibitem[{Zhou et~al.(2024)Zhou, Ning, Hong, Fu, Xu, Li, Lou, Wang, Yuan, Li et~al.}]{zhou2024survey}
Zixuan Zhou, Xuefei Ning, Ke~Hong, Tianyu Fu, Jiaming Xu, Shiyao Li, Yuming Lou, Luning Wang, Zhihang Yuan, Xiuhong Li, et~al. 2024.
\newblock A survey on efficient inference for large language models.
\newblock \emph{arXiv preprint arXiv:2404.14294}.

\bibitem[{Zhu et~al.(2024)Zhu, Li, Liu, Ma, and Wang}]{zhu2024survey}
Xunyu Zhu, Jian Li, Yong Liu, Can Ma, and Weiping Wang. 2024.
\newblock A survey on model compression for large language models.
\newblock \emph{Transactions of the Association for Computational Linguistics}, 12:1556--1577.

\end{thebibliography}
% Uncomment the following line if a specific bibliography style is required
% \bibliographystyle{acl_natbib}

\clearpage
\appendix
\textbf{\Large Appendix}

\renewcommand{\arraystretch}{1.25} % increase row height

\begin{table*}[!ht]
\centering
\resizebox{\textwidth}{!}{%
\begin{tabular}{p{0.10\textwidth}  
                p{0.13\textwidth}  
                p{0.32\textwidth}  
                p{0.40\textwidth}} 
\toprule
\textbf{Model Family} & \textbf{Quantization} & \textbf{Models \& Source Links} & \textbf{Key Details} \\
\midrule

\multirow{5}{*}[0pt]{\centering \llama{}} 
 & AWQ-int4 
 & \href{https://huggingface.co/hugging-quants/Meta-Llama-3.1-8B-Instruct-AWQ-INT4}{8B}, \href{https://huggingface.co/hugging-quants/Meta-Llama-3.1-70B-Instruct-AWQ-INT4}{70B} (HuggingQuants) 
 & Zero-point quantization, group size 128, GEMM kernels. \\

 & GPTQ-int4 
 & \href{https://huggingface.co/RedHatAI/Meta-Llama-3.1-8B-Instruct-quantized.w4a16}{8B}, \href{https://huggingface.co/RedHatAI/Meta-Llama-3.1-70B-Instruct-quantized.w4a16}{70B} (RedHat AI) 
 & Symmetric per-group quantization with linear scaling, 10\% damping, 768 calibration sequences. Only weights quantized. \\

 & GPTQ-int8 
 & \href{https://huggingface.co/RedHatAI/Meta-Llama-3.1-8B-Instruct-quantized.w8a16}{8B} (RedHat AI) 
 & Symmetric per-channel quantization with linear scaling, 1\% damping, 256 sequences of 8192 tokens. Only weights quantized. \\

 & GPTQ-int8 
 & Locally created 70B quantization 
 & W8A16 quantization adapted from RedHat AI’s \href{https://huggingface.co/RedHatAI/Meta-Llama-3.1-70B-Instruct-quantized.w8a8}{70B W8A8 recipe}, integrating SmoothQuant into GPTQ \citep{kurtic2024give}. \\

 & FP8 
 & \href{https://huggingface.co/RedHatAI/Meta-Llama-3.1-8B-Instruct-FP8-dynamic}{8B}, \href{https://huggingface.co/RedHatAI/Meta-Llama-3.1-70B-Instruct-FP8-dynamic}{70B} (RedHat AI) 
 & Symmetric per-channel quantization on weights and activations with per-token dynamic scaling. \\

 & BNB-nf4 
 & In-flight quantization done in \href{https://github.com/vllm-project/vllm}{vLLM}
 & Weights converted from BF16 to nf4 blocks at load time; activations remain in BF16. \\

\midrule

\multirow{4}{*}[0pt]{\centering \qwen{}} 
 & AWQ-int4 
 & \href{https://huggingface.co/Qwen/Qwen2.5-7B-Instruct-AWQ-Int4}{7B}, \href{https://huggingface.co/Qwen/Qwen2.5-32B-Instruct-AWQ-Int4}{32B}, \href{https://huggingface.co/Qwen/Qwen2.5-72B-Instruct-AWQ-Int4}{72B} 
 & Official \qwen{} release \citep{quantcollectionqwen25}. \\

 & GPTQ-int4 
 & \href{https://huggingface.co/Qwen/Qwen2.5-7B-Instruct-GPTQ-Int4}{7B}, \href{https://huggingface.co/Qwen/Qwen2.5-32B-Instruct-GPTQ-Int4}{32B}, \href{https://huggingface.co/Qwen/Qwen2.5-72B-Instruct-GPTQ-Int4}{72B} 
 & Official \qwen{} release \citep{quantcollectionqwen25}. \\

 & GPTQ-int8 
 & \href{https://huggingface.co/Qwen/Qwen2.5-7B-Instruct-GPTQ-Int8}{7B}, \href{https://huggingface.co/Qwen/Qwen2.5-32B-Instruct-GPTQ-Int8}{32B}, \href{https://huggingface.co/Qwen/Qwen2.5-72B-Instruct-GPTQ-Int8}{72B} 
 & Official \qwen{} release \citep{quantcollectionqwen25}. \\

 & FP8 
 & Locally created FP8 \qwen{} models using the same recipe as \llama{} family's FP8 detailed above.
 & Dynamic quantization using RedHat AI’s \llama{} 70B FP8 recipe \citep{kurtic2024give}. \\

 & BNB-nf4 
 & In-flight quantization done in \href{https://github.com/vllm-project/vllm}{vLLM} 
 & Weights converted from BF16 to nf4 blocks at load time; activations remain in BF16. \\

\bottomrule
\end{tabular}
}
\caption{Summary of quantized model sources, quantization formats, and implementation details.}
\label{tab:quantization-info}
\end{table*}

\section{Experimental setup: additional details}
\label{app:sec-exp_setup}

\subsection{Obtaining quantized models}
\label{app:sec-quantize_sources}

Our objective was to study how easily available quantized models handle long-context and long-form tasks. So, we primarily relied on quantized models provided widely by RedHat AI (former NeuralMagic; \cite{kurtic2024give}), HuggingQuants, and \qwen{}'s official repository, since these are popular resources for quantized versions of \llama{} and \qwen{}, with up to 300K downloads as of July 2025. 

\Cref{tab:quantization-info} provides an overview of how quantized models were obtained for this study. Some specifics are detailed below.

For \llama{} 70B, we observed noticeable accuracy degradation upon using the default  GPTQ-int8 (W8A16) quantization recipe. \citet{kurtic2024give} encountered the same issue with the W8A8 version and resolved it by integrating SmoothQuant \citep{xiao2023smoothquant} into their W8A8 GPTQ-int8 recipe. We use this same quantization recipe from \citet{kurtic2024give} and performed a W8A16 GPTQ-int8 quantization, by which we were able to largely retain the pre-quantization accuracy.

For \qwen{} models, FP8-Dynamic models were created locally using LLMCompressor \citep{llmcompressorblog, llmcompressorcode}, borrowing the same quantization recipe used by \citet{kurtic2024give} for \llama.

For BNB-nf4, we use in-flight quantization directly from the full-precision checkpoints \citep{vllmdocs}. We launch vLLM with quantization=\textit{bitsandbytes}. This converts BF16 weights to nf4 blocks at load time, and then runs inference by de-quantizing those blocks on-chip before each GEMM, while activations remain in BF16. We do not experiment with the 8-bit integer BNB quantization \citep{dettmers2022gpt3} since it is not supported on vLLM.\footnote{\url{https://github.com/vllm-project/vllm/issues/11655}.}

\subsection{Using \qwen{} for long context tasks:}
\label{app:qwen_long_ctx}
Though \qwen{} has a context length of 128K, the default configuration is set to support context lengths only up to 32K; for longer contexts, the official Qwen documentation \citep{qwenvllmdocs} suggests employing \textsc{YaRN} \citep{peng2024yarn} to extrapolate the RoPE scaling and achieve 4$\times$ the context length. We adopt the same approach and modify Qwen model configuration when running on context lengths larger than 32K.

\subsection{Computational Cost} 
We estimate the approximate compute consumption as follows: \nocha{}: 1800 hours, \ruler{}: 500 hours, \oneruler{}: 3200 hours, \csfour{}: 180 hours, and \factscore{}: 100 hours, totaling roughly 5800 GPU hours on NVIDIA A100 GPUs.

% % left bottom right top
% \begin{figure}[t]
%     \centering
%     \resizebox{0.99\linewidth}{!}{
%         \includegraphics[clip, trim=0pt 0pt 0pt 0pt]{images/all_quantwise.pdf}
%     }
%   \caption{Average drop in performance by quantization and task.\anmol{need to move legend to top of image again, also send to app, borrow color scheme from fig 3. also clean up x-axis and don't leave overalps}}
%   \label{fig:all_quantwise}
% \end{figure}

% \begin{figure*}[t]
%   \centering

%\begin{figure}[t] 
%    \centering    \includegraphics[width=\linewidth]{images/Ruler_Overall.pdf}
    %\vspace{-2pt}
%    \caption{Average $\Delta$-accuracy of different quantization methods relative to BF16 on the \ruler{} benchmark. (Average score of BF16: 66.27)}
%    \label{fig:ruler-overall}
%\end{figure}

\section{Benchmarks: additional details}
\label{app:sec-benchmarks}
 
 In this section, we present more details about each benchmark task, adding to the description in \S\ref{sec:exp_setup}.

\paragraph{\ruler:} While \citet{kurtic2024give} evaluate quantized LLMs on \ruler{} and report coarse-grained, aggregated results across quantized models and context lengths, we perform a more detailed analysis. We follow \citet{yen2024helmet} to select our datapoints for the multi-key needle, multi-key UUID, and multi-value tasks from the \ruler{} benchmark. In the multi-key task, the model must retrieve the value associated with a specific key given in the query, while ignoring others which serve as distractors. The multi-key UUID task tests a similar ability, but the values are 32-character long-form strings (UUIDs), making it harder to match or guess, testing the model’s ability to precisely extract the correct value without partial matching. Finally, the multi-value task increases the difficulty further by requiring the models to retrieve multiple values associated with the same key in the haystack. While the tokenizers for \llama{} and \qwen{} differ, \citet{yen2024helmet} use \llama{} tokenizer to tokenize and construct their benchmark. This implies that \llama{} and \qwen{} receive slightly different numbers of input tokens for the same input datapoint.

\paragraph{\oneruler:} This benchmark extends \ruler{} in two significant ways. First, \oneruler's NIAH task includes points without any correct needle present, requiring models to abstain appropriately. Second, \oneruler{} expands the task beyond English to encompass 26 languages representing diverse language families and writing systems. They find that accuracy on NIAH shows large drops on low-resource languages, especially ones that use non-Latin scripts. We evaluate models using two NIAH task types: NIAH-single and NIAH-none. In NIAH-single, a single correct needle is present without distractors, which models must retrieve accurately. In NIAH-none, no correct needle is present, but four distractor needles exist, requiring models to abstain from retrieving any needle.

\paragraph{\nocha:} \nocha{} evaluates long-context reasoning abilities by testing realistic claim verification and reasoning skills of LLMs on book-length texts. Since \llama{} and \qwen{} support context lengths of up to 128K tokens, this work considers the subset of datapoints with contexts up to that length, yielding 723 claim pairs drawn from 46 novels and three short story collections. While \ruler{} and \oneruler{} help analyze the impact of quantization across varying context lengths, \nocha{} enables a more nuanced investigation. First, it poses a significantly more challenging task for LLMs, with full-precision models from \llama{} and \qwen{} performing at approximately chance level \cite{karpinska2024one}. Additionally, \nocha{} facilitates further analysis through various dataset splits. A split based on \textit{true} versus \textit{false} claims enables the study of potential LLM biases. A difficulty-based split allows for examining the impact of question difficulty, and not just context length. Finally, a reasoning-scope-based split supports analysis across different spans of context required for reasoning.

\paragraph{\factscore{}:} \textit{Evaluation procedure}: the \veriscore{} metric runs a three-stage evaluation pipeline: it extracts individual claims from the text using a task-finetuned LLM, focusing specifically on generating sensible and verifiable factual claims. It then retrieves evidence for each claim through Google search results using the Serper API.\footnote{\url{https://serper.dev/}} Finally, each claim is verified against the retrieved evidence using another task-finetuned LLM. To account for the open-ended nature of the generation, it computes the F1 score at K, where K is the median number of claims extracted across all responses. \textit{Dataset and generation}: see \S\ref{appsec:factscore-example} for a complete example of biography generation prompt and response for the prompt. 

\paragraph{\csfour:} Each \csfour{} input prompt initially consists of a simple instruction and is subsequently expanded by introducing sets of constraints across three complexity levels: specifically, 9, 23, and 39 constraints. By incrementally increasing the number of constraints, we examine the model’s capability to write creatively while adhering to provided instructions. For evaluating constraint satisfaction, we improve upon the evaluation prompt used in \citet{atmakuru2024cs4}, employing GPT4o\footnote{refers to the \texttt{gpt-4o-2024-08-06} endpoint} as the judge (\cref{fig:cs-prompt}). Instead of providing the generated story along with all constraints for evaluation, we feed GPT4o three constraints at a time. GPT4o is also used to evaluate story quality using \bookscore. \bookscore{} evaluates the narrative quality based on language fluency, salience, internal consistency, and related aspects. It is a pairwise evaluation metric that enables comparing stories generated by quantized models against those produced by their full-precision counterparts. See \S\ref{app:cs4-example} for a complete example including the input constraints, generated story, and constraint-satisfaction evaluation.

\begin{figure*}[t]
  \centering
  \begin{varwidth}{0.97\textwidth}
    \small\ttfamily

You are an expert evaluator. You will be given a story followed by a set of three constraints.
Your task is to carefully read both of them and tell how many constraints are being
satisfied in the story. As the output, print "yes" or "no" for each constraint
based on whether it is being satisfied or not, followed by an explanation of
why it is satisfied/violated.
If a constraint is satisfied, print the sentence(s) from the story in
which it is satisfied. If a constraint is not being satisfied, explain how it is being violated. Be very strict in your evaluation.
Mark a constraint as satisfied ("yes") only if it is completely satisfied in
the story. For no satisfaction/partial satisfaction, mark a "no".
Your output should always follow the format: <constraint number><yes/no><explanation>.  

    \end{varwidth}
  \caption{Prompt used to evaluate constraint-satisfaction of stories in the \csfour benchmark.}
  \label{fig:cs-prompt}
\end{figure*}

\section{Further results}

\begin{figure*}[t]
    \centering
    \resizebox{0.75\linewidth}{!}{
        \includegraphics[clip, trim=0pt 0pt 0pt 0pt]{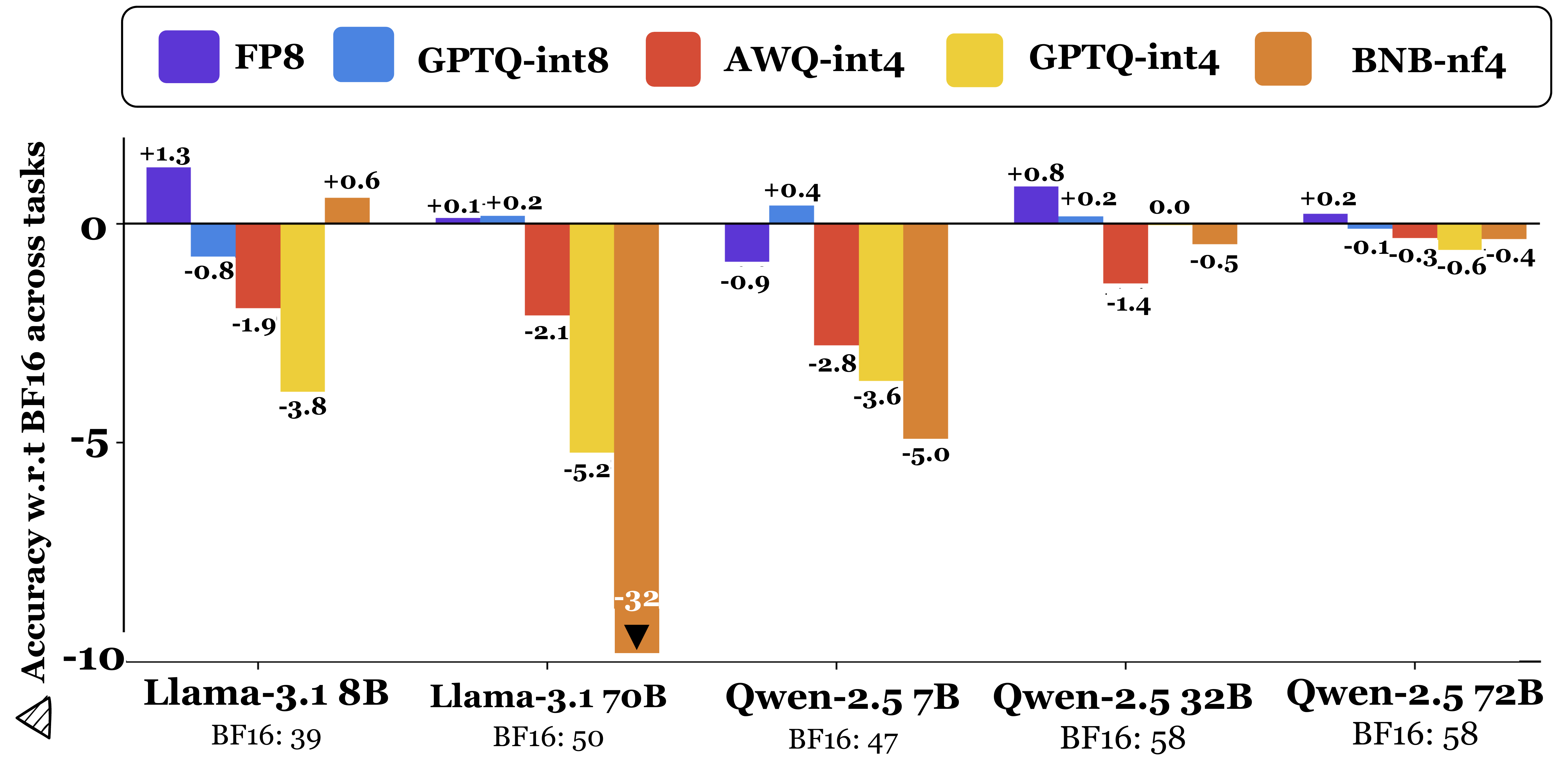}
    }
  \caption{Average performance of each quantization method on each model averaged across all tasks. Performance is illustrated using \deltacc{}, the difference w.r.t. the BF16 baseline. The BF16 baseline average results for each task (over which deltas are computed) are displayed above the x-axis. For \oneruler{} and \ruler{} we only aggregate the long context results from 64K and 128K context lengths.}
  \label{fig:all_modelwise}
\end{figure*}

This section provides and details more results, adding to \S\ref{sec:results}. \autoref{fig:all_modelwise} shows average results across models by the quantization method.

%\begin{figure}[t] 
%    \centering    \includegraphics[width=\linewidth]{images/CS4_Overall.pdf}
    %\vspace{-2pt}
%    \caption{Average $\Delta$-constraint-satisfaction of different quantization methods relative to BF16 on the \csfour{} benchmark. (Average score of BF16: 85.84)}
%    \label{fig:cs-overall}
%end{figure}

%\begin{figure}[t] % 'h' means here
%    \centering
    %\includegraphics[width=\linewidth]{images/SelfBLEU_heatmap.pdf}
    %\vspace{-2pt}
%    \caption{Average 1000*SelfBLEU scores of different model-quantization pairs on \csfour.}
%    \label{fig:SelfBLEU_heatmap}
%\end{figure}

%\begin{figure}[t] % 'h' means here
%    \centering
    %\includegraphics[width=\linewidth]{images/CS_Repetitions_heatmap.pdf}
    %\vspace{-2pt}
   % \caption{Average 1000*5-gram repetition rates of different model-quantization pairs on \csfour.}
    %\label{fig:CS_Repetitions_heatmap}
%\end{figure}

% \section{Task-wise detailed results analysis}
% \label{sec:task-wise_detailed}

\subsection{\ruler{}}
\label{appsec:ruler}
\noindent\textbf{GPTQ-int4 and BNB-nf4 consistently underperform across models on \ruler{}:} \Cref{fig:ruler-contextwise} shows the $\Delta$-accuracy of the quantized models relative to BF16 on \ruler. At the 8K context length, all methods perform similarly to BF16. As the context length increases, 8-bit methods remain robust, with accuracy drops under 5\%.\footnote{The $p$ values are larger than 0.05 in \Cref{tab:ruler-posthoc-context} \S\ref{app:stats-test} for FP8 and GPTQ-int8 at 64K and 128K context-lengths compared to BF16.} However, 4-bit methods, especially BNB-nf4, show significant degradation in longer contexts. A post-hoc analysis in \Cref{tab:ruler-posthoc-context} (\S\ref{app:stats-test}) reveals that the performance drop in BNB-nf4 compared to BF16 is statistically significant at the context lengths 64K and 128K  ($p < 0.05$ for both). A model-wise breakdown is shown in \cref{fig:ruler-model-wise}. Except FP8 quantization on \qwen{} 32B, 8-bit quantization methods exhibit less than a 3\% performance drop relative to BF16. On the other hand, GPTQ-int4 and BNB-nf4 show performance drops of at least 3\% on all models and over 10\% on several models: specifically, GPTQ-int4 on \llama{} 8B (21.2\%) and \qwen{} 72B (10.3\%), and BNB-nf4 on \llama{} 8B (30.7\%), \qwen{} 7B(29.7\%), and \qwen{} 32B(15.7\%).
 % Considering the impact of quantization on individual base models, \Cref{fig:ruler-model-wise} shows that BNB-nf4 consistently underperforms relative to BF16 and other quantized models, especially on \llama{} 8B (30.7\%), \qwen{} 7B (29.7\%), and \qwen{} 32B (15.7\%).

\begin{figure}[t] 
    \centering    \includegraphics[width=\linewidth]{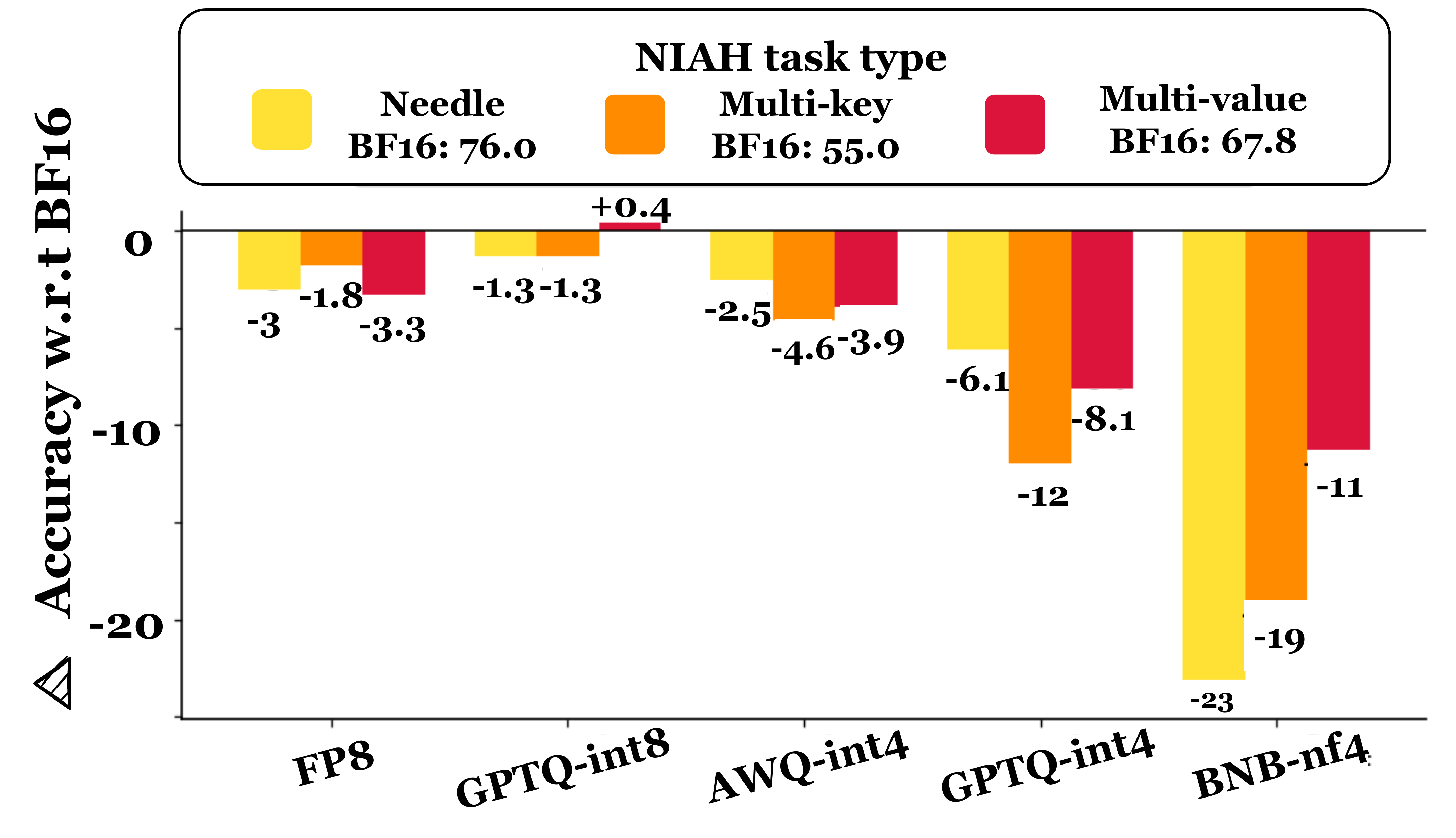}
    %\vspace{-2pt}
    \caption{Average $\Delta$-accuracy of different quantization methods relative to BF16 on the \ruler{} benchmark across different sub-tasks.}
    \label{fig:ruler-task}
\end{figure}

\begin{figure*}[t] % 'h' means here
    \centering
    \includegraphics[width=\linewidth]{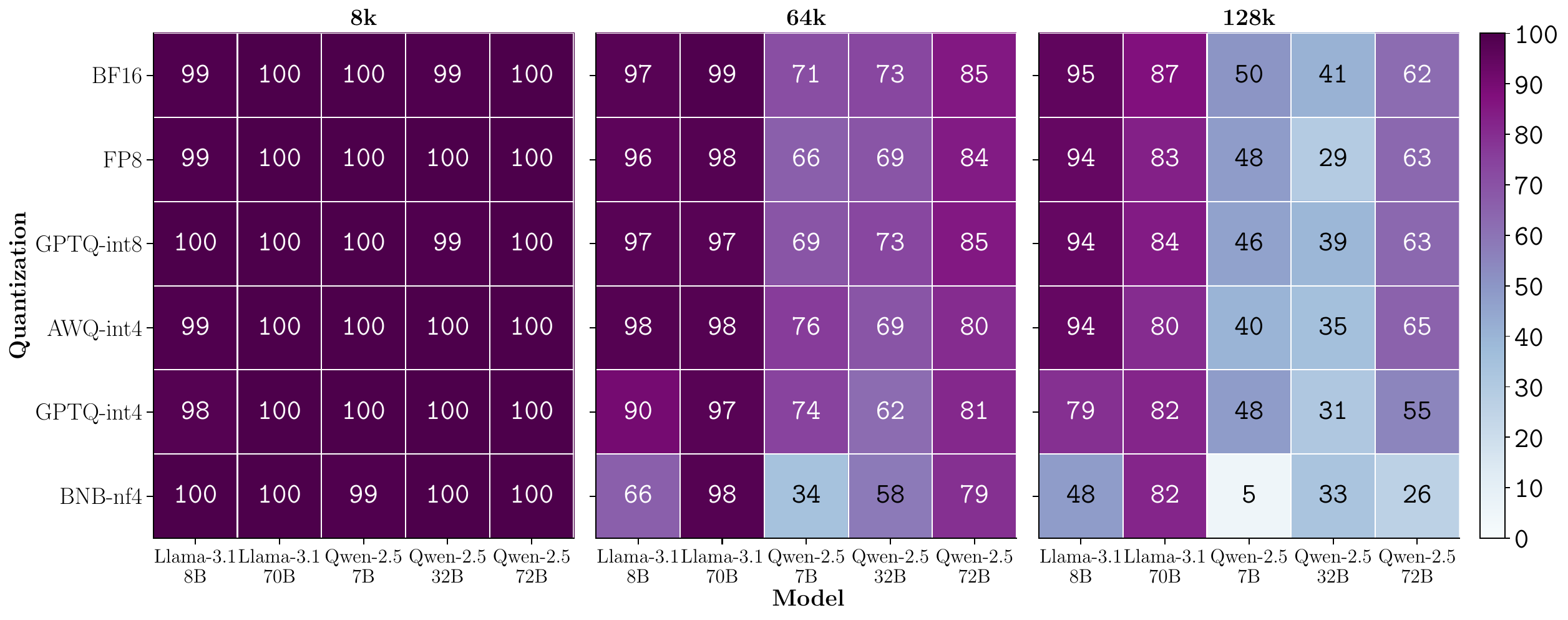}
    %\vspace{-2pt}
    \caption{Performance of all models and quantization methods across context lengths on the Needle task in \ruler.}
    \label{fig:ruler-needle}
\end{figure*}

\begin{figure*}[t] % 'h' means here
    \centering
    \includegraphics[width=\linewidth]{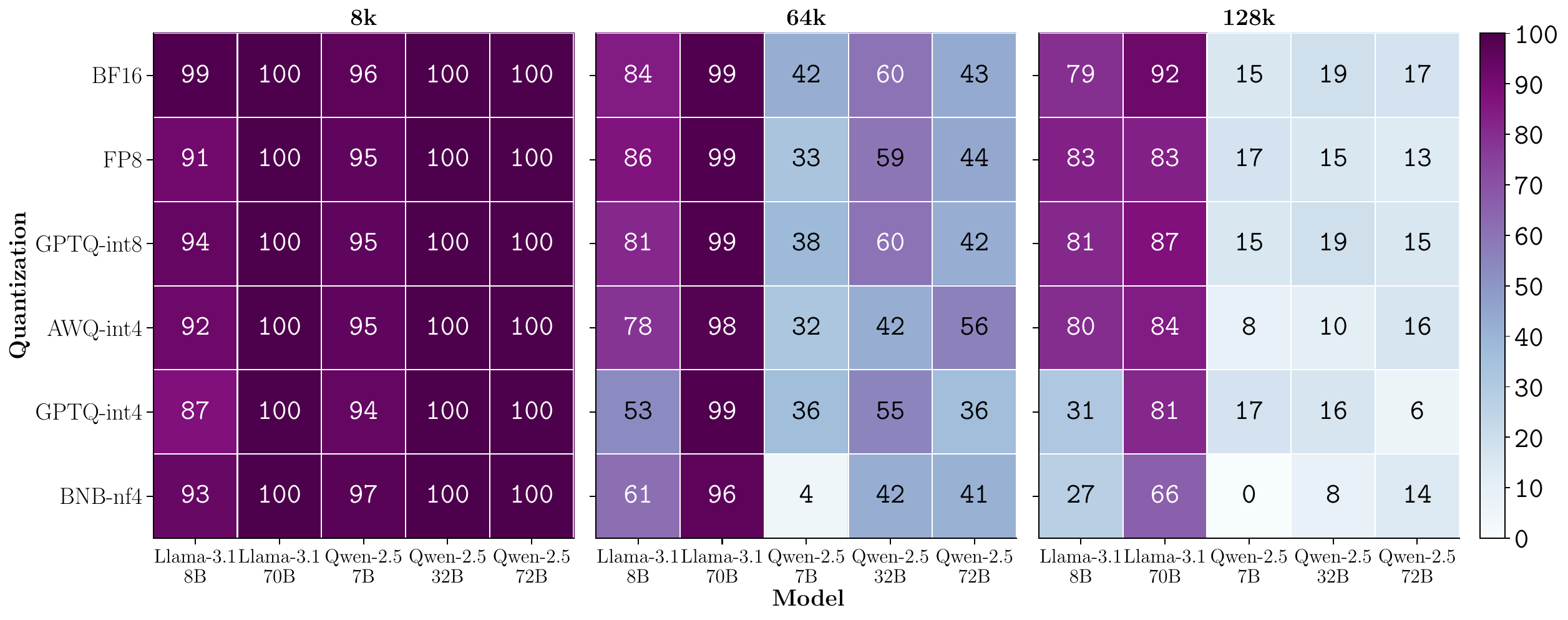}
    %\vspace{-2pt}
    \caption{Performance of all models and quantization methods across context lengths on the Multi-key task in \ruler.}
    \label{fig:ruler-multi-key}
\end{figure*}

\begin{figure*}[t] % 'h' means here
    \centering
    \includegraphics[width=\linewidth]{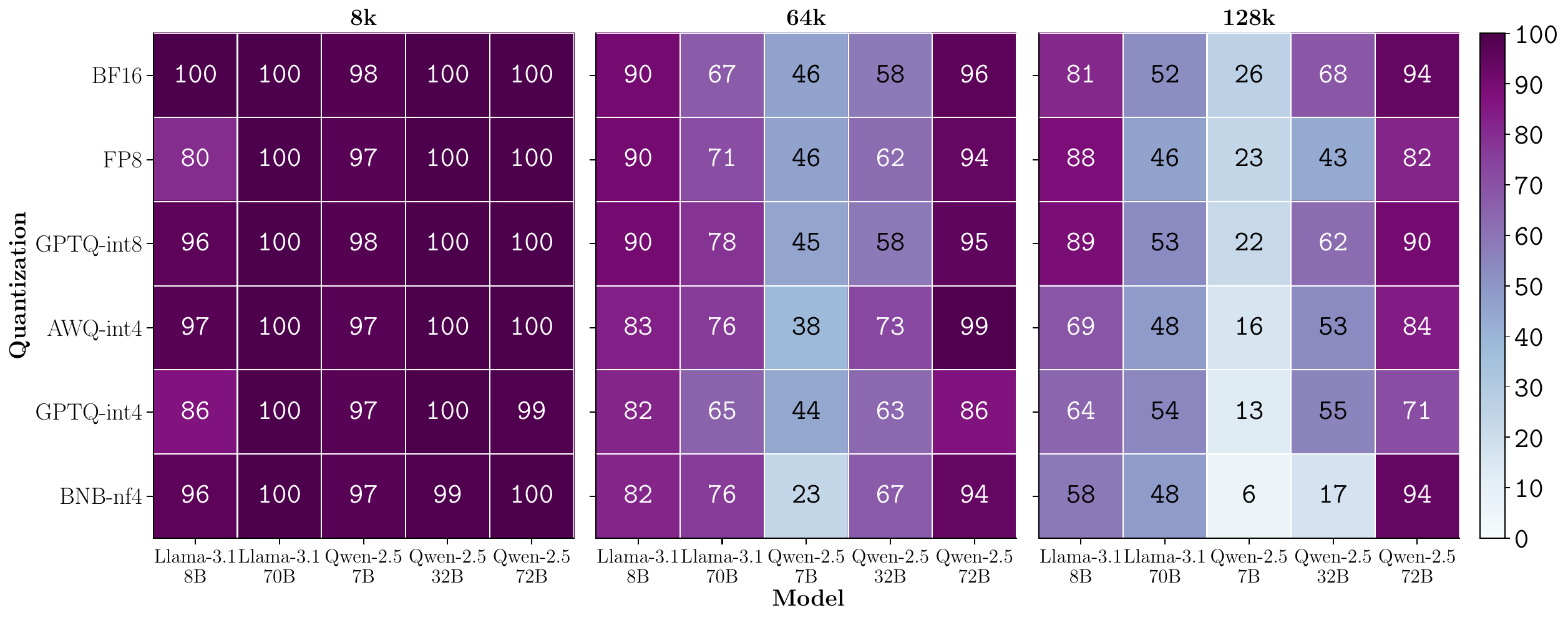}
    %\vspace{-2pt}
    \caption{Performance of all models and quantization methods across context lengths on the Multi-value task in \ruler.}
    \label{fig:ruler-multi-value}
\end{figure*}

\noindent\textbf{Quantization-induced degradation does not consistently align with task difficulty:} \cref{fig:ruler-task} analyzes the impact on different sub-tasks of \ruler on quantization. Absolute performance values for these sub-tasks are visualized as heat maps: \cref{fig:ruler-needle} for needle retrieval, \cref{fig:ruler-multi-key} for multi-key retrieval, and \cref{fig:ruler-multi-value} for multi-value retrieval. Although \ruler{} defines \textit{needle} as the easiest of the three tasks, \cref{fig:ruler-task} shows that the performance drop on \textit{needle} is not the lowest for the FP8, GPTQ-int8, and BNB-nf4 quantization methods. This suggests that different quantization methods struggle with different tasks and that increased task difficulty does not directly correlate with greater performance degradation. %\Cref{fig:ruler-needle,fig:ruler-multi-key,fig:ruler-multi-value} further support this observation. 

\noindent\textbf{Larger models are more robust to quantization:} Consistent with prior research \citep{lee2024comprehensive}, \Cref{fig:ruler-model-wise,fig:ruler-needle,fig:ruler-multi-key,fig:ruler-multi-value} demonstrate that within each model family, larger models are generally less affected by quantization and that different models and quantization methods find different tasks difficult.

\subsection{\oneruler{}} 
\begin{figure*}[t] % 'h' means here
    \centering
    \includegraphics[clip, trim=5pt 5pt 5pt 5pt, scale=0.50]{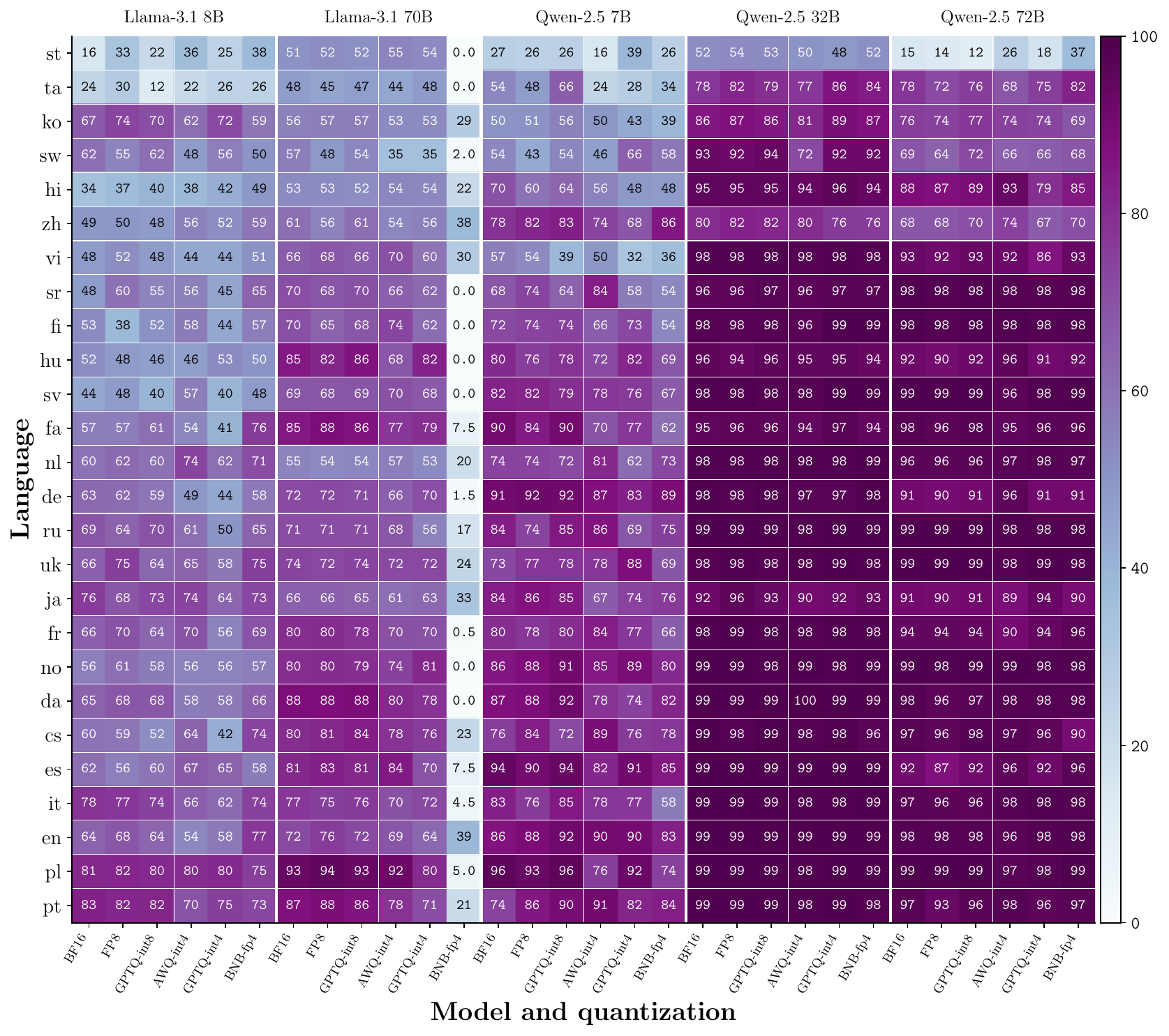}
    %\vspace{-2pt}
    \caption{\textbf{\oneruler}: Performance (accuracy) of all model and quantization configurations across the points in each language. Includes only 64K, 128K lengths due to focus on very long contexts.}
    \label{fig:oneruler-heatmap-lang}
\end{figure*}

\begin{figure*}[t]
  \centering
    \includegraphics[clip, trim=0pt 5pt 0pt 5pt, width=\linewidth]{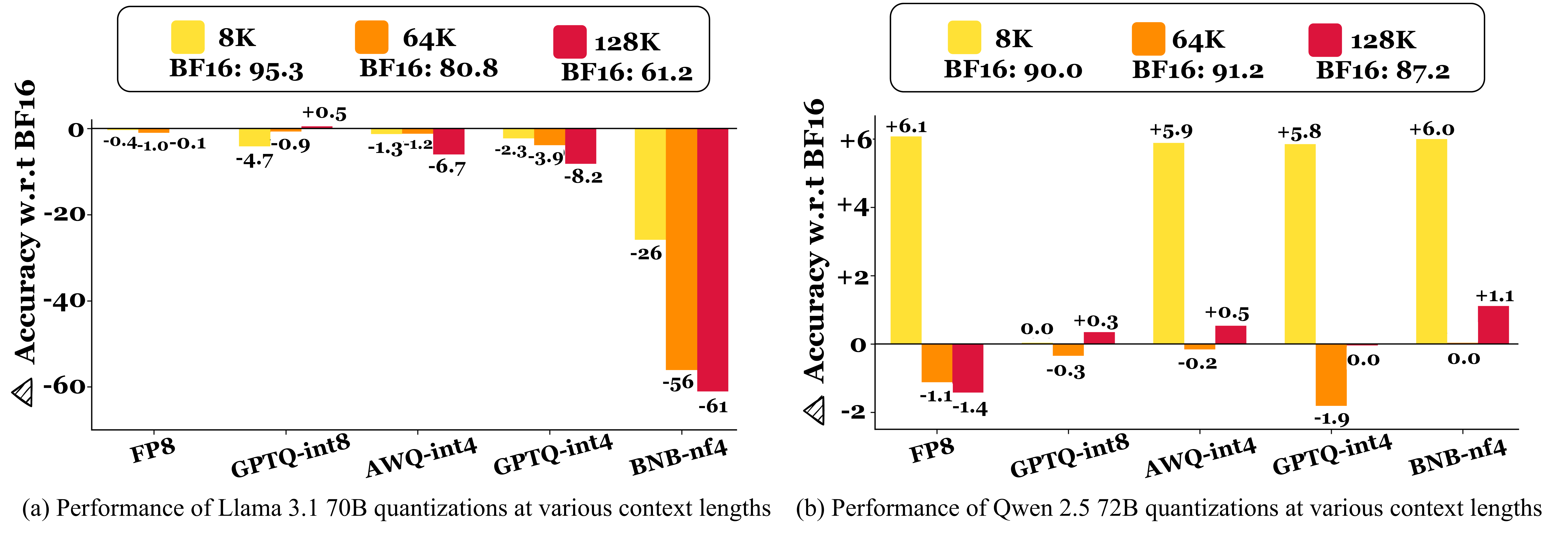}
  \caption{\oneruler{} \deltacc{}, averaged across models, with varying context length for \llama{} 70B and \qwen{} 72B. Both show starkly different responses to quantization.}
  \label{fig:oneruler-barplots-llama-qwen}
\end{figure*}

\noindent\textbf{GPTQ-int4 and BNB-nf4 consistently underperform across models on \oneruler:} \Cref{fig:oneruler-contextwise} presents the $\Delta$-accuracy of the quantized models relative to BF16 on \oneruler{}. FP8 and GPTQ-int8 perform comparably to BF16 across all context lengths. In contrast, 4-bit quantization—particularly BNB-nf4—leads to more pronounced performance degradation as context length increases. The impact of each quantization method, however, varies by model. For instance, BNB-nf4 yields the best results for \llama{} 8B but performs particularly poorly on \llama{} 70B, with a 66\% drop in accuracy. Interestingly, \qwen{} 32B and 72B maintain performance comparable to BF16 even under 4-bit quantization, showing strong robustness to quantization. \autoref{fig:oneruler-barplot-resource} shows no significant performance difference between high- and low-resource languages within each quantization method. While performance drops increase for non-English languages, we do not observe consistent trends distinguishing high- vs. low-resource languages (\autoref{fig:oneruler-barplot-resource}). Finally, \autoref{fig:oneruler-barplot-script} shows the performance of different quantization methods across language scripts, while \autoref{fig:oneruler-heatmap-lang} further breaks down the \oneruler{} results by individual languages across models. \autoref{fig:oneruler-barplots-llama-qwen} shows how \llama{} 70B and \qwen 72B, despite being approximately of the same size, perform very differently across context lengths on \oneruler.

% left bottom right top

\begin{figure}[t] 
    \centering    \includegraphics[width=\linewidth]{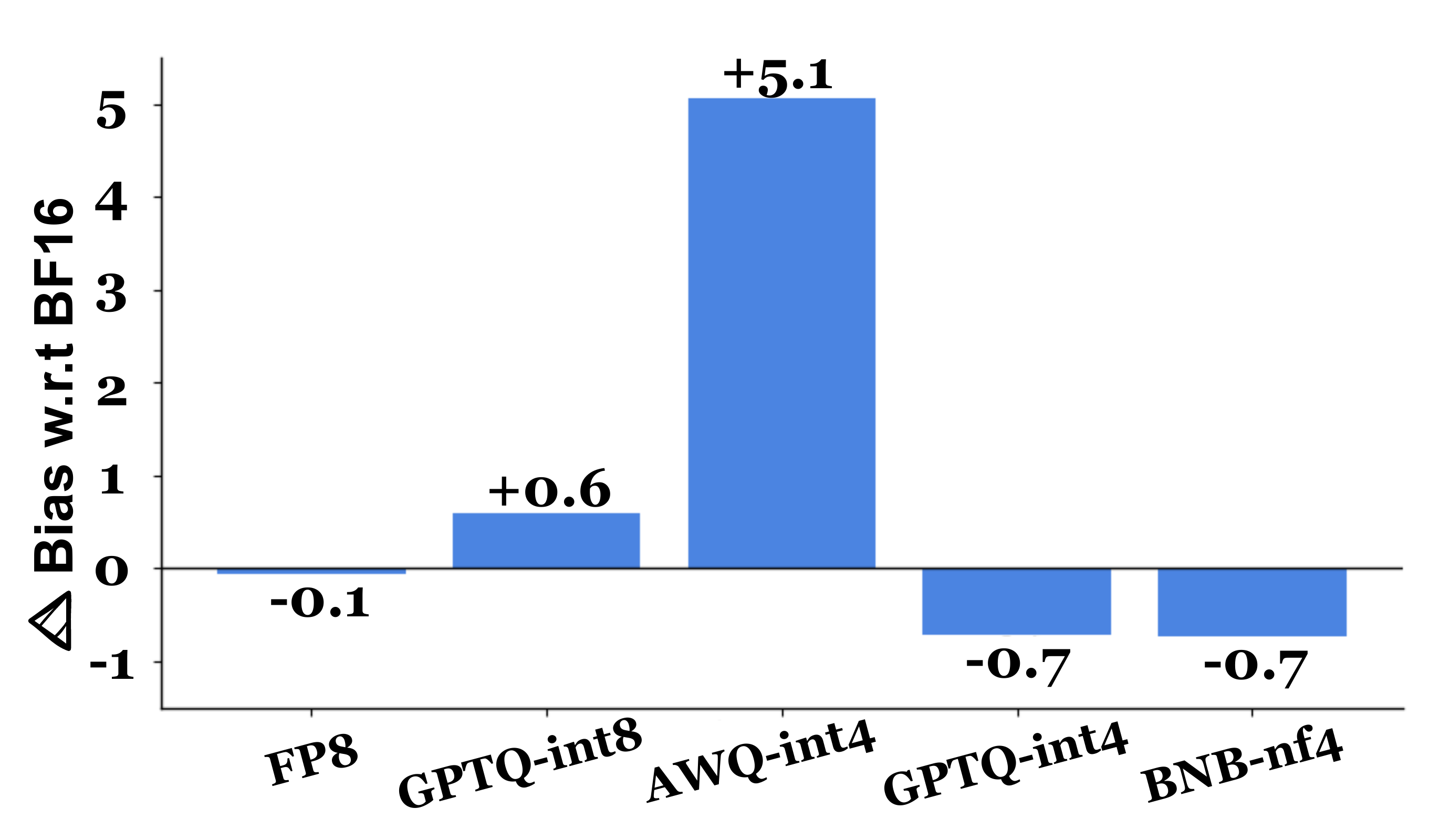}
    %\vspace{-2pt}
    \caption{Average $\Delta$-bias of different quantization methods relative to BF16 on the \nocha{} benchmark. Bias is defined as the performance difference between \textit{false} and \textit{true} claims, indicating a model's tendency to favor one label over the other.}
    \label{fig:nocha-bias}
\end{figure}

\subsection{\nocha{}} 
\label{appsec:nocha} 
\begin{figure}[t] % 'h' means here
    \centering
    \includegraphics[width=\linewidth]{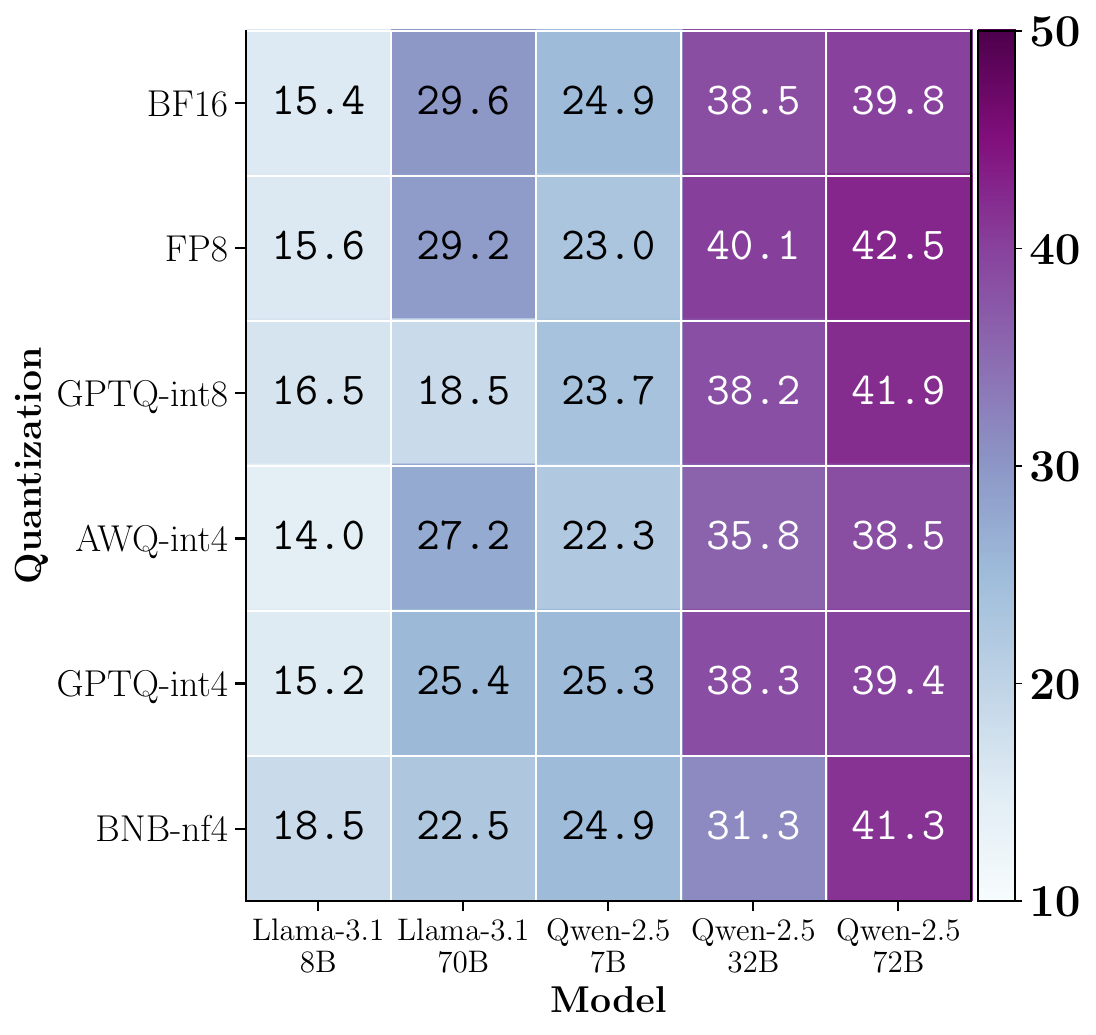}
    %\vspace{-2pt}
    \caption{Performance (accuracy) of all models and quantization methods on the \nocha{} benchmark.}
    \label{fig:nocha-heatmap}
\end{figure}

\noindent\textbf{BF16 performs around chance on \nocha:}
\Cref{fig:nocha-context-length} shows the $\Delta$-accuracy of quantized models relative to BF16 on \nocha{}. The magnitude of the $\Delta$-accuracy is smaller than that observed on \ruler{} and \oneruler{}. This is largely because even the unquantized models perform near chance level (\Cref{fig:nocha-heatmap}), leaving limited room for further degradation from quantization.

\begin{figure*}[t] % 'h' means here
   \centering
   \includegraphics[width=\linewidth]{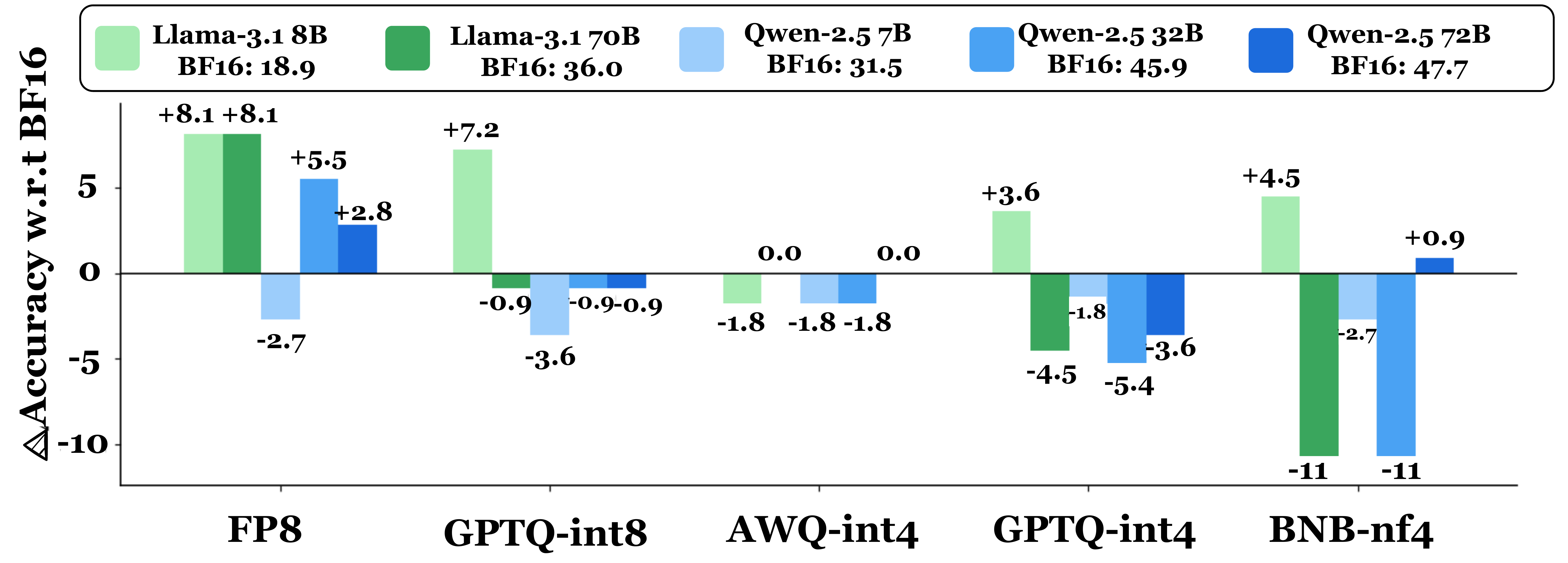}
    \vspace{-2pt}
   \caption{Average $\Delta$-accuracy of different quantization methods relative to BF16 on the \nocha{} benchmark across models when context length $<$75k.}
   \label{fig:nocha-short-model}
\end{figure*}

\begin{figure*}[t] % 'h' means here
   \centering
   \includegraphics[width=\linewidth]{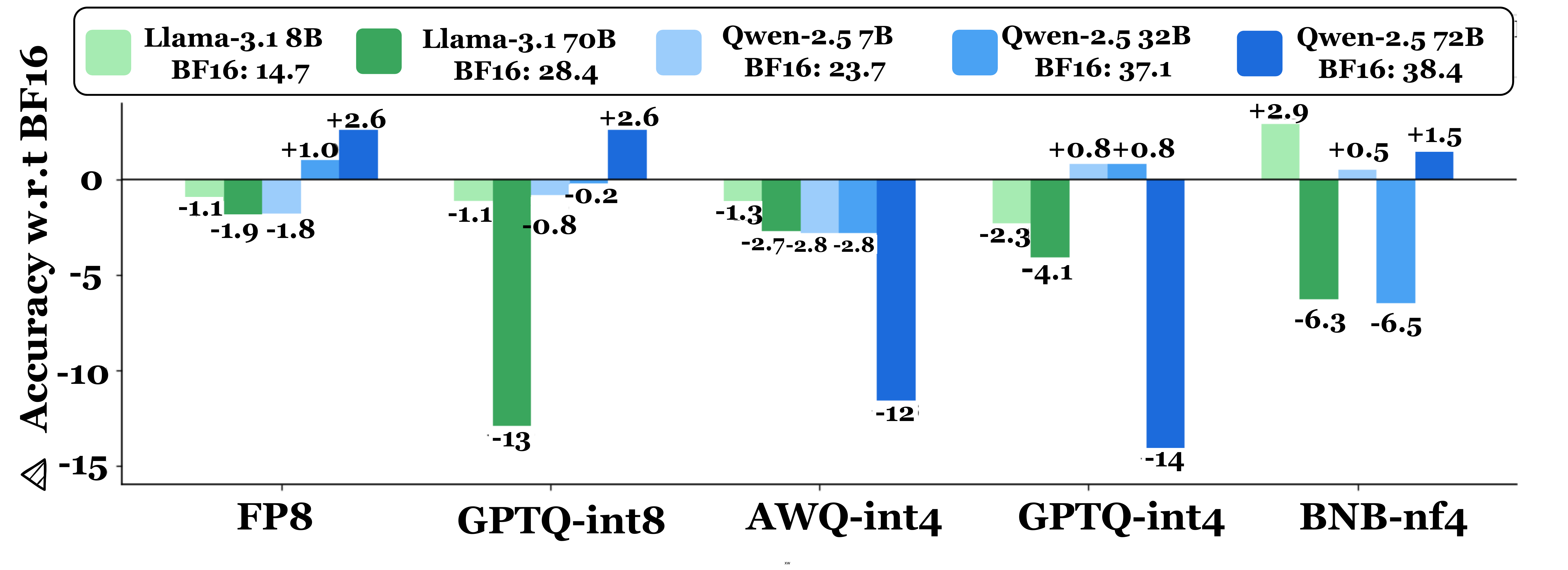}
    \vspace{-2pt}
   \caption{Average $\Delta$-accuracy of different quantization methods relative to BF16 on the \nocha{} benchmark across models when context length $>$75k.}
   \label{fig:nocha-long-model}
\end{figure*}

\begin{figure}[t] % 'h' means here
    \centering
    \includegraphics[width=\linewidth]{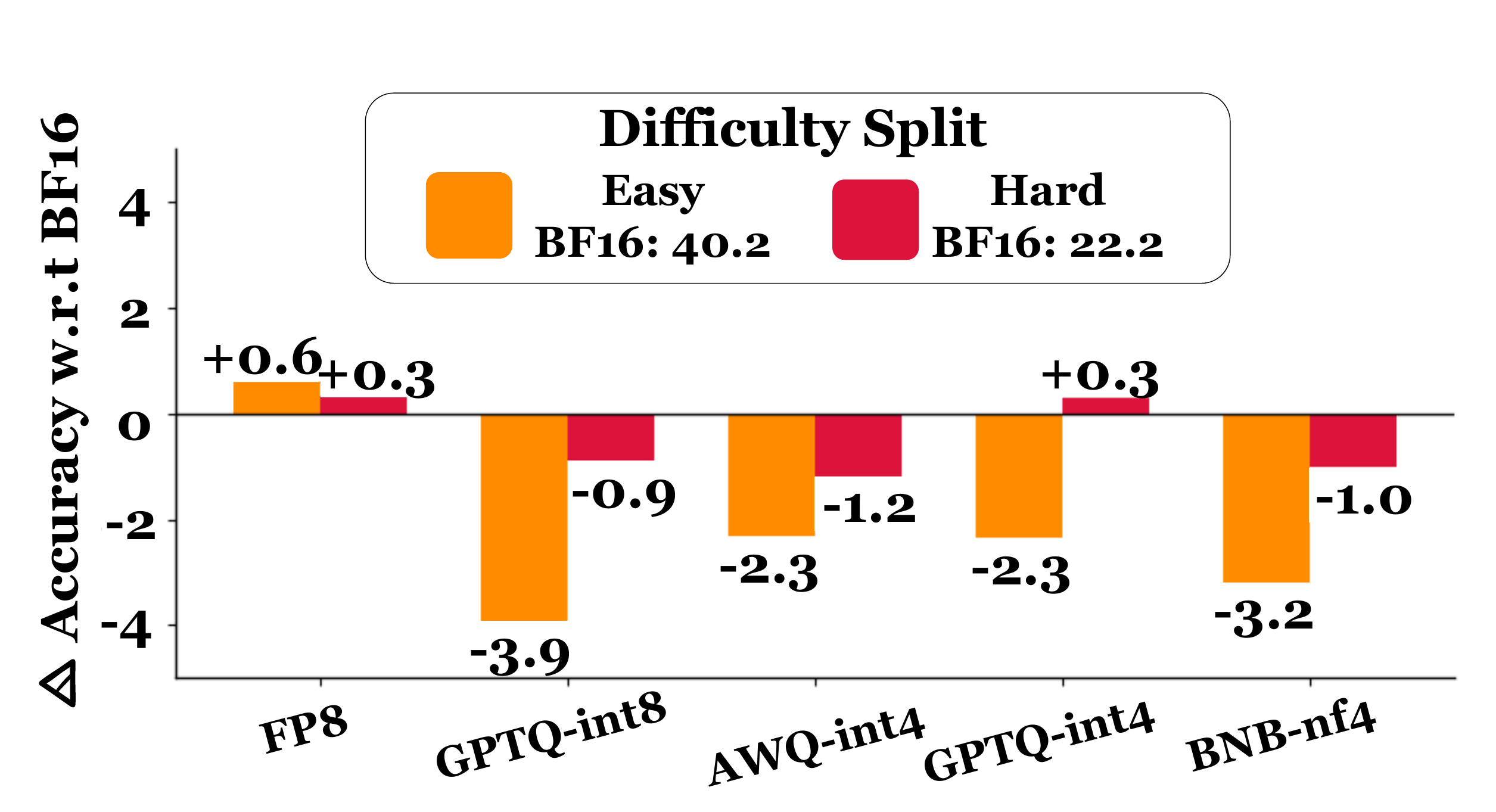}
    %\vspace{-2pt}
    \caption{Average $\Delta$-accuracy of different quantization methods relative to BF16 on the \nocha{} benchmark with varying difficulty level.}
    \label{fig:nocha-difficulty-split}
\end{figure}

\begin{figure*}[t] 
    \centering
    \includegraphics[width=\linewidth]{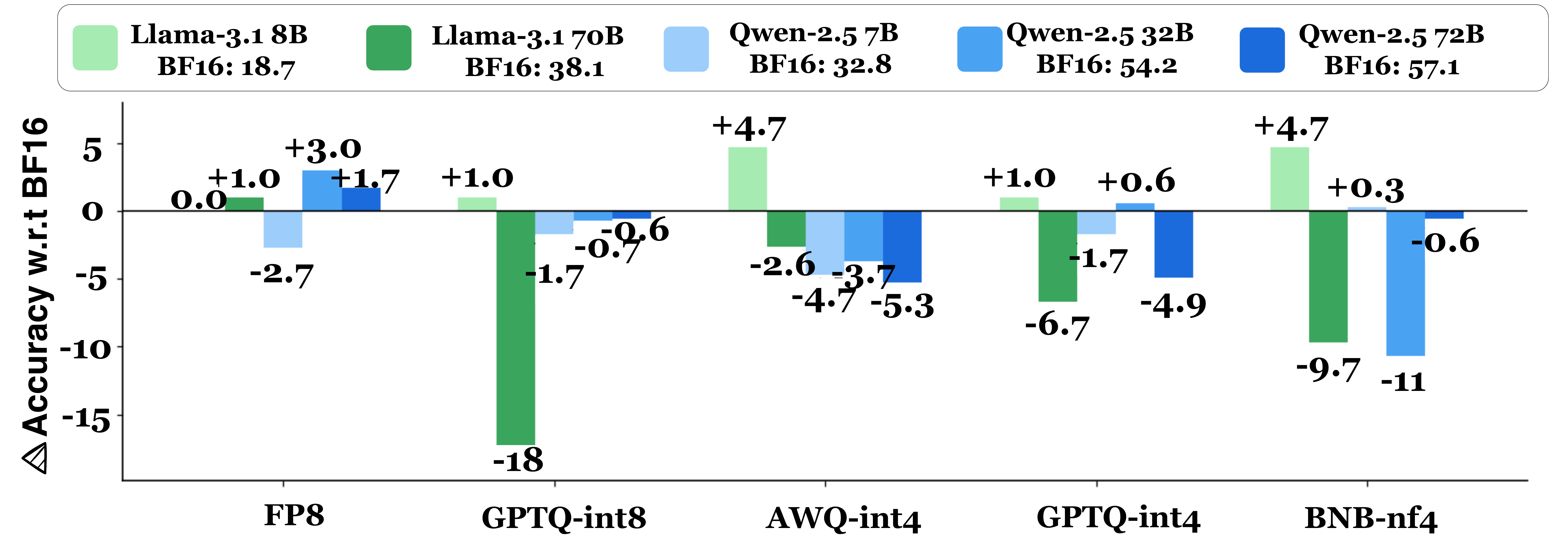}
    %\vspace{-2pt}
    \caption{Average $\Delta$-accuracy of different quantization methods relative to BF16 on the \textit{easy} split of \nocha{} across models.}
    \label{fig:nocha-easy-model}
\end{figure*}

\begin{figure*}[t] % 'h' means here
    \centering
    \includegraphics[width=\linewidth]{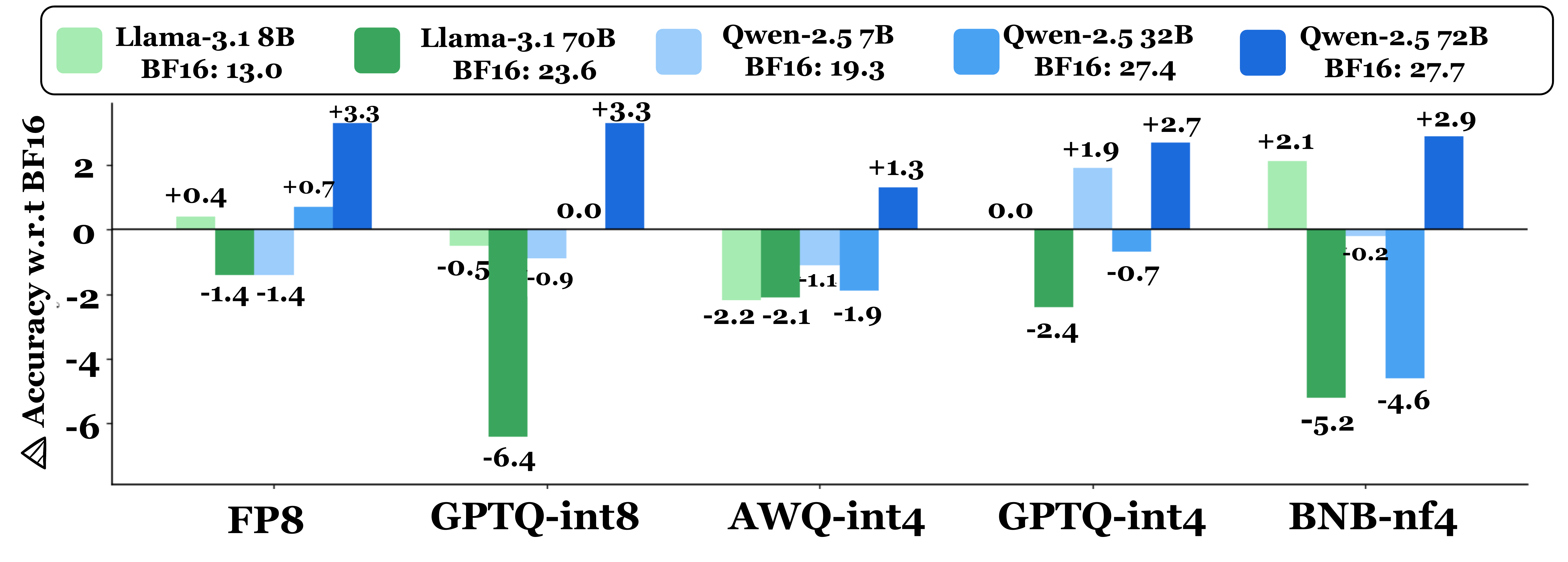}
    %\vspace{-2pt}
    \caption{Average $\Delta$-accuracy of different quantization methods relative to BF16 on the \textit{hard} split of \nocha{} across models.}
    \label{fig:nocha-hard-model}
\end{figure*}

\noindent\textbf{Performance drops become steeper beyond 75k context length:}
\Cref{fig:nocha-context-length} shows that most quantization methods undergo a steeper performance drop compared to BF16 when context length increases from below 75k to above 75k. \cref{fig:nocha-short-model,fig:nocha-long-model} take a closer look at \Cref{fig:nocha-context-length} by analyzing the performance of each quantization method across different models at varying context lengths. At context lengths below 75k (\cref{fig:nocha-short-model}), the effects of quantization vary arbitrarily across models. While GPTQ-int4 and BNB-nf4 generally underperform compared to BF16, \cref{fig:nocha-short-model} shows that all quantization methods—except AWQ-int4—outperform BF16 on \llama{} 8B. These trends stabilize as the context length increases beyond 75k (\cref{fig:nocha-long-model}). Although some quantization methods achieve up to 3\% better performance than their BF16 counterparts, most configurations show relative drops. We observe performance drops of 13\% for GPTQ-int8 on \llama{} 70B, 12\% for AWQ-int4 on \qwen{} 72B, and 14\% for GPTQ-int4 on \qwen{} 72B.\\\noindent\textbf{Quantization-induced degradation does not consistently align with task difficulty:} \cref{fig:nocha-difficulty-split} shows the average $\Delta$-accuracy of different quantization methods relative to BF16 across the \textit{easy} and \textit{hard} splits of \nocha. Together with \cref{fig:nocha-easy-model,fig:nocha-hard-model}, \cref{fig:nocha-difficulty-split} reiterates the observation from \S\ref{sec:results} that the magnitude of performance degradation due to quantization does not directly correlate with task difficulty. While \cref{fig:nocha-difficulty-split} may suggest that increased task difficulty leads to reduced performance degradation, \cref{fig:nocha-easy-model,fig:nocha-hard-model} show that this trend is arbitrary across models. Specifically for \nocha, the BF16 baseline values in \cref{fig:nocha-difficulty-split,fig:nocha-easy-model,fig:nocha-hard-model} are close to chance level, making any small gain or drop around these values difficult to interpret meaningfully. %Finally, assuming sentence-level reasoning to be the easiest and passage-level reasoning the hardest, \cref{fig:nocha-reasoning-len} further supports this claim by showing that the performance of different quantization methods vary differently with different reasoning lengths.
\\\noindent\textbf{Quantization generally preserves model bias, but effects vary across methods and architectures:} Each datapoint in \nocha{} consists of narrative-minimal claim pairs labeled as \textit{true} and \textit{false}, with points awarded only when a model correctly labels both claims within a pair. Analyzing the results for each label separately allows us to assess whether models exhibit bias toward one label. An unbiased model should perform similarly on both \textit{true} and \textit{false} claims; thus, the performance difference between them reflects the model’s bias. We observed that all models predicted \textit{false} more frequently than \textit{true}. On average, across models, \cref{fig:nocha-bias} shows that different quantization methods tend to preserve the bias of their BF16 counterparts, except for AWQ-int4 that amplifies the bias. 


%\begin{figure}[t] % 'h' means here
%   \centering
%   \includegraphics[width=\linewidth]{images/nocha_lenwise.pdf}
%    \vspace{-2pt}
%   \caption{Average $\Delta$-accuracy of different quantization methods relative to BF16 on the \nocha{} benchmark with varying evidence scope.}
%   \label{fig:nocha-reasoning-len}
%\end{figure}

%\begin{figure*}[t] % 'h' means here
%    \centering
%    \includegraphics[width=\linewidth]{images/Nocha_Bias_Model.pdf}
    %\vspace{-2pt}
%    \caption{Average $\Delta$-bias of different quantization methods relative to BF16 on the \nocha{} benchmark across models.}
%    \label{fig:nocha-bias-model}
%\end{figure*}

\subsection{\factscore{}}
\label{appsec:factscore}

\begin{figure}[t]
  \centering
  \includegraphics[width=\linewidth]{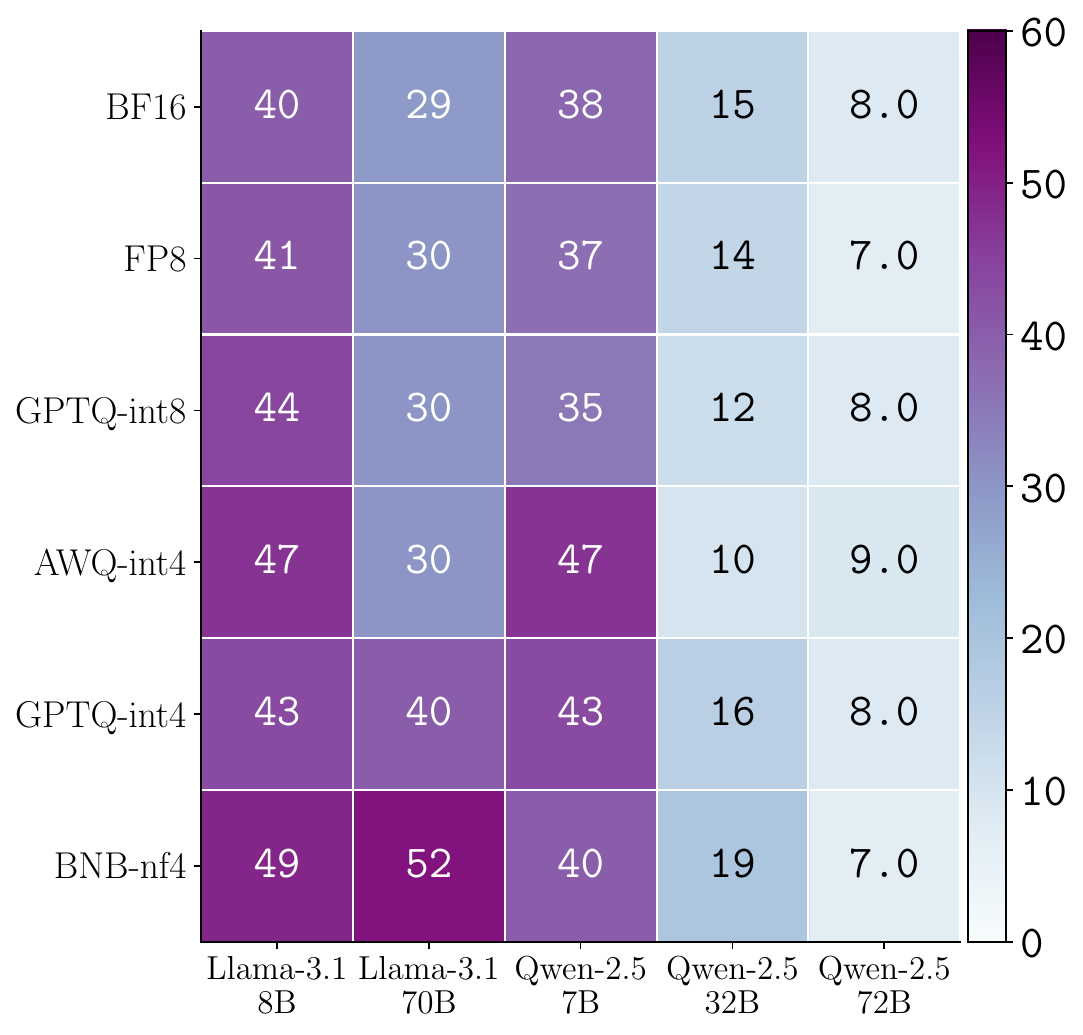}
  \caption{Abstain-rate (rate of responses without verifiable claims) on \factscore{}.}
  \label{fig:veriscore-noclaims-heatmap}
\end{figure}

\autoref{fig:veriscore_raritywise} shows the trends in \veriscore{} as it varies across quantization methods with the frequency of the input entity. \autoref{fig:veriscore_modelwise} shows how \veriscore{} varies across quantization methods with the different models, while \autoref{fig:veriscore-noclaims-heatmap} shows the rate of responses without verifiable claims on \factscore.

\begin{figure}[t] % 'h' means here
    \centering
    \includegraphics[width=\linewidth]{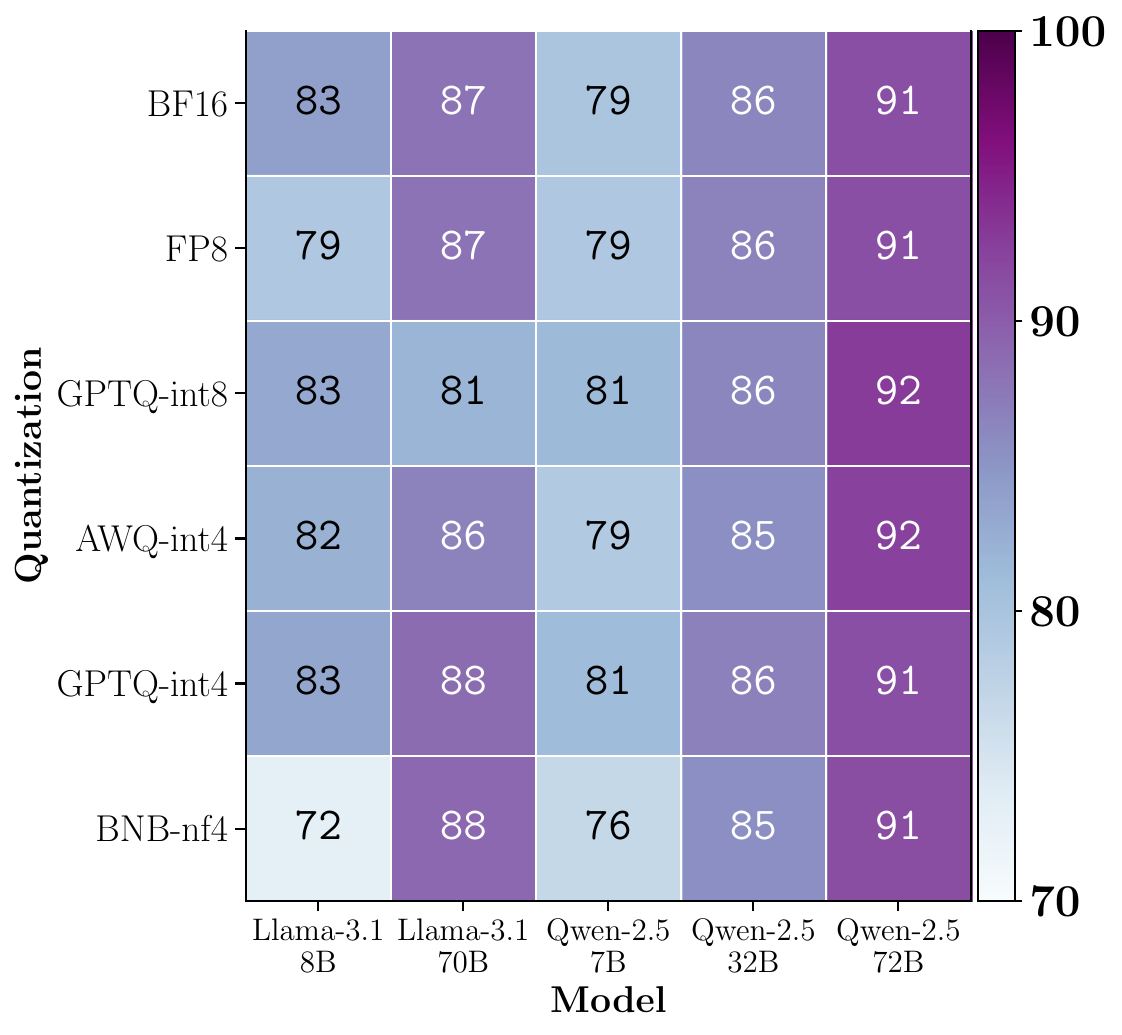}
    %\vspace{-2pt}
    \caption{Average constraint-satisfaction scores of different model-quantization pairs on \csfour.}
    \label{fig:cs-heatmap}
\end{figure}

\subsection{\csfour{}}
\label{appsec:csfour}
\noindent\textbf{Prompting the judge model with fewer constraints improves agreement with human evaluations:} \Cref{fig:cs-prompt} (\S\ref{app:sec-benchmarks}) shows the prompt used to evaluate constraint satisfaction for stories generated using the \csfour{} benchmark. Although each story-writing input can have up to 39 constraints, the judge model is prompted with only three constraints at a time, and the results are then aggregated. To assess whether this improves reliability, we conducted a meta-evaluation on 42 generations—21 where the judge model was prompted with 3 constraints at a time, and 21 where all constraints were shown at once. The 21 stories were evenly split across the three levels of input constraints. These 42 story–constraint pairs were manually annotated by one of the authors, who independently read each story and assessed which constraints were satisfied. Agreement scores reported in \cref{tab:cs-eval-agreement} indicate how many of the judge model’s evaluations matched the human annotations. Prompting with three constraints at a time resulted in closer agreement with human assessments. %\\\noindent\textbf{FP8 quantization shows a significant drop on the instruction-following ability of models:} In \Cref{fig:all_taskwise}, we compute the $\Delta$-accuracy relative to BF16 for each quantization method, averaged across models and data points. Along with \Cref{fig:cs-barplot,fig:cs-heatmap}, this figure further confirms that constraint satisfaction is largely unaffected by quantization, as even the worst-performing method in \Cref{fig:all_taskwise} shows only a 1.8\% drop in performance.

\begin{table}[t]
\centering
\resizebox{0.8\linewidth}{!}{%
  \small
  \renewcommand{\arraystretch}{1.1}
  \begin{tabular}{ccc}
    \hline
    \begin{tabular}[c]{@{}c@{}}No. of input \\ constraints for \\ writing the story\end{tabular} & 
    \begin{tabular}[c]{@{}c@{}}No. of constraints \\ given to judge-LLM \\ at a time\end{tabular} & 
    \begin{tabular}[c]{@{}c@{}}Avg. evaluations\\ agreeing with the\\ human\end{tabular} \\
    \hline
    7  & 3   & 6.57  \\
    7  & 7   & 6.57  \\
    23 & 3   & 21.00 \\
    23 & 23  & 19.46 \\
    39 & 3   & 37.28 \\
    39 & 39  & 32.00 \\
    \hline
  \end{tabular}
}
\caption{Agreement between human evaluation and LLM-as-judge under different judge-LLM input sizes. Agreement refers to the number of constraint evaluations (out of the total) where the judge model’s assessment matched human annotations, across varying numbers of input constraints and judge-prompt sizes.}
\label{tab:cs-eval-agreement}
\end{table}

% \begin{figure*}[t] % 'h' means here
%     \centering
%     \includegraphics[width=\linewidth]{images/BooookScore_Model.pdf}
%     \caption{\textbf{\csfour:} Average $\Delta$-BoookScore across models.}
%     \label{fig:BooookScore_modelwise}\textbf{\csfour: Average $\Delta$-BooookScore of different quantization methods relative to BF16 across varying models} 
%     % .
% \end{figure*}

  % \vspace{1em}
  % \begin{subfigure}[b]{0.65\textwidth}
  %   \centering
  %   \includegraphics[width=\textwidth]{images/BooookScore_Model.pdf}
  %   \caption{\textbf{\csfour:} Average $\Delta$-BoookScore across models.}
  %   \label{fig:BooookScore_modelwise}
  % \end{subfigure}
  % \vspace{-5pt}

\clearpage       
\onecolumn       
\section{An example from the \csfour~benchmark}
\label{app:cs4-example}

{\footnotesize
\setlength{\parindent}{0pt}
\ttfamily

\textbf{Input prompt for story writing:} Write a story about an introverted protagonist with uncontrollable intuition, who experiences others' emotions intensely, as she navigates a challenging encounter during the holiday season. The story should satisfy the following constraints: 

\begin{enumerate}
  \item The protagonist suffers physical discomfort when overwhelmed by emotions (nausea, shaking, etc.).
  \item The protagonist is challenged by the need to engage in public spaces.
  \item The unknown man realizes that the protagonist can feel his emotions.
  \item The protagonist uses humor and sarcasm to cope with her situation.
  \item The protagonist is an introverted character.
  \item The story includes communication via text messages.
  \item The story is set in a Starbucks on Michigan in Chicago a week before Christmas.
  \item The protagonist is forced to leave the meeting early due to being overwhelmed.
  \item The protagonist desires to live a more normal life despite her unique condition.
  \item There could exist a vaccine for controlling intuition.
  \item The protagonist devises coping strategies for managing her anxiety in public places.
  \item Scientists are working to find a solution for people who can't use the intuition vaccine.
  \item Tiffany threatens the protagonist to meet her.
  \item The protagonist struggles with accepting her condition.
  \item The protagonist must grapple with the thoughts and feelings of others in the Starbucks.
  \item The man looks at the protagonist with both desire and love.
  \item The protagonist is physically attractive.
  \item The protagonist encounters an unknown man who causes powerful and unique emotions.
  \item There is societal disapproval for people whose intuitions cannot be controlled by the vaccine.
  \item The protagonist has a heightened intuition.
  \item The setting should be during the holiday season.
  \item The protagonist feels other people's emotions intensely.
  \item The protagonist struggles with disentangling their own feelings from others'.
\end{enumerate}

\vspace{1em}
\textbf{Story generated by \llama~8B BF16 model:} In the cozy glow of a late-December morning, under the twinkling lights of a small Starbucks on Michigan Avenue in Chicago, Emily sat alone, sipping a steaming latte. The holiday season was in full swing, and the café was bustling with people. Emily, an introvert with a heightened intuition, felt her heart racing as she scanned the crowd. Her eyes met those of an unknown man across the room. He was tall, with piercing blue eyes and a charming smile. Emily felt a mix of emotions—curiosity, warmth, and a hint of fear. As he looked back, she could sense a deep longing and a touch of desire in him. The intensity of these feelings was overwhelming, causing her stomach to churn and her hands to shake slightly. "Are you okay?" a voice whispered through her mind, breaking the spell. She blinked, trying to focus. It was the man from across the room. He leaned closer, his eyes filled with concern. "You look like you're going to be sick." Emily laughed, a nervous chuckle that sounded hollow even to her ears. "I’m fine," she said out loud, her voice trembling. "Just a bit... overwhelmed." He nodded, understanding. "I get it. I’ve been there." He gestured towards the crowded café. "Want to step outside for a bit? It’s not easy in here." Emily hesitated but then nodded, grateful for his kindness. They stepped into the chilly air, and the man offered her a jacket. "Here, put this on. It’s freezing out here." As they walked, the man began to share stories, his laughter light and infectious. Emily found herself smiling, her anxiety easing slightly. She realized that she could control her reactions better when she engaged in conversation. "So, what brings you to Chicago this time of year?" she asked, trying to keep the conversation flowing. "I'm actually here for a conference," he replied, his eyes never leaving hers. "And you?" "Just visiting," Emily said, her cheeks flushing. She felt a wave of emotions wash over her—joy, admiration, and something she couldn’t quite place. She took a deep breath, her intuition telling her to be careful. Back inside the café, the man excused himself, promising to catch up later. Emily sat down, her mind racing. She had always struggled with her unique condition, feeling the emotions of others intensely and often struggling to separate them from her own. The idea of a vaccine that could control her intuition was appealing, but she feared losing the connection to others. Tiffany, her best friend, had threatened to meet her earlier that day, but Emily had pushed her away. She needed some space to think. Emily pulled out her phone and texted Tiffany, "I’m meeting someone for coffee. Can we talk later?" Tiffany replied almost immediately, "Sure thing. Be safe." Emily sighed, feeling a mix of relief and apprehension. She knew she had to find a way to live a more normal life, one where she could navigate public spaces without being overwhelmed. She decided to reach out to a support group for people like her, hoping to find others who understood her struggle. As the sun began to set, casting a warm golden light over the city, Emily gathered her things and left the café. She knew the road ahead would be challenging, but she also knew that she wasn’t alone. With each step, she felt a glimmer of hope, determined to find a way to live a more balanced life, one where her unique condition didn’t define her. 

\vspace{1em}
\textbf{Evaluation for constraint satisfaction using the prompt in \cref{fig:cs-prompt}:}
\begin{enumerate}
  \item Yes – “The intensity of these feelings was overwhelming, causing her stomach to churn and her hands to shake slightly.” —the story shows clear nausea and shaking when emotions peak, fully meeting the requirement for physical discomfort.
  \item Yes – “The holiday season was in full swing, and the café was bustling with people.” / “She knew she had to find a way to live a more normal life, one where she could navigate public spaces without being overwhelmed.” These lines depict a crowded public setting that stresses her and a stated struggle with such spaces.
  \item No – At no point does the unknown man discover that Emily can literally feel his emotions; he only notices that she looks unwell. There is no dialogue or narration indicating his realization of her empathic ability, so the constraint is unmet.
  \item No – Although Emily gives a nervous chuckle and says, “I’m fine… just a bit overwhelmed,” this is neither genuine humor nor sarcasm used as a coping mechanism. The story lacks any witty, humorous, or sarcastic remarks from her to manage the situation, so the criterion is not satisfied.
  \item Yes – “Emily, an introvert with a heightened intuition, felt her heart racing as she scanned the crowd.” The narrative explicitly identifies the protagonist as an introvert, fully fulfilling the constraint.
  \item Yes – “Emily pulled out her phone and texted Tiffany, ‘I’m meeting someone for coffee. Can we talk later?’ Tiffany replied almost immediately, ‘Sure thing. Be safe.’” — these two lines show a back-and-forth text exchange, satisfying the communication-via-text requirement.
  \item No – The narrative states only that events occur “in the cozy glow of a late-December morning, under the twinkling lights of a small Starbucks on Michigan Avenue in Chicago.” It never specifies “a week before Christmas.” Late December could be before or after Christmas, so the exact timing demanded by the constraint is missing.
  \item No – Although Emily briefly steps outside to get fresh air, she later returns and ultimately leaves the café at sunset of her own accord. The story never shows her being compelled to end the encounter prematurely because of overwhelm (e.g., cutting the meeting short, fleeing abruptly), so the requirement of being “forced to leave the meeting early” is not met.
  \item Yes – “She knew she had to find a way to live a more normal life, one where she could navigate public spaces without being overwhelmed.” — this sentence directly expresses her desire for a normal life despite her extraordinary intuition.
  \item Yes – “The idea of a vaccine that could control her intuition was appealing, but she feared losing the connection to others.” — this line explicitly introduces the existence of a vaccine designed to dampen or manage her intuitive ability, fulfilling the constraint.
  \item Yes – She actively develops ways to stay calm: “She realized that she could control her reactions better when she engaged in conversation,” and later, “She decided to reach out to a support group for people like her.” These lines show deliberate coping strategies for public-place anxiety.
  \item No – While a vaccine to blunt intuition is mentioned, the story never says that scientists are pursuing an alternative for people who cannot take it; no research effort is described.
  \item Yes – “Tiffany, her best friend, had threatened to meet her earlier that day, but Emily had pushed her away.” —the verb “threatened” makes the demand explicit.
  \item Yes – “She had always struggled with her unique condition, feeling the emotions of others intensely and often struggling to separate them from her own.” This directly shows her difficulty accepting and living with the condition.
  \item Yes – “Emily, an introvert with a heightened intuition, felt her heart racing as she scanned the crowd.” / “She could sense a deep longing and a touch of desire in him.” These sentences depict her grappling with others’ thoughts and feelings inside the Starbucks.
  \item No – The narrative states the man’s “deep longing and a touch of desire,” but it never attributes love to him. Because love is absent, the condition is unmet.
  \item No – Nowhere is Emily’s physical attractiveness described or implied; the story focuses on her inner state, not her appearance.
  \item Yes – “Her eyes met those of an unknown man across the room… She felt a mix of emotions—curiosity, warmth, and a hint of fear.” The unknown man clearly triggers powerful, unique feelings in her, satisfying the constraint.
  \item No – The story gives no indication of societal disapproval toward people whose intuition cannot be controlled; no stigma or judgment is depicted.
  \item Yes – “Emily, an introvert with a heightened intuition…” —the phrase explicitly confirms her heightened intuitive ability.
  \item Yes – “The holiday season was in full swing” and the date is “late-December”; the setting is clearly during the holiday season.
  \item Yes – “She had always struggled with her unique condition, feeling the emotions of others intensely…” —this line directly states that she experiences other people’s emotions intensely.
  \item Yes – “…often struggling to separate them from her own.” The narrative explicitly describes her difficulty disentangling her feelings from others’, fully satisfying the constraint.
\end{enumerate}

}
\clearpage        
\twocolumn
\normalfont
\clearpage       
\onecolumn       
\section{An example from the \factscore{} benchmark}
\label{appsec:factscore-example}

{\footnotesize
\setlength{\parindent}{0pt}
\ttfamily

\textbf{Input prompt:} Write the biography of Fumiko Kaneko.

\vspace{1em}
\textbf{Response:} Fumiko Kaneko (1903-1926) was a Japanese anarchist and nihilist who gained notoriety for her involvement in a plot to assassinate the Japanese Emperor Hirohito and other high-ranking government officials.\\

Born in 1903 in Aomori Prefecture, Japan, Kaneko grew up in a poor farming family. Her early life was marked by hardship and struggle, which may have contributed to her radicalization. In her late teens, she moved to Tokyo, where she became involved in the city's anarchist and nihilist circles.\\

In 1923, Kaneko met Pak Yeol, a Korean anarchist who had fled to Japan after being involved in the March 1st Movement, a Korean independence movement against Japanese colonial rule. The two became romantically involved and began to plan a series of bombings and assassinations targeting Japanese government officials and the royal family.\\

On January 9, 1926, Kaneko and Pak Yeol were arrested by Japanese police while attempting to plant a bomb near the Imperial Palace in Tokyo. The plot was foiled, and the two were charged with treason and attempted murder.\\

During her trial, Kaneko showed no remorse for her actions, stating that she had acted out of a desire to overthrow the Japanese government and monarchy, which she saw as oppressive and corrupt. Her statements were seen as a challenge to the authority of the Japanese state, and she became a symbol of resistance against the government.\\

Kaneko was sentenced to death and executed by hanging on March 23, 1926, at the age of 23. Pak Yeol was also executed on the same day. Their executions sparked widespread outrage and protests among anarchist and left-wing groups in Japan, who saw them as martyrs for the cause of revolution and social change.\\

Despite her short life and tragic end, Fumiko Kaneko's legacy as a radical activist and symbol of resistance has endured in Japan and beyond. Her story has inspired numerous books, films, and plays, and she remains a fascinating and complex figure in Japanese history.

}

\clearpage        
\twocolumn
\normalfont
\clearpage    
\section{Statistical tests of significance for comparing quantization methods}
\label{app:stats-test}

We conducted statistical analysis by fitting the generalized linear mixed-effects models (GLMMs) using the \texttt{glmer()} function from the \texttt{lme4} package in R \cite{bates2015fitting} to study if different quantization methods performed significantly differently from their BF16 counterparts. For improved convergence, we used a \textit{binomial logit link} and the \texttt{bobyqa} optimizer. We formatted our data in \texttt{long} format, with each model response represented as a binary categorical variable indicating whether the prediction was correct or incorrect. Quantization methods were included as the primary fixed effect of interest, with additional fixed effects for context length, task, and their interactions in different model variants. 

For each GLMM we report marginal $R^2$, the variance explained by fixed effects alone, and conditional $R^2$, the variance explained by fixed and random effects \cite{nakagawa2017coefficient}. We also list the variance components ($\sigma^2$, $\tau_{00}$) and intra-class correlation coefficients (ICC) for every grouping factor. Other aspects like task, context length, and language are added where appropriate to account for the non-independence of repeated measures and to allow the fixed effects of interest to be estimated after partialing out systematic differences due solely to model identity, task, or input length.

Following \cite{karpinska2024one}, we conducted pairwise comparisons of quantization levels with emmeans for every GLMM. \texttt{p-value}s were Bonferroni-adjusted \cite{westfall1997bayesian} and reported in the post-hoc tables. Log-odds estimates were exponentiated to odds ratios, which were then converted to probabilities to facilitate interpretation (a probability of 0.50 means the two methods are equally accurate).

To examine how quantization interacts with context length and task type within \ruler{}, we fit three GLMMs with \texttt{Model} as a random intercept. The main-effect model (\texttt{Binary\_label $\sim$ Quantization + (1|Model)}) tests whether any quantization method differs from BF16 (Table~\ref{tab:ruler-stats-model-glmm}; pairwise contrasts in Table~\ref{tab:ruler-posthoc-model}). The interaction model with context length (\texttt{Binary\_label $\sim$ Quantization * Context\_length + (1|Model) + (1|Task)}) evaluates whether quantization effects vary across 8k, 64k, and 128k tokens while controlling for task variability (Table~\ref{tab:ruler-stats-context_Len}; contrasts in Table~\ref{tab:ruler-posthoc-context}). The task interaction model (\texttt{Binary\_label $\sim$ Quantization * Task + (1|Model)}) compares performance of quantized models on multi-key and multi-value tasks on \ruler{}, using BF16 on needle as the reference (Table~\ref{tab:ruler-task-interaction}; contrasts in Table~\ref{tab:ruler-posthoc-task}). Across these models, marginal $R^2$ increases from .013 (quantization only) to .615 (quantization * context-length), while conditional $R^2$ peaks at .488, indicating that both fixed effects and grouping factors jointly capture a sizeable proportion of variance on \ruler{}. We conduct the same statistical analyses with \oneruler as in \ruler{} (\cref{tab:oneruler-stats-model-glmm,tab:oneruler-posthoc-model,tab:oneruler-context,tab:oneruler-posthoc-context,tab:oneruler-frequency-interaction,tab:oneruler-posthoc-frequency}), with the only change being the use of \textit{language frequency} instead of \textit{task}. 

\nocha{} analyses parallel the \ruler{} tests but center on the 75k-token split defined in the benchmark. The main-effect model (\texttt{Binary\_label $\sim$ Quantization + (1|Context\_length) + (1|Model)}) includes quantization as the predictor, with \texttt{Context\_length} and \texttt{Model} as random intercepts (Table~\ref{tab:nocha-stats}); pairwise contrasts appear in Table~\ref{tab:nocha-posthoc-model}. To test whether quantization errors increase for longer inputs, we also fit an interaction model (\texttt{Binary\_label $\sim$ Quantization * Context\_length + (1|Model)}), using BF16 at $\leq$75k tokens as the reference (Table~\ref{tab:nocha-context-split}; contrast results are reported in Table~\ref{tab:nocha-posthoc-context75k}). Random-effects variances are small for \nocha{} (\textit{ICC}$_\text{Model} \approx 0.045$), and both marginal and conditional $R^2$ values are low ($\approx$~0.016), indicating the dataset’s overall difficulty. 

To evaluate how quantization impacts performance on \factscore{}, we fit three linear and generalized linear mixed-effects models with \texttt{Model} as a random intercept. The main-effect model (\texttt{log\_veriscore $\sim$ Quantization + (1|topic) + (1|Model)}) estimates whether quantized models deviate from the BF16 baseline in overall generation quality (Table~\ref{tab:veriscore-quant-main}; pairwise contrasts in Table~\ref{tab:veriscore-quant-posthoc}). To assess no claim rate, we model the likelihood of no claims (\texttt{glmer(no\_claims $\sim$ Quantization + (1|Model) + (1|topic))}) in Table~\ref{tab:veriscore-empty_claims-main} with contrasts in Table~\ref{tab:veriscore-empty_claims-posthoc}. Lastly, we examine whether quantization interacts with category (\texttt{log\_veriscore $\sim$ Quantization * cat + (1|Model) + (1|topic)}) by modeling \veriscore{} across medium, very frequent, and very rare categories (Table~\ref{tab:veriscore-category-main}; post-hoc contrasts in Table~\ref{tab:veriscore-category-posthoc}).

\Cref{fig:cs-heatmap} in \S\ref{appsec:csfour} plots the absolute constraint satisfaction scores of each model-quantization pair. Here, we fit GLMMs to evaluate how quantization influences constraint satisfaction on \csfour{}. However, each data point here reflects the percentage of constraints satisfied, i.e., a success proportion, rather than a binary label. Since this can only take a finite set of values, we use a binomial GLMM with \texttt{cbind}. The main-effect model (\texttt{cbind(Success, Failures) $\sim$ Quantization + (1|Model)}) tests whether quantized models differ from BF16 in satisfying constraints (Table~\ref{tab:csfour-stats-model-quant}; pairwise contrasts in Table~\ref{tab:cs4-posthoc-model}). To analyze whether quantization interacts with the number of input constraints, we introduce input-level interactions in (\texttt{cbind(Success, Failures) $\sim$ Quantization * Input\_constraints + (1|Model)}), allowing us to track performance across 7, 23, and 39 input constraints (Table~\ref{tab:cs4-interact}; post-hoc results in Table~\ref{tab:cs4-interact-posthoc}).

\begin{table*}[ht]
\centering
\resizebox{0.7\textwidth}{!}{
\begin{tabular}{l
                S[table-format=1.3]
                S[table-format=1.3]
                S[table-format=1.3]
                S[table-format=1.3]
                S[table-format=1.5]
                c}
\toprule
\multicolumn{7}{l}{\texttt{glmer(Binary\_label $\sim$ Quantization + (1|Model), data = data, family = binomial)}} \\
\bottomrule
\textsc{Predictors} & \textsc{Estimate} & \textsc{OR} & \textsc{CI (lower)} & \textsc{CI (upper)} & \texttt{p-value} & \\
\midrule
Intercept            &  1.336 & 3.802 & 2.182 & 6.623 & 0.000 & *** \\
FP8                  & -0.144 & 0.866 & 0.783 & 0.957 & 0.050  &  \\
GPTQ-int8            & -0.042 & 0.959 & 0.867 & 1.062 & 0.42140  &     \\
AWQ-int4             & -0.158 & 0.854 & 0.772 & 0.944 & 0.00197  & **  \\
GPTQ-int4            & -0.365 & 0.694 & 0.629 & 0.766 & 0.000 & *** \\
BNB-nf4              & -0.647 & 0.524 & 0.475 & 0.577 & 0.000   & *** \\
\bottomrule
\textsc{Random Effects} & & & & & & \\
\bottomrule
\\
$\sigma^2$ (residual variance) & \multicolumn{5}{l}{1.000} & \\
$\tau_{00}$ (Model) & \multicolumn{5}{l}{0.408} & \\
ICC (Model) & \multicolumn{5}{l}{0.290} & \\
N (Model) & \multicolumn{5}{l}{5} & \\
Observations & \multicolumn{5}{l}{27000} & \\
\bottomrule
\\
$R^2$ (marginal) & \multicolumn{5}{l}{0.013} & \\
$R^2$ (conditional) & \multicolumn{5}{l}{0.122} & \\
\bottomrule
\end{tabular}
}
\caption{\textbf{\ruler{}}: Summary of generalized linear mixed model with \textbf{quantization} as the predictor of \textbf{accuracy}: \texttt{glmer(Binary\_label $\sim$ Quantization + (1|Model), data = data, family = binomial)}. The quantization method "BF16" was set as the reference level (intercept), with \texttt{Model} set as the grouping factor for a random intercept to account for variability across models. See Table~\ref{tab:ruler-posthoc-model} for post-hoc analysis.}
\label{tab:ruler-stats-model-glmm}
\vspace{0.5em}
\end{table*}

\begin{table*}[ht]
\centering
\resizebox{0.7\textwidth}{!}{
\begin{tabular}{l
                S[table-format=1.3]
                S[table-format=1.3]
                S[table-format=1.3]
                S[table-format=1.3]
                S[table-format=1.3]
                c}
\toprule
\textsc{Contrast} & \textsc{Estimate} & \textsc{SE} & \textsc{OR} & \textsc{Probability} & \texttt{p-value} & \\
\midrule
BF16 - FP8                 &  0.144 & 0.051 & 1.155 & 0.536 & 0.072 &     \\
BF16 - (GPTQ-int8)         &  0.042 & 0.052 & 1.043 & 0.510 & 1.00e+00 &     \\
BF16 - (AWQ-int4)          &  0.158 & 0.051 & 1.171 & 0.539 & 0.029 & *   \\
BF16 - (GPTQ-int4)         &  0.365 & 0.050 & 1.441 & 0.590 & 0.000 & *** \\
BF16 - (BNB-nf4)           &  0.647 & 0.049 & 1.910 & 0.656 & 0.000 & *** \\

(AWQ-int4) - FP8           & -0.014 & 0.050 & 0.986 & 0.497 & 1.00e+00 &     \\
(AWQ-int4) - (GPTQ-int8)   & -0.117 & 0.051 & 0.890 & 0.471 & 0.330 &     \\
(AWQ-int4) - (GPTQ-int4)   &  0.207 & 0.049 & 1.230 & 0.552 & 0.000 & *** \\
(AWQ-int4) - (BNB-nf4)     &  0.489 & 0.048 & 1.631 & 0.620 & 0.000 & *** \\

FP8 - (GPTQ-int4)          &  0.221 & 0.049 & 1.247 & 0.555 & 0.000 & *** \\
FP8 - (GPTQ-int8)          & -0.103 & 0.051 & 0.902 & 0.474 & 0.661 &     \\
FP8 - (BNB-nf4)            & -0.503 & 0.048 & 0.605 & 0.377 & 0.000 & *** \\

(GPTQ-int4) - (GPTQ-int8)  & -0.324 & 0.050 & 0.723 & 0.420 & 0.000 & *** \\
(GPTQ-int4) - (BNB-nf4)    & -0.282 & 0.047 & 0.754 & 0.430 & 0.000 & *** \\

(GPTQ-int8) - (BNB-nf4)    & -0.606 & 0.049 & 0.546 & 0.353 & 0.000 & *** \\
\bottomrule
\end{tabular}
}
\caption{\textbf{\ruler{}}: Post-hoc comparisons between quantization methods for \textbf{accuracy} (Table~\ref{tab:ruler-stats-model-glmm}) using Bonferroni adjustments for multiple pairwise comparisons. The probability values refer to the likelihood that the quantization method listed first in each contrast is more accurate than the one listed second (i.e., a value of 0.5 suggests that both methods are comparable in terms of accuracy).}
\label{tab:ruler-posthoc-model}
\vspace{0.5em}
\end{table*}

\begin{table*}[ht]
\centering
\resizebox{0.7\textwidth}{!}{%
\begin{tabular}{l
                S[table-format=2.6]
                S[table-format=1.6]
                S[table-format=1.6]
                S[table-format=1.6]
                S[table-format=1.6]
                c}
\toprule
\multicolumn{7}{c}{
\begin{tabular}{c}
\texttt{glmer(Binary\_label $\sim$ Quantization * Context\_length + (1|Model)} \\
\texttt{+ (1|Task), data = data, family = binomial)}
\end{tabular}
} \\
\bottomrule
\textsc{Predictors} & \textsc{Estimate} & \textsc{OR} & \textsc{CI (lower)} & \textsc{CI (upper)} & \texttt{p-value} & \\
\midrule
Intercept                               &  5.576814 & 264.228358 &  85.366506 & 817.845909 & 0.000 & *** \\
AWQ-int4                    & -0.813585 &   0.443266 &   0.215988 &   0.909702 & 0.026557 & * \\
BNB-nf4                     & -0.760965 &   0.467216 &   0.227141 &   0.961035 & 0.038643 & * \\
FP8                         & -1.480385 &   0.227550 &   0.117791 &   0.439583 & 0.000 & *** \\
GPTQ-int4                   & -1.507703 &   0.221418 &   0.114635 &   0.427670 & 0.000 & *** \\
GPTQ-int8                   & -0.705495 &   0.493864 &   0.237760 &   1.025831 & 0.058546 & \\
64k                      & -4.318882 &   0.013315 &   0.007365 &   0.024072 & 0.000 & *** \\
128k                     & -5.164246 &   0.005717 &   0.003168 &   0.010318 & 0.000 & *** \\
AWQ-int4:64k  &  0.756742 &   2.131320 &   1.018549 &   4.459800 & 0.044564 & * \\
BNB-nf4:64k   &  0.061311 &   1.063229 &   0.507867 &   2.225890 & 0.870801 & \\
FP8:64k       &  1.427561 &   4.168521 &   2.113363 &   8.222235 & 0.000 & *** \\
GPTQ-int4:64k &  1.169936 &   3.221785 &   1.634713 &   6.349676 & 0.000726 & *** \\
GPTQ-int8:64k &  0.693220 &   2.000145 &   0.944865 &   4.234023 & 0.070026 & \\
AWQ-int4:128k &  0.496279 &   1.642598 &   0.788071 &   3.423712 & 0.185375 & \\
BNB-nf4:128k  & -0.387598 &   0.678685 &   0.324766 &   1.418293 & 0.302682 & \\
FP8:128k      &  1.254862 &   3.507354 &   1.784755 &   6.892559 & 0.000272 & *** \\
GPTQ-int4:128k&  0.932412 &   2.540630 &   1.293009 &   4.992077 & 0.006817 & ** \\
GPTQ-int8:128k&  0.641986 &   1.900252 &   0.901099 &   4.007283 & 0.091718 & \\
\bottomrule
\textsc{Random Effects} & & & & & & \\
\bottomrule
\\
$\sigma^2$ (residual variance) & \multicolumn{5}{l}{1.000} & \\
$\tau_{00}$ (Model) & \multicolumn{5}{l}{0.807} & \\
$\tau_{00}$ (Task) & \multicolumn{5}{l}{0.281} & \\
ICC (Model) & \multicolumn{5}{l}{0.386} & \\
ICC (Task) & \multicolumn{5}{l}{0.134} & \\
N (Model) & \multicolumn{5}{l}{5} & \\
N (Task) & \multicolumn{5}{l}{3} & \\
Observations & \multicolumn{5}{l}{27000} & \\
\bottomrule
\\
$R^2$ (marginal) & \multicolumn{5}{l}{0.615} & \\
$R^2$ (conditional) & \multicolumn{5}{l}{0.488} & \\
\bottomrule
\end{tabular}
} 
\caption{\textbf{\ruler{}}: Summary of generalized linear mixed model with \textbf{quantization}, \textbf{context length}, and their interaction as predictors of \textbf{accuracy}: \texttt{glmer(Binary\_label $\sim$ Quantization * Context\_length + (1|Model) + (1|Task), data = data, family = binomial)}. The quantization method "BF16" at context length 8k was set as the reference level (intercept), with \texttt{Model} and \texttt{Task} set as the grouping factors for random intercepts to account for variability across models and tasks. See Table~\ref{tab:ruler-posthoc-context} for post-hoc analysis.}

\vspace{0.5em}
\label{tab:ruler-stats-context_Len}
\end{table*}

\begin{table*}[ht]
\centering
\resizebox{0.7\textwidth}{!}{
\begin{tabular}{l
                l
                S[table-format=1.3]
                S[table-format=1.3]
                S[table-format=1.3]
                S[table-format=1.3]
                S[table-format=1.3]
                c}
\toprule
\textsc{Context} & \textsc{Contrast} & \textsc{Estimate} & \textsc{SE} & \textsc{OR} & \textsc{Probability} & \texttt{p-value} & \\
\midrule
8k   & BF16 - (AWQ-int4)           & 0.814 & 0.367 & 2.256 & 0.693 & 0.398 & \\
8k   & BF16 - (BNB-nf4)            & 0.761 & 0.368 & 2.140 & 0.682 & 0.580 & \\
8k   & BF16 - FP8                  & 1.480 & 0.336 & 4.395 & 0.815 & 0.000 & *** \\
8k   & BF16 - (GPTQ-int4)          & 1.508 & 0.336 & 4.516 & 0.819 & 0.000 & *** \\
8k   & BF16 - (GPTQ-int8)          & 0.705 & 0.373 & 2.025 & 0.669 & 0.878 & \\
8k   & (AWQ-int4) - (BNB-nf4)      & -0.053 & 0.320 & 0.949 & 0.487 & 1.000 & \\
8k   & (AWQ-int4) - FP8            & 0.667 & 0.279 & 1.948 & 0.661 & 0.255 & \\
8k   & (AWQ-int4) - (GPTQ-int4)    & 0.694 & 0.277 & 2.002 & 0.667 & 0.183 & \\
8k   & (AWQ-int4) - (GPTQ-int8)    & -0.108 & 0.326 & 0.898 & 0.473 & 1.000 & \\
8k   & (BNB-nf4) - FP8             & 0.719 & 0.283 & 2.053 & 0.672 & 0.164 & \\
8k   & (BNB-nf4) - (GPTQ-int4)     & 0.747 & 0.281 & 2.110 & 0.678 & 0.119 & \\
8k   & (BNB-nf4) - (GPTQ-int8)     & -0.055 & 0.328 & 0.946 & 0.486 & 1.000 & \\
8k   & FP8 - (GPTQ-int4)           & 0.027 & 0.233 & 1.028 & 0.507 & 1.000 & \\
8k   & FP8 - (GPTQ-int8)           & -0.775 & 0.288 & 0.461 & 0.315 & 0.108 & \\
8k   & (GPTQ-int4) - (GPTQ-int8)   & -0.802 & 0.287 & 0.448 & 0.310 & 0.078 & \\
64k  & BF16 - (AWQ-int4)           & 0.057 & 0.090 & 1.058 & 0.514 & 1.000 & \\
64k  & BF16 - (BNB-nf4)            & 0.700 & 0.087 & 2.013 & 0.668 & 0.000 & *** \\
64k  & BF16 - FP8                  & 0.053 & 0.090 & 1.054 & 0.513 & 1.000 & \\
64k  & BF16 - (GPTQ-int4)          & 0.338 & 0.088 & 1.402 & 0.584 & 0.002 & ** \\
64k  & BF16 - (GPTQ-int8)          & 0.012 & 0.090 & 1.012 & 0.503 & 1.000 & \\
64k  & (AWQ-int4) - (BNB-nf4)      & 0.643 & 0.086 & 1.902 & 0.655 & 0.000 & *** \\
64k  & (AWQ-int4) - FP8            & -0.004 & 0.090 & 0.996 & 0.499 & 1.000 & \\
64k  & (AWQ-int4) - (GPTQ-int4)    & 0.281 & 0.088 & 1.324 & 0.570 & 0.021 & * \\
64k  & (AWQ-int4) - (GPTQ-int8)    & -0.045 & 0.090 & 0.956 & 0.489 & 1.000 & \\
64k  & (BNB-nf4) - FP8             & -0.647 & 0.086 & 0.524 & 0.344 & 0.000 & *** \\
64k  & (BNB-nf4) - (GPTQ-int4)     & -0.362 & 0.084 & 0.696 & 
0.411 & 0.000 & *** \\
64k  & (BNB-nf4) - (GPTQ-int8)     & -0.687 & 0.087 & 0.503 & 0.335 & 0.000 & *** \\
64k  & FP8 - (GPTQ-int4)           & 0.285 & 0.088 & 1.330 & 0.571 & 0.018 & * \\
64k  & FP8 - (GPTQ-int8)           & -0.041 & 0.090 & 0.960 & 0.490 & 1.000 & \\
64k  & (GPTQ-int4) - (GPTQ-int8)   & -0.325 & 0.088 & 0.722 & 0.419 & 0.003 & ** \\
128k & BF16 - (AWQ-int4)           & 0.317 & 0.081 & 1.373 & 0.579 & 0.001 & ** \\
128k & BF16 - (BNB-nf4)            & 1.149 & 0.083 & 3.154 & 0.759 & 0.000 & *** \\
128k & BF16 - FP8                  & 0.226 & 0.081 & 1.253 & 0.556 & 0.084 & \\
128k & BF16 - (GPTQ-int4)          & 0.575 & 0.081 & 1.778 & 0.640 & 0.000 & *** \\
128k & BF16 - (GPTQ-int8)          & 0.064 & 0.082 & 1.066 & 0.516 & 1.000 & \\
128k & (AWQ-int4) - (BNB-nf4)      & 0.831 & 0.082 & 2.296 & 0.697 & 0.000 & *** \\
128k & (AWQ-int4) - FP8            & -0.092 & 0.081 & 0.912 & 0.477 & 1.000 & \\
128k & (AWQ-int4) - (GPTQ-int4)    & 0.258 & 0.081 & 1.294 & 0.564 & 0.021 & * \\
128k & (AWQ-int4) - (GPTQ-int8)    & -0.254 & 0.081 & 0.776 & 0.437 & 0.027 & * \\
128k & (BNB-nf4) - FP8             & -0.923 & 0.083 & 0.397 & 0.284 & 0.000 & *** \\
128k & (BNB-nf4) - (GPTQ-int4)     & -0.573 & 0.082 & 0.564 & 0.360 & 0.000 & *** \\
128k & (BNB-nf4) - (GPTQ-int8)     & -1.085 & 0.083 & 0.338 & 0.253 & 0.000 & *** \\
128k & FP8 - (GPTQ-int4)           & 0.350 & 0.081 & 1.419 & 0.587 & 0.000 & *** \\
128k & FP8 - (GPTQ-int8)           & -0.162 & 0.081 & 0.850 & 0.460 & 0.695 & \\
128k & (GPTQ-int4) - (GPTQ-int8)   & -0.512 & 0.081 & 0.599 & 0.375 & 0.000 & *** \\
\bottomrule
\end{tabular}
}
\caption{\textbf{\ruler{}}: Post-hoc comparisons between quantization methods for \textbf{accuracy} (Table~\ref{tab:ruler-stats-context_Len}) across context lengths using Bonferroni adjustments for multiple pairwise comparisons. The probability values refer to the likelihood that the quantization method listed first in each contrast is more accurate than the one listed second (i.e., a value of 0.5 suggests that both methods are comparable in terms of accuracy).}

\label{tab:ruler-posthoc-context}
\vspace{0.5em}
\end{table*}

\begin{table*}[ht]
\centering
\resizebox{0.7\textwidth}{!}{%
\begin{tabular}{l
                S[table-format=2.6]
                S[table-format=1.6]
                S[table-format=1.6]
                S[table-format=1.6]
                S[table-format=1.6]
                c}
\toprule
\multicolumn{7}{l}{\texttt{glmer(Binary\_label $\sim$ Quantization * Task + (1|Model), data = data, family = binomial)}} \\
\bottomrule
\textsc{Predictors} & \textsc{Estimate} & \textsc{OR} & \textsc{CI (lower)} & \textsc{CI (upper)} & \texttt{p-value} & \\
\midrule
Intercept                              &  1.790484 & 5.992354 & 3.346766 & 10.729253 & 0.000 & *** \\
AWQ-int4                   & -0.120452 & 0.886520 & 0.729797 & 1.076900 & 0.224921 & \\
BNB-nf4                    & -0.934520 & 0.392774 & 0.328292 & 0.469922 & 0.000 & *** \\
FP8                        & -0.144514 & 0.865443 & 0.713120 & 1.050302 & 0.143442 & \\
GPTQ-int4                  & -0.291422 & 0.747200 & 0.617883 & 0.903582 & 0.002650 & ** \\
GPTQ-int8                  & -0.061292 & 0.940548 & 0.773028 & 1.144371 & 0.540238 & \\
multi-key                          & -0.876615 & 0.416189 & 0.347804 & 0.498021 & 0.000 & *** \\
multi-value                        & -0.381977 & 0.682511 & 0.565913 & 0.823132 & 0.000064 & *** \\
AWQ-int4:multi-key     & -0.060151 & 0.941623 & 0.732480 & 1.210481 & 0.638799 & \\
BNB-nf4:multi-key      &  0.312536 & 1.366887 & 1.078376 & 1.732586 & 0.009773 & ** \\
FP8:multi-key          &  0.053002 & 1.054432 & 0.820499 & 1.355062 & 0.678777 & \\
GPTQ-int4:multi-key    & -0.142218 & 0.867432 & 0.678218 & 1.109433 & 0.257302 & \\
GPTQ-int8:multi-key    & -0.003379 & 0.996626 & 0.773644 & 1.283878 & 0.979136 & \\
AWQ-int4:multi-value   & -0.053496 & 0.947910 & 0.729822 & 1.231167 & 0.688405 & \\
BNB-nf4:multi-value    &  0.481421 & 1.618372 & 1.264088 & 2.071951 & 0.000134 & *** \\
FP8:multi-value        & -0.071815 & 0.930703 & 0.717581 & 1.207123 & 0.588336 & \\
GPTQ-int4:multi-value  & -0.083880 & 0.919542 & 0.712088 & 1.187434 & 0.520217 & \\
GPTQ-int8:multi-value  &  0.061286 & 1.063203 & 0.815730 & 1.385754 & 0.650296 & \\
\bottomrule
\textsc{Random Effects} & & & & & & \\
\bottomrule
\\
$\sigma^2$ (residual variance) & \multicolumn{5}{l}{1.000} & \\
$\tau_{00}$ (Model) & \multicolumn{5}{l}{0.428} & \\
ICC (Model) & \multicolumn{5}{l}{0.300} & \\
N (Model) & \multicolumn{5}{l}{5} & \\
Observations & \multicolumn{5}{l}{27000} & \\
\bottomrule
\\
$R^2$ (marginal) & \multicolumn{5}{l}{0.157} & \\
$R^2$ (conditional) & \multicolumn{5}{l}{0.047} & \\
\bottomrule
\end{tabular}
}
\caption{\textbf{\ruler{}}: Summary of generalized linear mixed model with \textbf{quantization}, \textbf{task}, and their interaction as predictors of \textbf{accuracy}: \texttt{glmer(Binary\_label $\sim$ Quantization * Task + (1|Model), data = data, family = binomial)}. The quantization method "BF16" on the \textit{needle} task was set as the reference level (intercept), with \texttt{Model} set as the grouping factor for a random intercept to account for variability across models. See Table~\ref{tab:ruler-posthoc-task} for post-hoc analysis.}

\vspace{0.5em}
\label{tab:ruler-task-interaction}
\end{table*}

\begin{table*}[ht]
\centering
\resizebox{0.7\textwidth}{!}{
\begin{tabular}{l
                l
                S[table-format=1.3]
                S[table-format=1.3]
                S[table-format=1.3]
                S[table-format=1.3]
                S[table-format=1.3]
                c}
\toprule
\textsc{Task} & \textsc{Contrast} & \textsc{Estimate} & \textsc{SE} & \textsc{OR} & \textsc{Probability} & \texttt{p-value} & \\
\midrule
needle & BF16 - (AWQ-int4)           & 0.120 & 0.099 & 1.128 & 0.530 & 1.000 & \\
needle & BF16 - (BNB-nf4)            & 0.935 & 0.091 & 2.546 & 0.718 & 0.000 & *** \\
needle & BF16 - FP8                  & 0.145 & 0.099 & 1.155 & 0.536 & 1.000 & \\
needle & BF16 - (GPTQ-int4)          & 0.291 & 0.097 & 1.338 & 0.572 & 0.040 & * \\
needle & BF16 - (GPTQ-int8)          & 0.061 & 0.100 & 1.063 & 0.515 & 1.000 & \\
needle & (AWQ-int4) - (BNB-nf4)      & 0.814 & 0.091 & 2.257 & 0.693 & 0.000 & *** \\
needle & (AWQ-int4) - FP8            & 0.024 & 0.098 & 1.024 & 0.506 & 1.000 & \\
needle & (AWQ-int4) - (GPTQ-int4)    & 0.171 & 0.096 & 1.186 & 0.543 & 1.000 & \\
needle & (AWQ-int4) - (GPTQ-int8)    & -0.059 & 0.099 & 0.943 & 0.485 & 1.000 & \\
needle & (BNB-nf4) - FP8             & -0.790 & 0.090 & 0.454 & 0.312 & 0.000 & *** \\
needle & (BNB-nf4) - (GPTQ-int4)     & -0.643 & 0.088 & 0.526 & 0.345 & 0.000 & *** \\
needle & (BNB-nf4) - (GPTQ-int8)     & -0.873 & 0.092 & 0.418 & 0.295 & 0.000 & *** \\
needle & FP8 - (GPTQ-int4)           & 0.147 & 0.096 & 1.158 & 0.537 & 1.000 & \\
needle & FP8 - (GPTQ-int8)           & -0.083 & 0.099 & 0.920 & 0.479 & 1.000 & \\
needle & (GPTQ-int4) - (GPTQ-int8)   & -0.230 & 0.097 & 0.794 & 0.443 & 0.264 & \\
multi-key & BF16 - (AWQ-int4)        & 0.181 & 0.082 & 1.198 & 0.545 & 0.404 & \\
multi-key & BF16 - (BNB-nf4)         & 0.622 & 0.080 & 1.863 & 0.651 & 0.000 & *** \\
multi-key & BF16 - FP8               & 0.092 & 0.082 & 1.096 & 0.523 & 1.000 & \\
multi-key & BF16 - (GPTQ-int4)       & 0.434 & 0.081 & 1.543 & 0.607 & 0.000 & *** \\
multi-key & BF16 - (GPTQ-int8)       & 0.065 & 0.082 & 1.067 & 0.516 & 1.000 & \\
multi-key & (AWQ-int4) - (BNB-nf4)   & 0.441 & 0.079 & 1.555 & 0.609 & 0.000 & *** \\
multi-key & (AWQ-int4) - FP8         & -0.089 & 0.081 & 0.915 & 0.478 & 1.000 & \\
multi-key & (AWQ-int4) - (GPTQ-int4) & 0.253 & 0.080 & 1.288 & 0.563 & 0.022 & * \\
multi-key & (AWQ-int4) - (GPTQ-int8) & -0.116 & 0.081 & 0.891 & 0.471 & 1.000 & \\
multi-key & (BNB-nf4) - FP8          & -0.530 & 0.080 & 0.588 & 0.370 & 0.000 & *** \\
multi-key & (BNB-nf4) - (GPTQ-int4)  & -0.188 & 0.078 & 0.828 & 0.453 & 0.235 & \\
multi-key & (BNB-nf4) - (GPTQ-int8)  & -0.557 & 0.080 & 0.573 & 0.364 & 0.000 & *** \\
multi-key & FP8 - (GPTQ-int4)        & 0.342 & 0.080 & 1.408 & 0.585 & 0.000 & *** \\
multi-key & FP8 - (GPTQ-int8)        & -0.027 & 0.082 & 0.974 & 0.493 & 1.000 & \\
multi-key & (GPTQ-int4) - (GPTQ-int8)& -0.369 & 0.080 & 0.691 & 0.409 & 0.000 & *** \\
multi-value & BF16 - (AWQ-int4)      & 0.174 & 0.090 & 1.190 & 0.543 & 0.788 & \\
multi-value & BF16 - (BNB-nf4)       & 0.453 & 0.087 & 1.573 & 0.611 & 0.000 & *** \\
multi-value & BF16 - FP8             & 0.216 & 0.089 & 1.242 & 0.554 & 0.231 & \\
multi-value & BF16 - (GPTQ-int4)     & 0.375 & 0.088 & 1.455 & 0.593 & 0.000 & *** \\
multi-value & BF16 - (GPTQ-int8)     & 0.000 & 0.092 & 1.000 & 0.500 & 1.000 & \\
multi-value & (AWQ-int4) - (BNB-nf4) & 0.279 & 0.086 & 1.322 & 0.569 & 0.017 & * \\
multi-value & (AWQ-int4) - FP8       & 0.042 & 0.088 & 1.043 & 0.511 & 1.000 & \\
multi-value & (AWQ-int4) - (GPTQ-int4)& 0.201 & 0.086 & 1.223 & 0.550 & 0.297 & \\
multi-value & (AWQ-int4) - (GPTQ-int8)& -0.174 & 0.090 & 0.840 & 0.457 & 0.798 & \\
multi-value & (BNB-nf4) - FP8        & -0.237 & 0.085 & 0.789 & 0.441 & 0.084 & \\
multi-value & (BNB-nf4) - (GPTQ-int4)& -0.078 & 0.084 & 0.925 & 0.481 & 1.000 & \\
multi-value & (BNB-nf4) - (GPTQ-int8)& -0.453 & 0.088 & 0.636 & 0.389 & 0.000 & *** \\
multi-value & FP8 - (GPTQ-int4)      & 0.159 & 0.086 & 1.172 & 0.540 & 0.969 & \\
multi-value & FP8 - (GPTQ-int8)      & -0.216 & 0.090 & 0.805 & 0.446 & 0.236 & \\
multi-value & (GPTQ-int4) - (GPTQ-int8)& -0.375 & 0.088 & 0.687 & 0.407 & 0.000 & *** \\
\bottomrule
\end{tabular}
}
\caption{\textbf{\ruler{}}: Post-hoc comparisons between quantization methods for \textbf{accuracy} (Table~\ref{tab:ruler-task-interaction}) across tasks using Bonferroni adjustments for multiple pairwise comparisons. The probability values refer to the likelihood that the quantization method listed first in each contrast is more accurate than the one listed second.}
\label{tab:ruler-posthoc-task}
\vspace{0.5em}
\end{table*}

%%%%%%%%%%%% One-ruler Tables

\begin{table*}[ht]
\centering
\resizebox{0.7\textwidth}{!}{
\begin{tabular}{l
                S[table-format=1.3]
                S[table-format=1.3]
                S[table-format=1.3]
                S[table-format=1.3]
                S[table-format=1.5]
                c}
\toprule
\multicolumn{7}{c}{
\begin{tabular}{c}
\texttt{glmer(Binary\_label $\sim$ Quantization + (1|Model),} \\
\texttt{data = data, family = binomial)}
\end{tabular}
} \\
\bottomrule
\textsc{Predictors} & \textsc{Estimate} & \textsc{OR} & \textsc{CI (lower)} & \textsc{CI (upper)} & \texttt{p-value} & \\
\midrule
Intercept            &  1.710 & 5.532 & 3.712 & 8.243 & 0.000 & *** \\
AWQ-int4             & -0.096 & 0.908 & 0.876 & 0.943 & 0.000 & *** \\
BNB-nf4              & -0.636 & 0.529 & 0.511 & 0.548 & 0.000 & *** \\
FP8                  &  0.012 & 1.012 & 0.975 & 1.051 & 0.516 &     \\
GPTQ-int4            & -0.173 & 0.841 & 0.811 & 0.872 & 0.000 & *** \\
GPTQ-int8            & -0.024 & 0.976 & 0.940 & 1.013 & 0.200 &     \\
\bottomrule
\textsc{Random Effects} & & & & & & \\
\bottomrule
\\
$\sigma^2$ (residual variance) & \multicolumn{5}{l}{3.290} & \\
$\tau_{00}$ (Model) & \multicolumn{5}{l}{0.841} & \\
ICC (Model) & \multicolumn{5}{l}{0.204} & \\
N (Model) & \multicolumn{5}{l}{5} & \\
Observations & \multicolumn{5}{l}{234000} & \\
\bottomrule
\\
$R^2$ (marginal) & \multicolumn{5}{l}{0.012} & \\
$R^2$ (conditional) & \multicolumn{5}{l}{0.213} & \\
\bottomrule
\end{tabular}
}
\caption{\textbf{\oneruler{}}: Summary of generalized linear mixed model with \textbf{quantization} as the predictor of \textbf{accuracy}: \texttt{glmer(Binary\_label $\sim$ Quantization + (1|Model), data = data, family = binomial)}. The quantization method "BF16" was set as the reference level (intercept), with \texttt{Model} set as the grouping factor for a random intercept to account for variability across models. See Table~\ref{tab:oneruler-posthoc-model} for post-hoc analysis.}
\label{tab:oneruler-stats-model-glmm}
\vspace{0.5em}
\end{table*}

\begin{table*}[ht]
\centering
\resizebox{0.7\textwidth}{!}{
\begin{tabular}{l
                S[table-format=1.3]
                S[table-format=1.3]
                S[table-format=1.3]
                S[table-format=1.3]
                S[table-format=1.3]
                c}
\toprule
\textsc{Contrast} & \textsc{Estimate} & \textsc{SE} & \textsc{OR} & \textsc{Probability} & \texttt{p-value} & \\
\midrule
BF16 - (AWQ-int4)            & 0.096 & 0.019 & 1.101 & 0.524 & 0.000 & *** \\
BF16 - (BNB-nf4)             & 0.636 & 0.018 & 1.889 & 0.654 & 0.000 & *** \\
BF16 - FP8                  & -0.012 & 0.019 & 0.988 & 0.497 & 1.000 &     \\
BF16 - (GPTQ-int4)          & 0.173 & 0.019 & 1.189 & 0.543 & 0.000 & *** \\
BF16 - (GPTQ-int8)          & 0.024 & 0.019 & 1.025 & 0.506 & 1.000 &     \\

(AWQ-int4) - (BNB-nf4)       & 0.540 & 0.018 & 1.716 & 0.632 & 0.000 & *** \\
(AWQ-int4) - FP8            & -0.108 & 0.019 & 0.897 & 0.473 & 0.000 & *** \\
(AWQ-int4) - (GPTQ-int4)     & 0.077 & 0.019 & 1.081 & 0.519 & 0.000 & *** \\
(AWQ-int4) - (GPTQ-int8)    & -0.072 & 0.019 & 0.931 & 0.482 & 0.002 & **  \\

(BNB-nf4) - FP8             & -0.649 & 0.018 & 0.523 & 0.343 & 0.000 & *** \\
(BNB-nf4) - (GPTQ-int4)     & -0.463 & 0.018 & 0.630 & 0.386 & 0.000 & *** \\
(BNB-nf4) - (GPTQ-int8)     & -0.612 & 0.018 & 0.542 & 0.352 & 0.000 & *** \\

FP8 - (GPTQ-int4)            & 0.186 & 0.019 & 1.204 & 0.546 & 0.000 & *** \\
FP8 - (GPTQ-int8)            & 0.037 & 0.019 & 1.037 & 0.509 & 0.821 &     \\

(GPTQ-int4) - (GPTQ-int8)   & -0.149 & 0.019 & 0.861 & 0.463 & 0.000 & *** \\
\bottomrule
\end{tabular}
}
\caption{\textbf{\oneruler{}}: Post-hoc comparisons between quantization methods for \textbf{accuracy} (Table~\ref{tab:oneruler-stats-model-glmm}) using Bonferroni adjustments for multiple pairwise comparisons. The probability values refer to the likelihood that the quantization method listed first in each contrast is more accurate than the one listed second (i.e., a value of 0.5 suggests that both methods are comparable in terms of accuracy).}
\label{tab:oneruler-posthoc-model}
\vspace{0.5em}
\end{table*}

\begin{table*}[ht]
\centering
\resizebox{0.7\textwidth}{!}{%
\begin{tabular}{l
                S[table-format=2.6]
                S[table-format=2.6]
                S[table-format=1.6]
                S[table-format=1.6]
                S[table-format=1.6]
                c}
\toprule
\multicolumn{7}{c}{
\begin{tabular}{c}
\texttt{glmer(Binary\_label $\sim$ Quantization * Context\_length + (1|Model)} \\
\texttt{+ (1|Frequency), data = data, family = binomial)}
\end{tabular}
} \\
\bottomrule
\textsc{Predictors} & \textsc{Estimate} & \textsc{OR} & \textsc{CI (lower)} & \textsc{CI (upper)} & \texttt{p-value} & \\
\midrule
Intercept                               & 2.403159 & 11.058054 & 6.371058 & 19.193131 & 0.000 & *** \\
AWQ-int4                                & 0.009621 & 1.009668 & 0.935044 & 1.090247 & 0.806 &     \\
BNB-nf4                                 & -0.496620 & 0.608584 & 0.566110 & 0.654245 & 0.000 & *** \\
FP8                                     & 0.108448 & 1.114547 & 1.030326 & 1.205653 & 0.007 & **  \\
GPTQ-int4                               & -0.087498 & 0.916221 & 0.849374 & 0.988329 & 0.024 & *   \\
GPTQ-int8                               & -0.126709 & 0.880990 & 0.818044 & 0.948779 & 0.001 & *** \\
64k                                     & -0.701749 & 0.495718 & 0.463028 & 0.530715 & 0.000 & *** \\
128k                                    & -1.237181 & 0.290201 & 0.271461 & 0.310235 & 0.000 & *** \\
AWQ-int4:64k                            & -0.105650 & 0.899740 & 0.815131 & 0.993131 & 0.036 & *   \\
BNB-nf4:64k                             & -0.291492 & 0.747148 & 0.680579 & 0.820229 & 0.000 & *** \\
FP8:64k                                 & -0.135493 & 0.873285 & 0.790256 & 0.965039 & 0.008 & **  \\
GPTQ-int4:64k                           & -0.182414 & 0.833256 & 0.756546 & 0.917745 & 0.000 & *** \\
GPTQ-int8:64k                           & 0.070655 & 1.073210 & 0.974969 & 1.181351 & 0.149 &     \\
AWQ-int4:128k                           & -0.185697 & 0.830525 & 0.755099 & 0.913485 & 0.000 & *** \\
BNB-nf4:128k                            & -0.263432 & 0.768410 & 0.701288 & 0.841957 & 0.000 & *** \\
FP8:128k                                & -0.111812 & 0.894212 & 0.810726 & 0.986296 & 0.025 & *   \\
GPTQ-int4:128k                          & -0.095802 & 0.908644 & 0.826231 & 0.999278 & 0.048 & *   \\
GPTQ-int8:128k                          & 0.181624 & 1.199163 & 1.091596 & 1.317331 & 0.000 & *** \\
\bottomrule
\textsc{Random Effects} & & & & & & \\
\bottomrule
\\
$\sigma^2$ (residual variance) & \multicolumn{5}{l}{1.000} & \\
$\tau_{00}$ (Model) & \multicolumn{5}{l}{3.29} & \\
$\tau_{00}$ (Frequency) & \multicolumn{5}{l}{0.723} & \\
ICC (Model) & \multicolumn{5}{l}{0.218} & \\  % Not reported
ICC (Frequency) & \multicolumn{5}{l}{0.141} & \\  % Not reported
N (Model) & \multicolumn{5}{l}{5} & \\
N (Frequency) & \multicolumn{5}{l}{3} & \\
Observations & \multicolumn{5}{l}{234000} & \\
\bottomrule
\\
$R^2$ (marginal) & \multicolumn{5}{l}{0.065} & \\
$R^2$ (conditional) & \multicolumn{5}{l}{0.401} & \\
\bottomrule
\end{tabular}
}
\caption{\textbf{\oneruler{}}: Summary of generalized linear mixed model with \textbf{quantization}, \textbf{context length}, and their interaction as predictors of \textbf{accuracy}: \texttt{glmer(Binary\_label $\sim$ Quantization * Context\_length + (1|Model) + (1|Frequency), data = data, family = binomial)}. The quantization method "BF16" at context length 8k was set as the reference level (intercept), with \texttt{Model} and \texttt{Frequency} set as the grouping factors for random intercepts to account for variability across models and frequency bands. See Table~\ref{tab:oneruler-posthoc-context} for post-hoc analysis.}
\label{tab:oneruler-context}
\vspace{0.5em}
\end{table*}

\begin{table*}[ht]
\centering
\resizebox{0.7\textwidth}{!}{
\begin{tabular}{l
                l
                S[table-format=1.3]
                S[table-format=1.3]
                S[table-format=1.3]
                S[table-format=1.3]
                S[table-format=1.3]
                c}
\toprule
\textsc{Context} & \textsc{Contrast} & \textsc{Estimate} & \textsc{SE} & \textsc{OR} & \textsc{Probability} & \texttt{p-value} & \\
\midrule
8k & BF16 - (AWQ-int4)           & -0.010 & 0.039 & 0.990 & 0.498 & 1.000 & \\
8k & BF16 - (BNB-nf4)            &  0.497 & 0.037 & 1.643 & 0.622 & 0.000 & *** \\
8k & BF16 - FP8                  & -0.108 & 0.040 & 0.897 & 0.473 & 0.102 & \\
8k & BF16 - (GPTQ-int4)          &  0.087 & 0.039 & 1.091 & 0.522 & 0.354 & \\
8k & BF16 - (GPTQ-int8)          &  0.127 & 0.038 & 1.135 & 0.532 & 0.012 & * \\
8k & (AWQ-int4) - (BNB-nf4)      &  0.506 & 0.038 & 1.659 & 0.624 & 0.000 & *** \\
8k & (AWQ-int4) - FP8            & -0.099 & 0.042 & 0.906 & 0.475 & 0.278 & \\
8k & (AWQ-int4) - (GPTQ-int4)    &  0.097 & 0.041 & 1.102 & 0.524 & 0.253 & \\
8k & (AWQ-int4) - (GPTQ-int8)    &  0.136 & 0.040 & 1.146 & 0.534 & 0.011 & * \\
8k & (BNB-nf4) - FP8             & -0.605 & 0.039 & 0.546 & 0.353 & 0.000 & *** \\
8k & (BNB-nf4) - (GPTQ-int4)     & -0.409 & 0.038 & 0.664 & 0.399 & 0.000 & *** \\
8k & (BNB-nf4) - (GPTQ-int8)     & -0.370 & 0.038 & 0.691 & 0.409 & 0.000 & *** \\
8k & FP8 - (GPTQ-int4)           &  0.196 & 0.041 & 1.216 & 0.549 & 0.000 & *** \\
8k & FP8 - (GPTQ-int8)           &  0.235 & 0.041 & 1.265 & 0.559 & 0.000 & *** \\
8k & (GPTQ-int4) - (GPTQ-int8)   &  0.039 & 0.040 & 1.040 & 0.510 & 1.000 & \\

64k & BF16 - (AWQ-int4)          &  0.096 & 0.033 & 1.101 & 0.524 & 0.059 & \\
64k & BF16 - (BNB-nf4)           &  0.788 & 0.032 & 2.199 & 0.687 & 0.000 & *** \\
64k & BF16 - FP8                 &  0.027 & 0.033 & 1.027 & 0.507 & 1.000 & \\
64k & BF16 - (GPTQ-int4)         &  0.270 & 0.033 & 1.310 & 0.567 & 0.000 & *** \\
64k & BF16 - (GPTQ-int8)         &  0.056 & 0.033 & 1.058 & 0.514 & 1.000 & \\
64k & (AWQ-int4) - (BNB-nf4)     &  0.692 & 0.032 & 1.998 & 0.666 & 0.000 & *** \\
64k & (AWQ-int4) - FP8           & -0.069 & 0.034 & 0.933 & 0.483 & 0.618 & \\
64k & (AWQ-int4) - (GPTQ-int4)   &  0.174 & 0.033 & 1.190 & 0.543 & 0.000 & *** \\
64k & (AWQ-int4) - (GPTQ-int8)   & -0.040 & 0.034 & 0.961 & 0.490 & 1.000 & \\
64k & (BNB-nf4) - FP8            & -0.761 & 0.032 & 0.467 & 0.318 & 0.000 & *** \\
64k & (BNB-nf4) - (GPTQ-int4)    & -0.518 & 0.031 & 0.596 & 0.373 & 0.000 & *** \\
64k & (BNB-nf4) - (GPTQ-int8)    & -0.732 & 0.032 & 0.481 & 0.325 & 0.000 & *** \\
64k & FP8 - (GPTQ-int4)          &  0.243 & 0.033 & 1.275 & 0.560 & 0.000 & *** \\
64k & FP8 - (GPTQ-int8)          &  0.029 & 0.034 & 1.029 & 0.507 & 1.000 & \\
64k & (GPTQ-int4) - (GPTQ-int8)  & -0.214 & 0.033 & 0.807 & 0.447 & 0.000 & *** \\
128k & BF16 - (AWQ-int4)         &  0.176 & 0.030 & 1.193 & 0.544 & 0.000 & *** \\
128k & BF16 - (BNB-nf4)          &  0.760 & 0.029 & 2.138 & 0.681 & 0.000 & *** \\
128k & BF16 - FP8                &  0.003 & 0.031 & 1.003 & 0.501 & 1.000 & \\
128k & BF16 - (GPTQ-int4)        &  0.183 & 0.030 & 1.201 & 0.546 & 0.000 & *** \\
128k & BF16 - (GPTQ-int8)        & -0.055 & 0.031 & 0.947 & 0.486 & 1.000 & \\
128k & (AWQ-int4) - (BNB-nf4)    &  0.584 & 0.029 & 1.793 & 0.642 & 0.000 & *** \\
128k & (AWQ-int4) - FP8          & -0.173 & 0.030 & 0.841 & 0.457 & 0.000 & *** \\
128k & (AWQ-int4) - (GPTQ-int4)  &  0.007 & 0.030 & 1.007 & 0.502 & 1.000 & \\
128k & (AWQ-int4) - (GPTQ-int8)  & -0.231 & 0.031 & 0.794 & 0.443 & 0.000 & *** \\
128k & (BNB-nf4) - FP8           & -0.757 & 0.030 & 0.469 & 0.319 & 0.000 & *** \\
128k & (BNB-nf4) - (GPTQ-int4)   & -0.577 & 0.029 & 0.562 & 0.360 & 0.000 & *** \\
128k & (BNB-nf4) - (GPTQ-int8)   & -0.815 & 0.030 & 0.443 & 0.307 & 0.000 & *** \\
128k & FP8 - (GPTQ-int4)         &  0.180 & 0.030 & 1.197 & 0.545 & 0.000 & *** \\
128k & FP8 - (GPTQ-int8)         & -0.058 & 0.031 & 0.943 & 0.485 & 0.899 & \\
128k & (GPTQ-int4) - (GPTQ-int8) & -0.238 & 0.031 & 0.788 & 0.441 & 0.000 & *** \\
\bottomrule
\end{tabular}
}
\caption{\textbf{\oneruler{}}: Post-hoc comparisons between quantization methods for \textbf{accuracy} (Table~\ref{tab:oneruler-context}) across context lengths using Bonferroni adjustments for multiple pairwise comparisons. The probability values refer to the likelihood that the quantization method listed first in each contrast is more accurate than the one listed second.}
\label{tab:oneruler-posthoc-context}
\vspace{0.5em}
\end{table*}

\begin{table*}[ht]
\centering
\resizebox{0.7\textwidth}{!}{%
\begin{tabular}{l
                S[table-format=2.6]
                S[table-format=1.6]
                S[table-format=1.6]
                S[table-format=1.6]
                S[table-format=1.6]
                c}
\toprule
\multicolumn{7}{l}{\texttt{glmer(Binary\_label $\sim$ Quantization * Frequency + (1|Model), data = data, family = binomial)}} \\
\bottomrule
\textsc{Predictors} & \textsc{Estimate} & \textsc{OR} & \textsc{CI (lower)} & \textsc{CI (upper)} & \texttt{p-value} & \\
\midrule
Intercept                                        &  0.626521 & 1.871090 & 0.751868 & 4.656373 & 0.178 &     \\
AWQ-int4                                         & -0.175971 & 0.838642 & 0.776665 & 0.905566 & 0.000 & *** \\
BNB-nf4                                          & -0.379553 & 0.684167 & 0.633749 & 0.738597 & 0.000 & *** \\
FP8                                              & -0.009310 & 0.990733 & 0.917129 & 1.070244 & 0.813 &     \\
GPTQ-int4                                        & -0.146050 & 0.864115 & 0.800061 & 0.933296 & 0.000 & *** \\
GPTQ-int8                                        & -0.055703 & 0.945820 & 0.875489 & 1.021801 & 0.158 &     \\
English                                          &  1.667311 & 5.297903 & 4.624733 & 6.069058 & 0.000 & *** \\
High-resource                                    &  1.362028 & 3.904102 & 3.669582 & 4.153610 & 0.000 & *** \\
AWQ-int4:English                                 &  0.005429 & 1.005444 & 0.817509 & 1.236582 & 0.959 &     \\
BNB-nf4:English                                  &  0.110170 & 1.116468 & 0.916823 & 1.359587 & 0.273 &     \\
FP8:English                                      &  0.127334 & 1.135796 & 0.922990 & 1.397668 & 0.229 &     \\
GPTQ-int4:English                                & -0.012479 & 0.987599 & 0.800225 & 1.218846 & 0.907 &     \\
GPTQ-int8:English                                &  0.130988 & 1.139955 & 0.931042 & 1.395744 & 0.205 &     \\
AWQ-int4:High-resource                           &  0.099453 & 1.104567 & 1.011560 & 1.206125 & 0.027 & *   \\
BNB-nf4:High-resource                            & -0.376512 & 0.686251 & 0.629313 & 0.748340 & 0.000 & *** \\
FP8:High-resource                                &  0.025458 & 1.025785 & 0.938630 & 1.121032 & 0.574 &     \\
GPTQ-int4:High-resource                          & -0.049395 & 0.951805 & 0.871680 & 1.039296 & 0.271 &     \\
GPTQ-int8:High-resource                          &  0.035289 & 1.035919 & 0.948005 & 1.131987 & 0.435 &     \\
\bottomrule
\textsc{Random Effects} & & & & & & \\
\bottomrule
\\
$\sigma^2$ (residual variance) & \multicolumn{5}{l}{3.290} & \\
$\tau_{00}$ (Model) & \multicolumn{5}{l}{0.924} & \\
ICC (Model) & \multicolumn{5}{l}{0.219} & \\
N (Model) & \multicolumn{5}{l}{5} & \\
Observations & \multicolumn{5}{l}{234000} & \\
\bottomrule
\\
$R^2$ (marginal) & \multicolumn{5}{l}{0.067} & \\
$R^2$ (conditional) & \multicolumn{5}{l}{0.271} & \\
\bottomrule
\end{tabular}
}
\caption{\textbf{\oneruler{}}: Summary of generalized linear mixed model with \textbf{quantization}, \textbf{language frequency}, and their interaction as predictors of \textbf{accuracy}: \texttt{glmer(Binary\_label $\sim$ Quantization * Frequency + (1|Model), data = data, family = binomial)}. The quantization method "BF16" under low-resource frequency was set as the reference level (intercept), with \texttt{Model} set as the grouping factor for a random intercept to account for variability across models. See Table~\ref{tab:oneruler-posthoc-frequency} for post-hoc analysis.}
\label{tab:oneruler-frequency-interaction}
\vspace{0.5em}
\end{table*}

\begin{table*}[ht]
\centering
\resizebox{0.7\textwidth}{!}{
\begin{tabular}{l
                l
                S[table-format=1.3]
                S[table-format=1.3]
                S[table-format=1.3]
                S[table-format=1.3]
                S[table-format=1.3]
                c}
\toprule
\textsc{Frequency} & \textsc{Contrast} & \textsc{Estimate} & \textsc{SE} & \textsc{OR} & \textsc{Probability} & \texttt{p-value} & \\
\midrule
Low-resource & BF16 - (AWQ-int4)           & 0.176 & 0.039 & 1.192 & 0.544 & 0.001 & *** \\
Low-resource & BF16 - (BNB-nf4)            & 0.380 & 0.039 & 1.462 & 0.594 & 0.000 & *** \\
Low-resource & BF16 - FP8                  & 0.009 & 0.039 & 1.009 & 0.502 & 1.000 & \\
Low-resource & BF16 - (GPTQ-int4)          & 0.146 & 0.039 & 1.157 & 0.536 & 0.003 & **  \\
Low-resource & BF16 - (GPTQ-int8)          & 0.056 & 0.039 & 1.057 & 0.514 & 1.000 & \\
Low-resource & (AWQ-int4) - (BNB-nf4)      & 0.204 & 0.040 & 1.226 & 0.551 & 0.000 & *** \\
Low-resource & (AWQ-int4) - FP8            & -0.167 & 0.040 & 0.846 & 0.458 & 0.001 & **  \\
Low-resource & (AWQ-int4) - (GPTQ-int4)    & -0.030 & 0.040 & 0.971 & 0.493 & 1.000 & \\
Low-resource & (AWQ-int4) - (GPTQ-int8)    & -0.120 & 0.040 & 0.887 & 0.470 & 0.044 & *   \\
Low-resource & (BNB-nf4) - FP8             & -0.370 & 0.040 & 0.691 & 0.408 & 0.000 & *** \\
Low-resource & (BNB-nf4) - (GPTQ-int4)     & -0.234 & 0.040 & 0.792 & 0.442 & 0.000 & *** \\
Low-resource & (BNB-nf4) - (GPTQ-int8)     & -0.324 & 0.040 & 0.723 & 0.420 & 0.000 & *** \\
Low-resource & FP8 - (GPTQ-int4)           & 0.137 & 0.041 & 1.147 & 0.534 & 0.012 & *   \\
Low-resource & FP8 - (GPTQ-int8)           & 0.046 & 0.041 & 1.047 & 0.512 & 1.000 & \\
Low-resource & (GPTQ-int4) - (GPTQ-int8)   & -0.090 & 0.041 & 0.914 & 0.477 & 0.396 & \\
English & BF16 - (AWQ-int4)                & 0.171 & 0.101 & 1.186 & 0.543 & 1.000 & \\
English & BF16 - (BNB-nf4)                 & 0.269 & 0.096 & 1.309 & 0.567 & 0.076 & \\
English & BF16 - FP8                       & -0.118 & 0.102 & 0.889 & 0.471 & 1.000 & \\
English & BF16 - (GPTQ-int4)               & 0.159 & 0.102 & 1.172 & 0.540 & 1.000 & \\
English & BF16 - (GPTQ-int8)               & -0.075 & 0.100 & 0.927 & 0.481 & 1.000 & \\
English & (AWQ-int4) - (BNB-nf4)           & 0.099 & 0.107 & 1.104 & 0.525 & 1.000 & \\
English & (AWQ-int4) - FP8                 & -0.289 & 0.113 & 0.749 & 0.428 & 0.162 & \\
English & (AWQ-int4) - (GPTQ-int4)         & -0.012 & 0.108 & 0.988 & 0.497 & 1.000 & \\
English & (AWQ-int4) - (GPTQ-int8)         & -0.246 & 0.110 & 0.782 & 0.439 & 0.376 & \\
English & (BNB-nf4) - FP8                  & -0.387 & 0.113 & 0.679 & 0.404 & 0.009 & **  \\
English & (BNB-nf4) - (GPTQ-int4)          & -0.111 & 0.108 & 0.895 & 0.472 & 1.000 & \\
English & (BNB-nf4) - (GPTQ-int8)          & -0.345 & 0.109 & 0.708 & 0.415 & 0.023 & *   \\
English & FP8 - (GPTQ-int4)                & 0.277 & 0.116 & 1.319 & 0.569 & 0.251 & \\
English & FP8 - (GPTQ-int8)                & 0.043 & 0.117 & 1.044 & 0.511 & 1.000 & \\
English & (GPTQ-int4) - (GPTQ-int8)        & -0.234 & 0.111 & 0.792 & 0.442 & 0.535 & \\
High-resource & BF16 - (AWQ-int4)          & 0.077 & 0.023 & 1.080 & 0.519 & 0.011 & *   \\
High-resource & BF16 - (BNB-nf4)           & 0.756 & 0.021 & 2.130 & 0.680 & 0.000 & *** \\
High-resource & BF16 - FP8                 & -0.016 & 0.023 & 0.984 & 0.496 & 1.000 & \\
High-resource & BF16 - (GPTQ-int4)         & 0.195 & 0.022 & 1.216 & 0.549 & 0.000 & *** \\
High-resource & BF16 - (GPTQ-int8)         & 0.020 & 0.023 & 1.021 & 0.505 & 1.000 & \\
High-resource & (AWQ-int4) - (BNB-nf4)     & 0.680 & 0.021 & 1.973 & 0.664 & 0.000 & *** \\
High-resource & (AWQ-int4) - FP8           & -0.093 & 0.023 & 0.911 & 0.477 & 0.001 & **  \\
High-resource & (AWQ-int4) - (GPTQ-int4)   & 0.119 & 0.022 & 1.126 & 0.530 & 0.000 & *** \\
High-resource & (AWQ-int4) - (GPTQ-int8)   & -0.056 & 0.023 & 0.945 & 0.486 & 0.197 & \\
High-resource & (BNB-nf4) - FP8            & -0.772 & 0.021 & 0.462 & 0.316 & 0.000 & *** \\
High-resource & (BNB-nf4) - (GPTQ-int4)    & -0.561 & 0.021 & 0.571 & 0.363 & 0.000 & *** \\
High-resource & (BNB-nf4) - (GPTQ-int8)    & -0.736 & 0.021 & 0.479 & 0.324 & 0.000 & *** \\
High-resource & FP8 - (GPTQ-int4)          & 0.212 & 0.022 & 1.236 & 0.553 & 0.000 & *** \\
High-resource & FP8 - (GPTQ-int8)          & 0.037 & 0.023 & 1.037 & 0.509 & 1.000 & \\
High-resource & (GPTQ-int4) - (GPTQ-int8)  & -0.175 & 0.022 & 0.839 & 0.456 & 0.000 & *** \\
\bottomrule
\end{tabular}
}
\caption{\textbf{\oneruler{}}: Post-hoc comparisons between quantization methods for \textbf{accuracy} (Table~\ref{tab:oneruler-frequency-interaction}) across language frequency using Bonferroni adjustments for multiple pairwise comparisons. The probability values refer to the likelihood that the quantization method listed first in each contrast is more accurate than the one listed second.}
\label{tab:oneruler-posthoc-frequency}
\vspace{0.5em}
\end{table*}

%%%%%% NOCHA TABLES

\begin{table*}[ht]
\centering
\resizebox{0.7\textwidth}{!}{
\begin{tabular}{l
                S[table-format=1.3]
                S[table-format=1.3]
                S[table-format=1.3]
                S[table-format=1.3]
                S[table-format=1.5]
                c}
\toprule
\multicolumn{7}{c}{\texttt{glmer(Binary\_label $\sim$ Quantization + (1|Context\_length) + (1|Model),}} \\
\multicolumn{7}{c}{\texttt{~~~~data = data, family = binomial)}} \\

\bottomrule
\textsc{Predictors} & \textsc{Estimate} & \textsc{OR} & \textsc{CI (lower)} & \textsc{CI (upper)} & \texttt{p-value} & \\
\midrule
Intercept        &  0.510 & 1.666 & 1.302 & 2.132 & 0.000 & *** \\
FP8              &  0.049 & 1.050 & 0.981 & 1.123 & 0.159 &     \\
GPTQ-int8        & -0.018 & 0.983 & 0.918 & 1.052 & 0.613 &     \\
AWQ-int4         & -0.005 & 0.995 & 0.931 & 1.065 & 0.893 &     \\
GPTQ-int4        & -0.002 & 0.998 & 0.933 & 1.067 & 0.952 &     \\
BNB-nf4          & -0.046 & 0.955 & 0.893 & 1.022 & 0.181 &     \\
\bottomrule
\textsc{Random Effects} & & & & & & \\
\bottomrule
\\
$\sigma^2$ (residual variance) & \multicolumn{5}{l}{1.000} & \\
$\tau_{00}$ (Model) & \multicolumn{5}{l}{0.047} & \\
ICC (Model) & \multicolumn{5}{l}{0.045} & \\
N (Model) & \multicolumn{5}{l}{5} & \\
Observations & \multicolumn{5}{l}{42781} & \\
\bottomrule
\\
$R^2$ (marginal) & \multicolumn{5}{l}{0.016} & \\
$R^2$ (conditional) & \multicolumn{5}{l}{0.001} & \\
\bottomrule
\end{tabular}
}
\caption{\textbf{\nocha{}}: Summary of generalized linear mixed model with \textbf{quantization} as the predictor of \textbf{accuracy}: \texttt{glmer(Binary\_label $\sim$ Quantization + (1|Context\_length) + (1|Model), data = data, family = binomial)}. The quantization method "BF16" was set as the reference level (intercept), with \texttt{Context\_length} and \texttt{Model} set as the grouping factors for random intercepts to account for variability across context lengths and models. See Table~\ref{tab:nocha-posthoc-model} for post-hoc analysis.}

\label{tab:nocha-stats}
\vspace{0.5em}
\end{table*}

\begin{table*}[ht]
\centering
\resizebox{0.7\textwidth}{!}{
\begin{tabular}{l
                S[table-format=1.3]
                S[table-format=1.3]
                S[table-format=1.3]
                S[table-format=1.3]
                S[table-format=1.5]
                c}
\toprule
\textsc{Contrast} & \textsc{Estimate} & \textsc{SE} & \textsc{OR} & \textsc{Probability} & \texttt{p-value} & \\
\midrule
BF16 - FP8        & -0.049 & 0.034 & 0.953 & 0.488 & 1.0000000 & \\
BF16 - GPTQ-int8      &  0.018 & 0.035 & 1.018 & 0.504 & 1.0000000 & \\
BF16 - AWQ-int4       &  0.005 & 0.034 & 1.005 & 0.501 & 1.0000000 & \\
BF16 - GPTQ-int4      &  0.002 & 0.034 & 1.002 & 0.501 & 1.0000000 & \\
BF16 - BNB-nf4       &  0.046 & 0.034 & 1.047 & 0.511 & 1.0000000 & \\

AWQ-int4- FP8         & -0.053 & 0.034 & 0.948 & 0.487 & 1.0000000 & \\
AWQ-int4- GPTQ-int8       &  0.013 & 0.035 & 1.013 & 0.503 & 1.0000000 & \\
AWQ-int4- GPTQ-int4       & -0.003 & 0.034 & 0.997 & 0.499 & 1.0000000 & \\
AWQ-int4- BNB-nf4        &  0.041 & 0.034 & 1.042 & 0.510 & 1.0000000 & \\

BNB-nf4 - FP8        & -0.094 & 0.034 & 0.910 & 0.476 & 0.0917821 & \\
BNB-nf4 - GPTQ-int8      & -0.028 & 0.035 & 0.972 & 0.493 & 1.0000000 & \\
BNB-nf4 - GPTQ-int4      & -0.044 & 0.034 & 0.957 & 0.489 & 1.0000000 & \\

FP8 - GPTQ-int8       &  0.066 & 0.035 & 1.068 & 0.517 & 0.8819317 & \\
FP8 - GPTQ-int4       &  0.051 & 0.034 & 1.052 & 0.513 & 1.0000000 & \\

GPTQ-int4 - GPTQ-int8     &  0.016 & 0.035 & 1.016 & 0.504 & 1.0000000 & \\
\bottomrule
\end{tabular}
}
\caption{\textbf{\nocha{}}: Post-hoc comparisons between quantization methods for \textbf{accuracy} (Table~\ref{tab:nocha-stats}) across all context lengths using Bonferroni adjustments for multiple pairwise comparisons. The probability values refer to the likelihood that the quantization method listed first in each contrast is more accurate than the one listed second (i.e., a value of 0.5 suggests that both methods are comparable in terms of accuracy).}

\label{tab:nocha-posthoc-model}
\vspace{0.5em}
\end{table*}

\begin{table*}[ht]
\centering
\resizebox{0.7\textwidth}{!}{%
\begin{tabular}{l
                S[table-format=2.6]
                S[table-format=1.6]
                S[table-format=1.6]
                S[table-format=1.6]
                S[table-format=1.6]
                c}
\toprule
\multicolumn{7}{c}{\texttt{glmer(Binary\_label $\sim$ Quantization * Context\_length + (1|Model),}} \\
\multicolumn{7}{c}{\texttt{~~~~data = data, family = binomial)}} \\
\bottomrule
\textsc{Predictors} & \textsc{Estimate} & \textsc{OR} & \textsc{CI (lower)} & \textsc{CI (upper)} & \texttt{p-value} & \\
\midrule
Intercept                                       & 0.587889 & 1.800185 & 1.440018 & 2.250434 & 0.000 & *** \\
FP8                                 & 0.126486 & 1.134834 & 0.955165 & 1.348300 & 0.1503   &     \\
GPTQ-int8                         & 0.015436 & 1.015556 & 0.856346 & 1.204367 & 0.8592   &     \\
AWQ-int4                            & 0.021609 & 1.021845 & 0.862439 & 1.210713 & 0.8028   &     \\
GPTQ-int4                         & -0.027324 & 0.973046 & 0.821471 & 1.152590 & 0.7518   &     \\
BNB-nf4                             & -0.090726 & 0.913268 & 0.771689 & 1.080821 & 0.2911   &     \\
Context length$>$75k               & -0.159741 & 0.852365 & 0.749282 & 0.969630 & 0.0151   & *   \\
FP8:$>$75k  & -0.091412 & 0.912642 & 0.757431 & 1.099657 & 0.3365   &     \\
GPTQ-int8:$>$75k & -0.039194 & 0.961564 & 0.798941 & 1.157289 & 0.6784   &     \\
AWQ-int4:$>$75k & -0.030805 & 0.969665 & 0.806970 & 1.165160 & 0.7424   &     \\
GPTQ-int4:$>$75k & 0.029694 & 1.030140 & 0.857525 & 1.237501 & 0.7510   &     \\
BNB-nf4:$>$75k & 0.052923 & 1.054348 & 0.878413 & 1.265521 & 0.5699   &     \\
\bottomrule
\textsc{Random Effects} & & & & & & \\
\bottomrule
\\
$\sigma^2$ (residual variance) & \multicolumn{5}{l}{1.000} & \\
$\tau_{00}$ (Model) & \multicolumn{5}{l}{0.047} & \\
ICC (Model) & \multicolumn{5}{l}{0.045} & \\
N (Model) & \multicolumn{5}{l}{5} & \\
Observations & \multicolumn{5}{l}{42781} & \\
\bottomrule
\\
$R^2$ (marginal) & \multicolumn{5}{l}{0.016} & \\
$R^2$ (conditional) & \multicolumn{5}{l}{0.001} & \\
\bottomrule
\end{tabular}
}
\caption{\textbf{\nocha{}}: Summary of generalized linear mixed model with \textbf{quantization}, \textbf{context length}, and their interaction as predictors of \textbf{accuracy}: \texttt{glmer(Binary\_label $\sim$ Quantization * Context\_length + (1|Model), data = data, family = binomial)}. The quantization method "BF16" at context length $\leq$75k was set as the reference level (intercept), with \texttt{Model} set as the grouping factor for a random intercept to account for variability across models. See Table~\ref{tab:nocha-posthoc-context75k} for post-hoc analysis.}
\vspace{0.5em}
\label{tab:nocha-context-split}
\end{table*}

\begin{table*}[ht]
\centering
\resizebox{0.7\textwidth}{!}{
\begin{tabular}{c
                l
                S[table-format=1.3]
                S[table-format=1.3]
                S[table-format=1.3]
                S[table-format=1.3]
                S[table-format=1.6]
                c}
\toprule
\textsc{Context} & \textsc{Contrast} & \textsc{Estimate} & \textsc{SE} & \textsc{OR} & \textsc{Probability} & \texttt{p-value} & \\
\midrule
$<$75k & BF16 - (AWQ-int4)          & -0.022 & 0.087 & 0.979 & 0.495 & 1.000000 & \\
$<$75k & BF16 - (BNB-nf4)          &  0.091 & 0.086 & 1.095 & 0.523 & 1.000000 & \\
$<$75k & BF16 - FP8           & -0.126 & 0.088 & 0.881 & 0.468 & 1.000000 & \\
$<$75k & BF16 - (GPTQ-int4)         &  0.027 & 0.086 & 1.028 & 0.507 & 1.000000 & \\
$<$75k & BF16 - (GPTQ-int8)         & -0.015 & 0.087 & 0.985 & 0.496 & 1.000000 & \\
$<$75k & (AWQ-int4)- (BNB-nf4)           &  0.112 & 0.088 & 1.119 & 0.528 & 1.000000 & \\
$<$75k & (AWQ-int4)- FP8            & -0.105 & 0.090 & 0.900 & 0.474 & 1.000000 & \\
$<$75k & (AWQ-int4)- (GPTQ-int4)          &  0.049 & 0.089 & 1.050 & 0.512 & 1.000000 & \\
$<$75k & (AWQ-int4)- (GPTQ-int8)          &  0.006 & 0.089 & 1.006 & 0.502 & 1.000000 & \\
$<$75k & (BNB-nf4) - FP8           & -0.217 & 0.089 & 0.805 & 0.446 & 0.215798 & \\
$<$75k & (BNB-nf4) - (GPTQ-int4)         & -0.063 & 0.088 & 0.939 & 0.484 & 1.000000 & \\
$<$75k & (BNB-nf4) - (GPTQ-int8)         & -0.106 & 0.088 & 0.899 & 0.473 & 1.000000 & \\
$<$75k & FP8 - (GPTQ-int4)          &  0.154 & 0.089 & 1.166 & 0.538 & 1.000000 & \\
$<$75k & FP8 - (GPTQ-int8)          &  0.111 & 0.090 & 1.117 & 0.528 & 1.000000 & \\
$<$75k & (GPTQ-int4) - (GPTQ-int8)        & -0.043 & 0.089 & 0.958 & 0.489 & 1.000000 & \\
$>$75k & BF16 - (AWQ-int4)       &  0.009 & 0.037 & 1.009 & 0.502 & 1.000000 & \\
$>$75k & BF16 - (BNB-nf4)       &  0.038 & 0.037 & 1.039 & 0.509 & 1.000000 & \\
$>$75k & BF16 - FP8        & -0.035 & 0.037 & 0.966 & 0.491 & 1.000000 & \\
$>$75k & BF16 - (GPTQ-int4)      & -0.002 & 0.037 & 0.998 & 0.499 & 1.000000 & \\
$>$75k & BF16 - (GPTQ-int8)      &  0.024 & 0.038 & 1.024 & 0.506 & 1.000000 & \\
$>$75k & (AWQ-int4)- (BNB-nf4)        &  0.029 & 0.037 & 1.029 & 0.507 & 1.000000 & \\
$>$75k & (AWQ-int4)- FP8         & -0.044 & 0.037 & 0.957 & 0.489 & 1.000000 & \\
$>$75k & (AWQ-int4)- (GPTQ-int4)       & -0.012 & 0.037 & 0.989 & 0.497 & 1.000000 & \\
$>$75k & (AWQ-int4)- (GPTQ-int8)       &  0.015 & 0.038 & 1.015 & 0.504 & 1.000000 & \\
$>$75k & (BNB-nf4) - FP8        & -0.073 & 0.037 & 0.930 & 0.482 & 0.757545 & \\
$>$75k & (BNB-nf4) - (GPTQ-int4)      & -0.040 & 0.037 & 0.961 & 0.490 & 1.000000 & \\
$>$75k & (BNB-nf4) - (GPTQ-int8)      & -0.014 & 0.038 & 0.986 & 0.496 & 1.000000 & \\
$>$75k & FP8 - (GPTQ-int4)       &  0.033 & 0.037 & 1.033 & 0.508 & 1.000000 & \\
$>$75k & FP8 - (GPTQ-int8)       &  0.059 & 0.038 & 1.061 & 0.515 & 1.000000 & \\
$>$75k & (GPTQ-int4) - (GPTQ-int8)     &  0.026 & 0.038 & 1.026 & 0.507 & 1.000000 & \\
\bottomrule
\end{tabular}
}
\caption{\textbf{\nocha{}}: Post-hoc comparisons between quantization methods for \textbf{accuracy} (Table~\ref{tab:nocha-context-split}) across context lengths below and above 75k tokens using Bonferroni adjustments for multiple pairwise comparisons. The probability values refer to the likelihood that the quantization method listed first in each contrast is more accurate than the one listed second (i.e., a value of 0.5 suggests that both methods are comparable in terms of accuracy).}
\label{tab:nocha-posthoc-context75k}
\vspace{0.5em}
\end{table*}

%%%%%%%%%%%%%%%%% Veriscore tables

\begin{table*}[ht]
\centering
\resizebox{0.7\textwidth}{!}{
\begin{tabular}{l
                S[table-format=1.3]
                S[table-format=1.3]
                S[table-format=1.3]
                S[table-format=1.3]
                S[table-format=1.5]
                c}
\toprule
\multicolumn{7}{c}{\texttt{lmer(log\_veriscore $\sim$ Quantization + (1|topic) + (1|Model),}} \\
\multicolumn{7}{c}{\texttt{~~~~data = data)}} \\
\bottomrule
\textsc{Predictors} & \textsc{Estimate} & \textsc{CI (lower)} & \textsc{CI (upper)} & \textsc{SE} & \texttt{p-value} & \\
\midrule
Intercept            &  2.183 & 1.754 & 2.611 & 0.219 & 0.000 & *** \\
AWQ-int4             & -0.079 & -0.181 & 0.023 & 0.052 & 0.129 &     \\
BNB-nf4             & -0.228 & -0.330 & -0.126 & 0.052 & 0.000 & *** \\
FP8                  &  0.074 & -0.028 & 0.176 & 0.052 & 0.156 &     \\
GPTQ-int4            & -0.127 & -0.229 & -0.025 & 0.052 & 0.015 & *   \\
GPTQ-int8            & -0.003 & -0.105 & 0.099 & 0.052 & 0.956 &     \\
\bottomrule
\textsc{Random Effects} & & & & & & \\
\bottomrule
\\
$\sigma^2$ (Residual)          & \multicolumn{5}{l}{0.679} & \\
$\tau_{00}$ (Model)            & \multicolumn{5}{l}{0.118} & \\
$\tau_{00}$ (Topic)            & \multicolumn{5}{l}{2.274} & \\
ICC (Model)                    & \multicolumn{5}{l}{0.039} & \\
ICC (Topic)                    & \multicolumn{5}{l}{0.740} & \\
N (Model)                      & \multicolumn{5}{l}{5} & \\
N (Topic)                      & \multicolumn{5}{l}{100} & \\
Observations                   & \multicolumn{5}{l}{3000} & \\
\bottomrule
\\
$R^2$ (marginal)               & \multicolumn{5}{l}{0.003} & \\
$R^2$ (conditional)            & \multicolumn{5}{l}{0.780} & \\
\bottomrule
\end{tabular}
}
\caption{\factscore: Summary of linear mixed-effects model with \textbf{quantization} as the predictor of \textbf{\veriscore{}}: \texttt{lmer(log\_veriscore $\sim$ Quantization + (1|topic) + (1|Model), data = data)}. The quantization method "BF16" was set as the reference level (intercept), with \texttt{topic} and \texttt{Model} set as grouping factors for random intercepts to account for variability. See Table~\ref{tab:veriscore-quant-posthoc} for post-hoc analysis.}
\label{tab:veriscore-quant-main}
\vspace{0.5em}
\end{table*}

\begin{table*}[ht]
\centering
\resizebox{0.7\textwidth}{!}{
\begin{tabular}{l
                S[table-format=1.3]
                S[table-format=1.3]
                S[table-format=1.3]
                S[table-format=1.3]
                S[table-format=1.3]
                c}
\toprule
\textsc{Contrast} & \textsc{Estimate} & \textsc{SE} & \textsc{OR} & \textsc{Probability} & \texttt{p-value} & \\
\midrule
BF16 - (AWQ-int4)         &  0.079 & 0.052 & 1.082 & 0.520 & 1.000 & \\
BF16 - (BNB-nf4)         &  0.228 & 0.052 & 1.256 & 0.557 & 0.000 & *** \\
BF16 - FP8                & -0.074 & 0.052 & 0.929 & 0.482 & 1.000 & \\
BF16 - (GPTQ-int4)        &  0.127 & 0.052 & 1.135 & 0.532 & 0.227 & \\
BF16 - (GPTQ-int8)        &  0.003 & 0.052 & 1.003 & 0.501 & 1.000 & \\

(AWQ-int4) - (BNB-nf4)    &  0.149 & 0.052 & 1.161 & 0.537 & 0.063 & \\
(AWQ-int4) - FP8           & -0.153 & 0.052 & 0.858 & 0.462 & 0.050 & * \\
(AWQ-int4) - (GPTQ-int4)   &  0.048 & 0.052 & 1.049 & 0.512 & 1.000 & \\
(AWQ-int4) - (GPTQ-int8)   & -0.076 & 0.052 & 0.927 & 0.481 & 1.000 & \\

(BNB-nf4) - FP8           & -0.302 & 0.052 & 0.739 & 0.425 & 0.000 & *** \\
(BNB-nf4) - (GPTQ-int4)   & -0.102 & 0.052 & 0.903 & 0.474 & 0.770 & \\
(BNB-nf4) - (GPTQ-int8)   & -0.225 & 0.052 & 0.798 & 0.444 & 0.000 & *** \\

FP8 - (GPTQ-int4)          &  0.201 & 0.052 & 1.222 & 0.550 & 0.002 & ** \\
FP8 - (GPTQ-int8)          &  0.077 & 0.052 & 1.080 & 0.519 & 1.000 & \\

(GPTQ-int4) - (GPTQ-int8)  & -0.124 & 0.052 & 0.884 & 0.469 & 0.263 & \\
\bottomrule
\end{tabular}
}
\caption{\textbf{\factscore{}}: Post-hoc comparisons between quantization methods for \textbf{\veriscore{}} (Table~\ref{tab:veriscore-quant-main}) using Bonferroni adjustments for multiple pairwise comparisons. The odds ratios and probabilities are derived from linear model estimates. A probability near 0.5 indicates that the two methods are comparably accurate (in \veriscore{}).}
\label{tab:veriscore-quant-posthoc}
\vspace{0.5em}
\end{table*}

\begin{table*}[ht]
\centering
\resizebox{0.7\textwidth}{!}{
\begin{tabular}{l
                S[table-format=1.3]
                S[table-format=1.3]
                S[table-format=1.3]
                S[table-format=1.3]
                S[table-format=1.3]
                c}
\toprule
\multicolumn{7}{c}{\texttt{glmer(no\_claims $\sim$ Quantization + (1|Model)}} \\
\multicolumn{7}{c}{\texttt{+ (1|topic), data = data, family = binomial)}} \\
\bottomrule
\textsc{Predictors} & \textsc{Estimate} & \textsc{OR} & \textsc{CI (lower)} & \textsc{CI (upper)} & \texttt{p-value} & \\
\midrule
Intercept                  & -2.961 & 0.052 & 0.010 & 0.262 & 0.000 & *** \\
AWQ-int4                   &  0.351 & 1.421 & 0.941 & 2.144 & 0.095 &     \\
BNB-nf4                   &  0.831 & 2.295 & 1.522 & 3.459 & 0.000 & *** \\
FP8                        & -0.048 & 0.954 & 0.630 & 1.444 & 0.822 &     \\
GPTQ-int4                  &  0.466 & 1.594 & 1.057 & 2.404 & 0.026 & *   \\
GPTQ-int8                  & -0.024 & 0.977 & 0.645 & 1.479 & 0.911 &     \\
\bottomrule
\textsc{Random Effects} & & & & & & \\
\bottomrule
\\
$\sigma^2$ (residual variance, logistic) & \multicolumn{5}{l}{3.290} & \\
$\tau_{00}$ (Model)                      & \multicolumn{5}{l}{2.669} & \\
$\tau_{00}$ (Topic)                      & \multicolumn{5}{l}{11.401} & \\
ICC (Model)                              & \multicolumn{5}{l}{0.154} & \\
ICC (Topic)                              & \multicolumn{5}{l}{0.657} & \\
N (Model)                                & \multicolumn{5}{l}{5} & \\
N (Topic)                                & \multicolumn{5}{l}{100} & \\
Observations                             & \multicolumn{5}{l}{3000} & \\
\bottomrule
\\
$R^2$ (marginal)                         & \multicolumn{5}{l}{0.006} & \\
$R^2$ (conditional)                      & \multicolumn{5}{l}{0.812} & \\
\bottomrule
\end{tabular}
}
\caption{\textbf{\factscore{}}: Summary of generalized linear mixed model with \textbf{quantization} as the predictor of \textbf{no claim rate}: \texttt{glmer(no\_claims $\sim$ Quantization + (1|Model) + (1|topic), data = data, family = binomial)}. The quantization method "BF16" was set as the reference level (intercept), with \texttt{Model} and \texttt{topic} set as grouping factors for random intercepts to account for variability across models and topics. See Table~\ref{tab:veriscore-empty_claims-posthoc} for post-hoc analysis.}
\label{tab:veriscore-empty_claims-main}
\vspace{0.5em}
\end{table*}

\begin{table*}[ht]
\centering
\resizebox{0.7\textwidth}{!}{
\begin{tabular}{l
                S[table-format=1.3]
                S[table-format=1.3]
                S[table-format=1.3]
                S[table-format=1.3]
                S[table-format=1.3]
                c}
\toprule
\textsc{Contrast} & \textsc{Estimate} & \textsc{SE} & \textsc{OR} & \textsc{Probability} & \texttt{p-value} & \\
\midrule
BF16 - (AWQ-int4)         & -0.351 & 0.210 & 0.704 & 0.413 & 1.000 &     \\
BF16 - (BNB-nf4)         & -0.831 & 0.209 & 0.436 & 0.304 & 0.001 & **  \\
BF16 - FP8                &  0.048 & 0.212 & 1.049 & 0.512 & 1.000 &     \\
BF16 - (GPTQ-int4)        & -0.466 & 0.210 & 0.627 & 0.385 & 0.392 &     \\
BF16 - (GPTQ-int8)        &  0.024 & 0.212 & 1.024 & 0.506 & 1.000 &     \\

(AWQ-int4) - (BNB-nf4)    & -0.479 & 0.206 & 0.619 & 0.382 & 0.302 &     \\
(AWQ-int4) - FP8           &  0.399 & 0.210 & 1.490 & 0.598 & 0.871 &     \\
(AWQ-int4) - (GPTQ-int4)   & -0.115 & 0.207 & 0.891 & 0.471 & 1.000 &     \\
(AWQ-int4) - (GPTQ-int8)   &  0.375 & 0.210 & 1.455 & 0.593 & 1.000 &     \\

(BNB-nf4) - FP8           &  0.878 & 0.210 & 2.406 & 0.706 & 0.000 & *** \\
(BNB-nf4) - (GPTQ-int4)   &  0.364 & 0.206 & 1.439 & 0.590 & 1.000 &     \\
(BNB-nf4) - (GPTQ-int8)   &  0.854 & 0.210 & 2.350 & 0.701 & 0.001 & **  \\

FP8 - (GPTQ-int4)          & -0.514 & 0.210 & 0.598 & 0.374 & 0.217 &     \\
FP8 - (GPTQ-int8)          & -0.024 & 0.212 & 0.976 & 0.494 & 1.000 &     \\

(GPTQ-int4) - (GPTQ-int8)  &  0.490 & 0.210 & 1.632 & 0.620 & 0.293 &     \\
\bottomrule
\end{tabular}
}
\caption{\textbf{\factscore{}}: Post-hoc comparisons between quantization methods for \textbf{no claim rate} (Table~\ref{tab:veriscore-empty_claims-main}) using Bonferroni adjustments for multiple pairwise comparisons. Results are reported on the log-odds ratio scale. The probability values refer to the likelihood that the quantization method listed first in each contrast abstains more than the one listed second (i.e., a value of 0.5 suggests comparable rate of empty claims).}
\label{tab:veriscore-empty_claims-posthoc}
\vspace{0.5em}
\end{table*}

\begin{table*}[ht]
\centering
\resizebox{0.7\textwidth}{!}{
\begin{tabular}{l
                S[table-format=1.3]
                S[table-format=1.3]
                S[table-format=1.3]
                S[table-format=1.3]
                S[table-format=1.3]
                c}
\toprule
\multicolumn{7}{c}{\texttt{lmer(log\_veriscore $\sim$ Quantization * cat + (1|Model)}} \\
\multicolumn{7}{c}{\texttt{+ (1|topic), data = data)}} \\
\bottomrule
\textsc{Predictors} & \textsc{Estimate} & \textsc{CI (lower)} & \textsc{CI (upper)} & \textsc{SE} & \texttt{p-value} & \\
\midrule
Intercept                               &  2.042 & 1.639 & 2.445 & 0.206 & 0.000 & *** \\
AWQ-int4                                & -0.142 & -0.320 & 0.035 & 0.091 & 0.116 &     \\
BNB-nf4                                & -0.430 & -0.607 & -0.252 & 0.091 & 0.000 & *** \\
FP8                                     &  0.089 & -0.089 & 0.267 & 0.091 & 0.326 &     \\
GPTQ-int4                               & -0.197 & -0.374 & -0.019 & 0.091 & 0.030 & *   \\
GPTQ-int8                               &  0.023 & -0.155 & 0.200 & 0.091 & 0.803 &     \\
very freq                               &  1.831 & 1.456 & 2.205 & 0.191 & 0.000 & *** \\
very rare                               & -1.460 & -1.837 & -1.082 & 0.193 & 0.000 & *** \\
AWQ-int4 × very freq                    &  0.014 & -0.236 & 0.263 & 0.127 & 0.915 &     \\
BNB-nf4 × very freq                    &  0.354 & 0.105 & 0.604 & 0.127 & 0.005 & **  \\
FP8 × very freq                         & -0.064 & -0.314 & 0.185 & 0.127 & 0.612 &     \\
GPTQ-int4 × very freq                   &  0.084 & -0.166 & 0.333 & 0.127 & 0.511 &     \\
GPTQ-int8 × very freq                   & -0.060 & -0.309 & 0.189 & 0.127 & 0.638 &     \\
AWQ-int4 × very rare                    &  0.177 & -0.074 & 0.428 & 0.128 & 0.166 &     \\
BNB-nf4 × very rare                    &  0.245 & -0.006 & 0.496 & 0.128 & 0.056 & .   \\
FP8 × very rare                         &  0.021 & -0.231 & 0.272 & 0.128 & 0.873 &     \\
GPTQ-int4 × very rare                   &  0.126 & -0.125 & 0.377 & 0.128 & 0.327 &     \\
GPTQ-int8 × very rare                   & -0.016 & -0.267 & 0.236 & 0.128 & 0.903 &     \\
\bottomrule
\textsc{Random Effects} & & & & & & \\
\bottomrule
\\
$\sigma^2$ (Residual)                   & \multicolumn{5}{l}{0.677} & \\
$\tau_{00}$ (Model)                    & \multicolumn{5}{l}{0.118} & \\
$\tau_{00}$ (Topic)                    & \multicolumn{5}{l}{0.477} & \\
ICC (Model)                            & \multicolumn{5}{l}{0.093} & \\
ICC (Topic)                            & \multicolumn{5}{l}{0.375} & \\
N (Model)                              & \multicolumn{5}{l}{5} & \\
N (Topic)                              & \multicolumn{5}{l}{100} & \\
Observations                           & \multicolumn{5}{l}{3000} & \\
\bottomrule
\\
$R^2$ (marginal)                       & \multicolumn{5}{l}{0.586} & \\
$R^2$ (conditional)                    & \multicolumn{5}{l}{0.780} & \\
\bottomrule
\end{tabular}
}
\caption{\textbf{\factscore{}}: Summary of linear mixed-effects model with \textbf{quantization} and \textbf{category} as predictors of \textbf{veriscore{}}: \texttt{lmer(log\_veriscore $\sim$ Quantization * cat + (1|Model) + (1|topic), data = data)}. The quantization method "BF16" and category "very freq" were set as reference levels. Random intercepts were included for \texttt{Model} and \texttt{topic} to account for grouping variability. See Table~\ref{tab:veriscore-category-posthoc} for post-hoc analysis.}
\label{tab:veriscore-category-main}
\vspace{0.5em}
\end{table*}

\begin{table*}[ht]
\centering
\resizebox{0.7\textwidth}{!}{
\begin{tabular}{l
                l
                S[table-format=1.3]
                S[table-format=1.3]
                S[table-format=1.3]
                S[table-format=1.3]
                c}
\toprule
\textsc{Category} & \textsc{Contrast} & \textsc{Estimate} & \textsc{SE} & \textsc{OR} & \texttt{p-value} & \\
\midrule
medium     & BF16 - (AWQ-int4)            &  0.142 & 0.091 & 1.153 & 1.000 & \\
medium     & BF16 - (BNB-nf4)            &  0.430 & 0.091 & 1.537 & 0.000 & *** \\
medium     & BF16 - FP8                   & -0.089 & 0.091 & 0.915 & 1.000 & \\
medium     & BF16 - (GPTQ-int4)           &  0.197 & 0.091 & 1.217 & 0.451 & \\
medium     & BF16 - (GPTQ-int8)           & -0.023 & 0.091 & 0.977 & 1.000 & \\
medium     & (AWQ-int4) - (BNB-nf4)      &  0.287 & 0.091 & 1.333 & 0.023 & * \\
medium     & (AWQ-int4) - FP8             & -0.231 & 0.091 & 0.794 & 0.161 & \\
medium     & (AWQ-int4) - (GPTQ-int4)     &  0.054 & 0.091 & 1.056 & 1.000 & \\
medium     & (AWQ-int4) - (GPTQ-int8)     & -0.165 & 0.091 & 0.848 & 1.000 & \\
medium     & (BNB-nf4) - FP8             & -0.519 & 0.091 & 0.595 & 0.000 & *** \\
medium     & (BNB-nf4) - (GPTQ-int4)     & -0.233 & 0.091 & 0.792 & 0.153 & \\
medium     & (BNB-nf4) - (GPTQ-int8)     & -0.452 & 0.091 & 0.636 & 0.000 & *** \\
medium     & FP8 - (GPTQ-int4)            &  0.286 & 0.091 & 1.331 & 0.025 & * \\
medium     & FP8 - (GPTQ-int8)            &  0.066 & 0.091 & 1.069 & 1.000 & \\
medium     & (GPTQ-int4) - (GPTQ-int8)    & -0.219 & 0.091 & 0.803 & 0.234 & \\
very freq  & BF16 - (AWQ-int4)            &  0.129 & 0.089 & 1.137 & 1.000 & \\
very freq  & BF16 - (BNB-nf4)            &  0.075 & 0.089 & 1.078 & 1.000 & \\
very freq  & BF16 - FP8                   & -0.025 & 0.089 & 0.975 & 1.000 & \\
very freq  & BF16 - (GPTQ-int4)           &  0.113 & 0.089 & 1.120 & 1.000 & \\
very freq  & BF16 - (GPTQ-int8)           &  0.037 & 0.089 & 1.038 & 1.000 & \\
very freq  & (AWQ-int4) - (BNB-nf4)      & -0.053 & 0.089 & 0.948 & 1.000 & \\
very freq  & (AWQ-int4) - FP8             & -0.153 & 0.089 & 0.858 & 1.000 & \\
very freq  & (AWQ-int4) - (GPTQ-int4)     & -0.016 & 0.089 & 0.985 & 1.000 & \\
very freq  & (AWQ-int4) - (GPTQ-int8)     & -0.091 & 0.089 & 0.913 & 1.000 & \\
very freq  & (BNB-nf4) - FP8             & -0.100 & 0.089 & 0.905 & 1.000 & \\
very freq  & (BNB-nf4) - (GPTQ-int4)     &  0.038 & 0.089 & 1.038 & 1.000 & \\
very freq  & (BNB-nf4) - (GPTQ-int8)     & -0.038 & 0.089 & 0.963 & 1.000 & \\
very freq  & FP8 - (GPTQ-int4)            &  0.138 & 0.089 & 1.148 & 1.000 & \\
very freq  & FP8 - (GPTQ-int8)            &  0.062 & 0.089 & 1.064 & 1.000 & \\
very freq  & (GPTQ-int4) - (GPTQ-int8)    & -0.076 & 0.089 & 0.927 & 1.000 & \\
very rare  & BF16 - (AWQ-int4)            & -0.035 & 0.091 & 0.966 & 1.000 & \\
very rare  & BF16 - (BNB-nf4)            &  0.185 & 0.091 & 1.203 & 0.621 & \\
very rare  & BF16 - FP8                   & -0.110 & 0.091 & 0.896 & 1.000 & \\
very rare  & BF16 - (GPTQ-int4)           &  0.071 & 0.091 & 1.073 & 1.000 & \\
very rare  & BF16 - (GPTQ-int8)           & -0.007 & 0.091 & 0.993 & 1.000 & \\
very rare  & (AWQ-int4) - (BNB-nf4)      &  0.220 & 0.091 & 1.246 & 0.229 & \\
very rare  & (AWQ-int4) - FP8             & -0.075 & 0.091 & 0.928 & 1.000 & \\
very rare  & (AWQ-int4) - (GPTQ-int4)     &  0.106 & 0.091 & 1.112 & 1.000 & \\
very rare  & (AWQ-int4) - (GPTQ-int8)     &  0.028 & 0.091 & 1.029 & 1.000 & \\
very rare  & (BNB-nf4) - FP8             & -0.294 & 0.091 & 0.745 & 0.018 & * \\
very rare  & (BNB-nf4) - (GPTQ-int4)     & -0.114 & 0.091 & 0.892 & 1.000 & \\
very rare  & (BNB-nf4) - (GPTQ-int8)     & -0.192 & 0.091 & 0.826 & 0.514 & \\
very rare  & FP8 - (GPTQ-int4)            &  0.180 & 0.091 & 1.197 & 0.697 & \\
very rare  & FP8 - (GPTQ-int8)            &  0.103 & 0.091 & 1.109 & 1.000 & \\
very rare  & (GPTQ-int4) - (GPTQ-int8)    & -0.078 & 0.091 & 0.925 & 1.000 & \\
\bottomrule
\end{tabular}
}
\caption{\textbf{\factscore{}}: Post-hoc comparisons between quantization methods for \textbf{\veriscore{}} across \textbf{category} using Bonferroni adjustments for multiple pairwise comparisons. The odds ratios are derived from linear mixed-effects estimates exponentiated for interpretability.}
\label{tab:veriscore-category-posthoc}
\vspace{0.5em}
\end{table*}

%%%%%%%%%%%%%%% CS4 tables
\begin{table*}[ht]
\centering
\begin{tabular}{l
                S[table-format=1.3]
                S[table-format=1.3]
                S[table-format=1.3]
                S[table-format=1.3]
                S[table-format=1.5]
                c}
\toprule
\multicolumn{7}{c}{\makecell[c]{\texttt{glmer(cbind(Success, Failures) $\sim$ Quantization + (1|Model),}\\
\texttt{data = data, family = binomial)}}} \\
\bottomrule
\textsc{Predictors} & \textsc{Estimate} & \textsc{OR} & \textsc{CI (lower)} & \textsc{CI (upper)} & \texttt{p-value} & \\
\midrule
Intercept            &  1.573 & 4.822 & 3.473 & 6.694 & 0.000 & *** \\
AWQ-int4             &  0.022 & 1.022 & 0.966 & 1.080 & 0.45032  &     \\
BNB-nf4              & -0.085 & 0.918 & 0.869 & 0.970 & 0.00237  & **  \\
FP8                  & -0.152 & 0.859 & 0.814 & 0.907 & 0.00000  & *** \\
GPTQ-int4            &  0.051 & 1.052 & 0.995 & 1.113 & 0.07544  & .   \\
GPTQ-int8            & -0.016 & 0.984 & 0.931 & 1.040 & 0.57107  &     \\
\bottomrule
\textsc{Random Effects} & & & & & & \\
\bottomrule
\\
$\sigma^2$ (residual variance) & \multicolumn{5}{l}{1.000} & \\
$\tau_{00}$ (Model) & \multicolumn{5}{l}{0.138} & \\
ICC (Model) & \multicolumn{5}{l}{0.121} & \\
N (Model) & \multicolumn{5}{l}{5} & \\
Observations & \multicolumn{5}{l}{4500} & \\
\bottomrule
\\
$R^2$ (marginal) & \multicolumn{5}{l}{0.016} & \\
$R^2$ (conditional) & \multicolumn{5}{l}{0.500} & \\
\bottomrule
\end{tabular}
\caption{\textbf{\csfour{}}: Summary of generalized linear mixed model with \textbf{quantization} as the predictor of \textbf{accuracy}: \texttt{glmer(cbind(Success, Failures) $\sim$ Quantization + (1|Model), data = data, family = binomial)}. The quantization method "BF16" was set as the reference level (intercept), with \texttt{Model} set as the grouping factor for a random intercept to account for variability across models. Refer to Table~\ref{tab:cs4-posthoc-model} for post-hoc analysis.}
\label{tab:csfour-stats-model-quant}
\vspace{0.5em}
\end{table*}

\begin{table*}[ht]
\centering
\begin{tabular}{l
                S[table-format=1.3]
                S[table-format=1.3]
                S[table-format=1.3]
                S[table-format=1.3]
                S[table-format=1.5]
                c}
\toprule
\textsc{Contrast} & \textsc{Estimate} & \textsc{SE} & \textsc{OR} & \textsc{Probability} & \texttt{p-value} & \\
\midrule
BF16 - (AWQ-int4)         & -0.022 & 0.028 & 0.979 & 0.495 & 1.00e+00 &     \\
BF16 - (BNB-nf4)          &  0.085 & 0.028 & 1.089 & 0.521 & 0.035 & *   \\
BF16 - FP8                &  0.152 & 0.028 & 1.164 & 0.538 & 0.000 & *** \\
BF16 - (GPTQ-int4)        & -0.051 & 0.029 & 0.950 & 0.487 & 1.00e+00 &     \\
BF16 - (GPTQ-int8)        &  0.016 & 0.028 & 1.016 & 0.504 & 1.00e+00 &     \\

(AWQ-int4) - (BNB-nf4)     &  0.107 & 0.028 & 1.113 & 0.527 & 0.002 & **  \\
(AWQ-int4) - FP8           &  0.173 & 0.028 & 1.189 & 0.543 & 0.000 & *** \\
(AWQ-int4) - (GPTQ-int4)   & -0.029 & 0.029 & 0.971 & 0.493 & 1.00e+00 &     \\
(AWQ-int4) - (GPTQ-int8)   &  0.038 & 0.028 & 1.038 & 0.509 & 1.00e+00 &     \\

(BNB-nf4) - FP8            &  0.067 & 0.027 & 1.069 & 0.517 & 0.222 &     \\
(BNB-nf4) - (GPTQ-int4)    & -0.136 & 0.028 & 0.873 & 0.466 & 0.000 & *** \\
(BNB-nf4) - (GPTQ-int8)    & -0.069 & 0.028 & 0.933 & 0.483 & 0.201 &     \\

FP8 - (GPTQ-int4)          & -0.203 & 0.028 & 0.816 & 0.449 & 0.000 & *** \\
FP8 - (GPTQ-int8)          & -0.136 & 0.028 & 0.873 & 0.466 & 0.000 & *** \\

(GPTQ-int4) - (GPTQ-int8)  &  0.067 & 0.029 & 1.069 & 0.517 & 0.286 &     \\
\bottomrule
\end{tabular}
\caption{\textbf{\csfour{}}: Post-hoc comparisons between quantization methods for \textbf{accuracy} (Table~\ref{tab:csfour-stats-model-quant}) using Bonferroni adjustments for multiple pairwise comparisons. The probability values refer to the likelihood that the quantization method listed first in each contrast is more accurate than the one listed second (i.e., a value of 0.5 suggests that both methods are comparable in terms of accuracy).}
\label{tab:cs4-posthoc-model}
\vspace{0.5em}
\end{table*}

\begin{table*}[ht]
\centering
\resizebox{\textwidth}{!}{%
\begin{tabular}{l
                S[table-format=2.6]
                S[table-format=2.6]
                S[table-format=1.6]
                S[table-format=2.6]
                S[table-format=1.6]
                c}
\toprule
\multicolumn{7}{c}{\makecell[c]{\texttt{glmer(cbind(Success, Failures) $\sim$ Quantization * Input\_constraints + (1|Model),}\\
\texttt{data = data, family = binomial)}}} \\

\bottomrule
\textsc{Predictors} & \textsc{Estimate} & \textsc{OR} & \textsc{CI (lower)} & \textsc{CI (upper)} & \texttt{p-value} & \\
\midrule
Intercept                                &  2.546482 & 12.762130 &  8.774426 & 18.562122 & 0.0000 & *** \\
AWQ-int4                                 &  0.075677 &  1.078614 &  0.836472 &  1.390851 & 0.560    &     \\
BNB-nf4                                  & -0.116433 &  0.890090 &  0.697100 &  1.136509 & 0.350    &     \\
FP8                                      & -0.032167 &  0.968345 &  0.755271 &  1.241532 & 0.800    &     \\
GPTQ-int4                                &  0.203962 &  1.226251 &  0.943980 &  1.592927 & 0.126    &     \\
GPTQ-int8                                & -0.101537 &  0.903448 &  0.707105 &  1.154310 & 0.417    &     \\
23 constraints                           & -0.757714 &  0.468737 &  0.387096 &  0.567596 & 0.0000 & *** \\
39 constraints                           & -1.200708 &  0.300981 &  0.250515 &  0.361613 & 0.0000 & *** \\
AWQ-int4:23 constraints                  & -0.085379 &  0.918164 &  0.697849 &  1.208035 & 0.542    &     \\
BNB-nf4:23 constraints                   &  0.098463 &  1.103473 &  0.846399 &  1.438629 & 0.467    &     \\
FP8:23 constraints                       & -0.170574 &  0.843181 &  0.645007 &  1.102241 & 0.212    &     \\
GPTQ-int4:23 constraints                 & -0.180124 &  0.835166 &  0.630303 &  1.106614 & 0.210    &     \\
GPTQ-int8:23 constraints                 &  0.148169 &  1.159709 &  0.888618 &  1.513502 & 0.275    &     \\
AWQ-int4:39 constraints                  & -0.043569 &  0.957366 &  0.735537 &  1.246096 & 0.746    &     \\
BNB-nf4:39 constraints                   &  0.002067 &  1.002069 &  0.777500 &  1.291502 & 0.987    &     \\
FP8:39 constraints                       & -0.109034 &  0.896700 &  0.693027 &  1.160229 & 0.407    &     \\
GPTQ-int4:39 constraints                 & -0.151087 &  0.859773 &  0.655834 &  1.127129 & 0.274    &     \\
GPTQ-int8:39 constraints                 &  0.064470 &  1.066594 &  0.826900 &  1.375768 & 0.620    &     \\
\bottomrule
\textsc{Random Effects} & & & & & & \\
\bottomrule
\\
$\sigma^2$ (residual variance) & \multicolumn{5}{l}{1.000} & \\
$\tau_{00}$ (Model) & \multicolumn{5}{l}{0.142} & \\
ICC (Model) & \multicolumn{5}{l}{0.124} & \\
N (Model) & \multicolumn{5}{l}{5} & \\
Observations & \multicolumn{5}{l}{4500} & \\
\bottomrule
\\
$R^2$ (marginal) & \multicolumn{5}{l}{0.487} & \\
$R^2$ (conditional) & \multicolumn{5}{l}{0.743} & \\
\bottomrule
\end{tabular}
}
\caption{\textbf{\csfour{}}: Summary of generalized linear mixed model with \textbf{quantization}, \textbf{input constraint level}, and their interaction as predictors of \textbf{accuracy}: \texttt{glmer(cbind(Success, Failures) $\sim$ Quantization * Input\_constraints + (1|Model), data = data, family = binomial)}. The quantization method "BF16" at 7 constraints was set as the reference level (intercept), with \texttt{Model} set as the grouping factor for a random intercept to account for variability across models. See Table~\ref{tab:cs4-interact-posthoc} for post-hoc analysis.}
\label{tab:cs4-interact}
\vspace{0.5em}
\end{table*}

\begin{table*}[ht]
\centering
\resizebox{0.7\textwidth}{!}{
\begin{tabular}{l
                l
                S[table-format=1.3]
                S[table-format=1.3]
                S[table-format=1.3]
                S[table-format=1.3]
                S[table-format=1.3]
                c}
\toprule
\textsc{Constraints} & \textsc{Contrast} & \textsc{Estimate} & \textsc{SE} & \textsc{OR} & \textsc{Probability} & \texttt{p-value} & \\
\midrule
7 & BF16 - (AWQ-int4)           & -0.076 & 0.130 & 0.927 & 0.481 & 1.000 & \\
7 & BF16 - (BNB-nf4)            &  0.116 & 0.125 & 1.123 & 0.529 & 1.000 & \\
7 & BF16 - FP8                  &  0.032 & 0.127 & 1.033 & 0.508 & 1.000 & \\
7 & BF16 - (GPTQ-int4)          & -0.204 & 0.133 & 0.815 & 0.449 & 1.000 & \\
7 & BF16 - (GPTQ-int8)          &  0.102 & 0.125 & 1.107 & 0.525 & 1.000 & \\
7 & (AWQ-int4) - (BNB-nf4)      &  0.192 & 0.127 & 1.212 & 0.548 & 1.000 & \\
7 & (AWQ-int4) - FP8            &  0.108 & 0.129 & 1.114 & 0.527 & 1.000 & \\
7 & (AWQ-int4) - (GPTQ-int4)    & -0.128 & 0.136 & 0.880 & 0.468 & 1.000 & \\
7 & (AWQ-int4) - (GPTQ-int8)    &  0.177 & 0.127 & 1.194 & 0.544 & 1.000 & \\
7 & (BNB-nf4) - FP8             & -0.084 & 0.124 & 0.919 & 0.479 & 1.000 & \\
7 & (BNB-nf4) - (GPTQ-int4)     & -0.320 & 0.131 & 0.726 & 0.421 & 0.213 & \\
7 & (BNB-nf4) - (GPTQ-int8)     & -0.015 & 0.122 & 0.985 & 0.496 & 1.000 & \\
7 & FP8 - (GPTQ-int4)           & -0.236 & 0.133 & 0.790 & 0.441 & 1.000 & \\
7 & FP8 - (GPTQ-int8)           &  0.069 & 0.124 & 1.072 & 0.517 & 1.000 & \\
7 & (GPTQ-int4) - (GPTQ-int8)   &  0.305 & 0.131 & 1.357 & 0.576 & 0.296 & \\
23 & BF16 - (AWQ-int4)           &  0.010 & 0.053 & 1.010 & 0.502 & 1.000 & \\
23 & BF16 - (BNB-nf4)            &  0.018 & 0.053 & 1.018 & 0.504 & 1.000 & \\
23 & BF16 - FP8                  &  0.203 & 0.051 & 1.225 & 0.551 & 0.001 & ** \\
23 & BF16 - (GPTQ-int4)          & -0.024 & 0.053 & 0.976 & 0.494 & 1.000 & \\
23 & BF16 - (GPTQ-int8)          & -0.047 & 0.053 & 0.954 & 0.488 & 1.000 & \\
23 & (AWQ-int4) - (BNB-nf4)      &  0.008 & 0.052 & 1.008 & 0.502 & 1.000 & \\
23 & (AWQ-int4) - FP8            &  0.193 & 0.051 & 1.213 & 0.548 & 0.002 & ** \\
23 & (AWQ-int4) - (GPTQ-int4)    & -0.034 & 0.053 & 0.967 & 0.492 & 1.000 & \\
23 & (AWQ-int4) - (GPTQ-int8)    & -0.056 & 0.053 & 0.945 & 0.486 & 1.000 & \\
23 & (BNB-nf4) - FP8             &  0.185 & 0.051 & 1.203 & 0.546 & 0.004 & ** \\
23 & (BNB-nf4) - (GPTQ-int4)     & -0.042 & 0.053 & 0.959 & 0.490 & 1.000 & \\
23 & (BNB-nf4) - (GPTQ-int8)     & -0.065 & 0.053 & 0.937 & 0.484 & 1.000 & \\
23 & FP8 - (GPTQ-int4)           & -0.227 & 0.051 & 0.797 & 0.444 & 0.000 & *** \\
23 & FP8 - (GPTQ-int8)           & -0.249 & 0.052 & 0.779 & 0.438 & 0.000 & *** \\
23 & (GPTQ-int4) - (GPTQ-int8)   & -0.023 & 0.053 & 0.977 & 0.494 & 1.000 & \\
39 & BF16 - (AWQ-int4)           & -0.032 & 0.035 & 0.968 & 0.492 & 1.000 & \\
39 & BF16 - (BNB-nf4)            &  0.114 & 0.035 & 1.121 & 0.529 & 0.015 & * \\
39 & BF16 - FP8                  &  0.141 & 0.035 & 1.152 & 0.535 & 0.001 & ** \\
39 & BF16 - (GPTQ-int4)          & -0.053 & 0.036 & 0.948 & 0.487 & 1.000 & \\
39 & BF16 - (GPTQ-int8)          &  0.037 & 0.035 & 1.038 & 0.509 & 1.000 & \\
39 & (AWQ-int4) - (BNB-nf4)      &  0.146 & 0.035 & 1.158 & 0.537 & 0.000 & *** \\
39 & (AWQ-int4) - FP8            &  0.173 & 0.035 & 1.189 & 0.543 & 0.000 & *** \\
39 & (AWQ-int4) - (GPTQ-int4)    & -0.021 & 0.036 & 0.979 & 0.495 & 1.000 & \\
39 & (AWQ-int4) - (GPTQ-int8)    &  0.069 & 0.035 & 1.072 & 0.517 & 0.752 & \\
39 & (BNB-nf4) - FP8             &  0.027 & 0.034 & 1.027 & 0.507 & 1.000 & \\
39 & (BNB-nf4) - (GPTQ-int4)     & -0.167 & 0.035 & 0.846 & 0.458 & 0.000 & *** \\
39 & (BNB-nf4) - (GPTQ-int8)     & -0.077 & 0.035 & 0.926 & 0.481 & 0.384 & \\
39 & FP8 - (GPTQ-int4)           & -0.194 & 0.035 & 0.824 & 0.452 & 0.000 & *** \\
39 & FP8 - (GPTQ-int8)           & -0.104 & 0.035 & 0.901 & 0.474 & 0.038 & * \\
39 & (GPTQ-int4) - (GPTQ-int8)   &  0.090 & 0.035 & 1.094 & 0.522 & 0.167 & \\
\bottomrule
\end{tabular}
}
\caption{\textbf{\csfour{}}: Post-hoc comparisons between quantization methods for \textbf{accuracy} (Table~\ref{tab:cs4-interact}) across constraint levels using Bonferroni adjustments for multiple pairwise comparisons. The probability values refer to the likelihood that the quantization method listed first in each contrast is more accurate than the one listed second (i.e., a value of 0.5 suggests that both methods are comparable in terms of accuracy).}
\label{tab:cs4-interact-posthoc}
\vspace{0.5em}
\end{table*}

\end{document}